\documentclass[english]{article}

\usepackage{geometry}
\usepackage[T1]{fontenc}
\usepackage[latin9]{inputenc}
\usepackage{bm}
\usepackage{amsmath,mathtools}
\usepackage{amssymb}
\usepackage[unicode=true,
 bookmarks=false,
 breaklinks=false,pdfborder={0 0 1},colorlinks=false]
 {hyperref}
\hypersetup{
 colorlinks,citecolor=blue,filecolor=blue,linkcolor=blue,urlcolor=blue}

\makeatletter
\usepackage{amsthm}
\usepackage{cite}  
\usepackage{comment}
\usepackage{natbib}
\usepackage{booktabs}

\usepackage{graphicx}

\usepackage[linesnumbered,ruled,vlined]{algorithm2e}

\SetCommentSty{mycommfont}
\usepackage{algorithmic}

\usepackage{float}
\usepackage{multirow}
 
\usepackage{dsfont}
\usepackage{tcolorbox}

\usepackage{color}
\definecolor{yxc}{RGB}{255,0,0}
\definecolor{yjc}{RGB}{125,0,0}
\definecolor{ytw}{RGB}{255,69,0}
\definecolor{gen}{RGB}{0,0,200}

\allowdisplaybreaks

\DeclareMathOperator{\ind}{\mathds{1}}  

\newcommand{\real}{\mathbb{R}}

\newcommand{\dist}{\mathsf{KL}_{\mathsf{gen}}}
\newcommand{\KL}{\mathsf{KL}}

\newcommand{\trunc}{\mathsf{adjust}}

\definecolor{yanxi}{RGB}{0,200,100}

\title{Fast Computation of Optimal Transport via \\ Entropy-Regularized Extragradient Methods}

\author{Gen Li\thanks{Department of Statistics, Chinese University of Hong Kong.} 
	\and Yanxi Chen\thanks{Department of Electrical and Computer Engineering, Princeton University, Princeton, NJ 08544, U.S.A.}
	\and Yu Huang\thanks{Department of Statistics and Data Science, Wharton School, University of Pennsylvania, Philadelphia, PA 19104, U.S.A.}
	\and Yuejie Chi\thanks{Department of Electrical and Computer Engineering, Carnegie Mellon University, Pittsburgh, PA 15213, U.S.A.}
	\and H.~Vincent Poor\footnotemark[2]
	\and Yuxin Chen\footnotemark[3]}
\date{January 2023; Revised: June 2024}

\makeatother

\begin{document}

\theoremstyle{plain} \newtheorem{lemma}{\textbf{Lemma}}\newtheorem{proposition}{\textbf{Proposition}}\newtheorem{theorem}{\textbf{Theorem}}

\theoremstyle{remark}\newtheorem{remark}{\textbf{Remark}}

\maketitle

\begin{abstract}
	Efficient computation of the optimal transport distance between two distributions serves as an algorithm subroutine that empowers various applications. 
	This paper develops a scalable first-order optimization-based method that computes optimal transport to within $\varepsilon$ additive accuracy 
	with runtime $\widetilde{O}( n^2/\varepsilon)$, where $n$ denotes the dimension of the probability distributions of interest.  
	Our algorithm achieves the state-of-the-art computational guarantees among all first-order methods, while exhibiting favorable numerical performance compared to classical algorithms like Sinkhorn and Greenkhorn. 
	Underlying our algorithm designs are two key elements: (a) converting the original problem into a bilinear minimax problem over probability distributions; 
	(b) exploiting the extragradient idea --- in conjunction with entropy regularization and adaptive learning rates --- to accelerate convergence. 
\end{abstract}


\tableofcontents

\section{Introduction}
\label{sec:intro}

Quantifying the distance between two probability distributions is an algorithm subroutine that permeates and empowers a wealth of modern data science applications.  
For instance, how to measure the difference between the model distribution and the real distribution in generative adversarial networks \cite{arjovsky2017wasserstein}, 
how to evaluate the intrinsic dissimilarity between two point clouds in computer graphics \cite{solomon2015convolutional,kim2013guided},  
and how to assess the distribution shift in transfer learning \cite{gayraud2017optimal},   
are all representative examples built upon probability distances.

This paper is about computing an elementary instance within this arena that gains increasing popularity, 
that is,  the optimal transport distance between two distributions \cite{peyre2019computational}.  
It sometimes goes by the name of the earth mover's distance \cite{werman1985distance,rubner2000earth,pele2009fast} or the Wasserstein distance \cite{villani2009optimal}. 
In light of the celebrated Kantorovich relaxation \cite{kantorovich1942translocation,peyre2019computational}, computing the optimal transport between two $n$-dimensional probability distributions can be cast as solving a linear program over probability matrices with fixed marginals:  
\begin{align}
	\begin{array}{cl}
	\mathop{\text{minimize}}\limits_{\bm{P} \in \mathbb{R}^{n\times n}} \qquad &\langle \bm{W}, \bm{P}\rangle \qquad \\
	\text{subject to} \qquad & \bm{P} \ge \bm{0}, \, \bm{P}\bm{1} = \bm{r},  \,\bm{P}^{\top}\bm{1} = \bm{c}.  
	\end{array}
	\label{eq:original}
\end{align}
Here, $\bm{r}=[r_i]_{1\leq i\leq n}$ and $\bm{c}=[c_i]_{1\leq i\leq n}$ are $n$-dimensional probability vectors representing the prescribed row and column marginals, respectively, 
 $\bm{W} = [w_{i,j}]_{1\leq i,j\leq n}\in \mathbb{R}_+^{n\times n}$ stands for a given non-negative cost matrix (so that the objective function $\langle \bm{W}, \bm{P} \rangle$ 
measures the total transportation cost), and $\bm{1}$ denotes the all-one vector.  
In a nutshell, the optimal transport problem amounts to finding the most cost-efficient reshaping of one distribution into another, 
or equivalently, the most economical coupling of the two distributions.


At first glance, the optimal transport problem \eqref{eq:original} seems to be readily solvable via relatively mature toolboxes in linear programming.   
Nevertheless, the unprecedentedly large problem dimensionality in contemporary applications calls for a thorough (re)-examination of existing algorithms, 
so as to ensure feasibility of computing optimal transport at scale. 
For example, the linear-programming-based algorithm \cite{lee2014path} requires a runtime $\widetilde{O}(n^{2.5})$, 
which takes much longer than the time needed to read the cost matrix $\bm{W}$. 
In comparison, another alternative tailored to this problem, called Sinkhorn iteration  \cite{sinkhorn1967diagonal,cuturi2013sinkhorn}, 
exploits the special structure underlying the solution to an entropy-regularized variant of \eqref{eq:original}.  
This classical approach and its variants have been shown to be near linear-time  \cite{altschuler2017near,dvurechensky2018computational,lin2022efficiency}, 
attaining $\varepsilon$ additive accuracy\footnote{A {\em feasible} point $\widetilde{\bm{P}}$ is said to achieve $\varepsilon$ additive accuracy if $\langle \bm{W}, \widetilde{\bm{P}} \rangle\leq \langle \bm{W}, \bm{P}^{\star} \rangle + \varepsilon$, where $\bm{P}^{\star}$ is an optimal solution.} with a computational complexity of  $\widetilde{O}(n^{2}/\varepsilon^2)$.
Despite their favorable scaling in $n$, however, the Sinkhorn-type algorithms fall short of achieving optimal scaling in $\varepsilon$, 
thereby stimulating further pursuit for theoretical improvement.    
\cite{blanchet2018towards,quanrud2018approximating} led this line of studies by developing the first algorithms with theoretical runtime $\widetilde{O}(n^{2}/\varepsilon)$, 
although practical implementation of these algorithms remain unavailable.\footnote{These methods invoke  black-box subroutines (e.g., positive linear programs, matrix scaling) that remain impractical so far.}  
\cite{jambulapati2019direct} went on to propose an implementable {\em first-order method} --- i.e., dual extrapolation --- that enjoys matching complexity $\widetilde{O}(n^{2}/\varepsilon)$;   
however, this method is numerically outperformed by Sinkhorn iteration as reported in their experiments.   
Another recent breakthrough in theoretical computer science \cite{van2020bipartite}  solved a more general class of problems called maximum cardinality bipartite matching 
and showed that even logarithmic $\varepsilon$-dependency is feasible; the lack of practical implementation, once again, inhibits real-world adoption.

In sum, despite an exciting line of theoretical advances towards solving optimal transport, 
there is significant mismatch between the state-of-the-art theoretical results and the practical runtime. 
This motivates one to pursue other algorithmic alternatives that could be appealing in both theory and practice.


\begin{table}

\begin{centering}
\small
\begin{tabular}{ccccc}
\toprule 
\multirow{2}*{reference} & \multirow{2}*{runtime} & \multirow{2}*{algorithm} & first-order & \multirow{2}*{implementable?}   \tabularnewline
 &  &  & method? &    \tabularnewline
\toprule 
\cite{altschuler2017near} & $n^2 / \varepsilon^3$ & Sinkhorn & yes & yes \tabularnewline
\cite{dvurechensky2018computational} & $n^2 / \varepsilon^2$ & Sinkhorn & yes & yes \tabularnewline
\cite{lin2022efficiency} & $n^2 / \varepsilon^2$  & Greenkhorn & yes & yes \tabularnewline
\midrule
	\cite{dvurechensky2018computational} & \multirow{2}*{$n^{2.5} / \varepsilon$}  & \multirow{2}*{APDAGD} & \multirow{2}*{yes} & \multirow{2}*{yes} \tabularnewline
\cite{lin2022efficiency} &   &  &  &  \tabularnewline


\cite{guminov2021combination} & $n^{2.5} / \varepsilon$ & AAM & yes & yes \tabularnewline

\cite{lin2022efficiency} & $n^{2.5} / \varepsilon$  & APDAMD & yes & yes \tabularnewline

\cite{guo2020fast} & $n^{2.5} / \varepsilon$  & APDRCD & yes & yes \tabularnewline

\cite{an2022efficient} & $n^{2.5} / \varepsilon$  & Nesterov's smoothing & yes & yes \tabularnewline

\cite{chambolle2022accelerated} & $n^{2.5} / \varepsilon$ & HPD & yes & yes \tabularnewline

\cite{xie2022accelerated} & $n^{2.5} / \varepsilon$ & PDASGD & yes & yes \tabularnewline
\midrule
\cite{blanchet2018towards} &  $n^2 / \varepsilon$ & packing linear program & yes & --- \tabularnewline
\cite{quanrud2018approximating} & $n^2 / \varepsilon$ & packing linear program & yes & --- \tabularnewline

\cite{blanchet2018towards} &  $n^2 / \varepsilon$ & matrix scaling & no & --- \tabularnewline

\cite{lahn2019graph} &  $n^2 / \varepsilon + n / \varepsilon^2$ & combinatorial & no & yes \tabularnewline

\multirow{2}*{\cite{van2020bipartite}} & \multirow{2}*{$n^2 \log^2(1/\varepsilon)$} & max-cardinality & \multirow{2}*{no} & \multirow{2}*{---} \tabularnewline

 & & bipartite matching &  &  \tabularnewline

\midrule
\cite{jambulapati2019direct} & $n^2 / \varepsilon$ & dual extrapolation & yes & yes \tabularnewline

\midrule
\textbf{This work} &  $n^2 / \varepsilon$  &  extragradient  & yes & yes  \tabularnewline
\bottomrule
\end{tabular}
\par\end{centering}

\caption{Comparisons with prior works. Here, we assume $\|\bm{W}\|_{\infty} = 1$ without loss of generality,  
	and we omit all logarithmic factors except for \cite{van2020bipartite} (as its $\varepsilon$-dependency is only logarithmic). 
	In the last column, ``---'' means that no practical implementation has been available so far. 
\label{tab:literature}}

\end{table}

\paragraph{Main contributions.} 
This paper contributes to the above growing literature by proposing a scalable algorithm tailored to the optimal transport problem.  
Our focus is first-order optimization-based methods --- a family of practically appealing algorithms for large-scale optimization. 
Our algorithm is built around several key ideas.  
\begin{itemize}
	\item[1)] We start with an $\ell_1$-penalized variant of the original problem, and reformulate it into a bilinear minimax problem over two sets of probability distributions.  
		
	\item[2)] In an attempt to solve this minimax problem, we design a variant of entropy-regularized extragradient methods. 
		On a high level, the algorithm performs two mirror-descent-type updates per iteration, with learning rates chosen adaptively in accordance with the corresponding row or column marginals. 
\end{itemize}
%
Encouragingly, the proposed entropy-regularized extragradient method is capable of achieving $\varepsilon$ additive accuracy with\footnote{Here and throughout, $f(n, 1/\varepsilon)=O( g(n,1/\varepsilon))$ means there exists a universal constant $C$ such that $|f(n,1/\varepsilon)|\leq C \cdot g(n,1/\varepsilon)$ for all $n$ and $1/\varepsilon$. The notation $\widetilde{O}(\cdot)$ is defined similarly except that it hides all logarithmic factors. }
\begin{equation}
	\widetilde{O}\big( 1/\varepsilon \big) \text{ iterations} \qquad \text{or} \qquad \widetilde{O}\big( n^2/\varepsilon \big) \text{ runtime}, 
\end{equation}
thus constituting a nearly linear-time algorithm with desired iteration complexity; see Theorem~\ref{thm:theory-OT}. 
Table~\ref{tab:literature} provides more detailed comparisons with previous results. 
In short, our algorithm enjoys computational guarantees that match the best-known theory (i.e., \cite{jambulapati2019direct})
among all first-order methods for computing optimal transport,  
while at the same time compare favorably to 
the classical Sinkhorn and Greenkhorn algorithms in numerical experiments.

\paragraph{Notation.} 
Let $\Delta_n \coloneqq \{\bm{x}\in \mathbb{R}^n \mid \bm{x}\geq \bm{0}, \bm{1}^{\top}\bm{x}=1\}$ and $\Delta_{n\times n} \coloneqq \{\bm{X}\in \mathbb{R}^{n\times n} \mid \bm{1}^{\top}\bm{X}\bm{1}=1, X_{i,j}\geq 0, \forall i,j \}$
denote the $n$-dimensional and $n\times n$-dimensional probability simplices, respectively. 
For any probability vector $\bm{p}\in \Delta_d$, its entropy is defined by $\mathcal{H}(\bm{p}) \coloneqq - \sum_{i=1}^d p_i \log p_i$.  
For any probability vectors $\bm{p}, \bm{q}\in \Delta_d$, 
the Kullback-Leibler (KL) divergence of between $\bm{p}$ and $\bm{q}$ is defined by $\KL(\bm{p}\parallel \bm{q}) \coloneqq \sum_i p_i \log \frac{p_i}{q_i} $. 
For any matrix $\bm{W}=[W_{i,j}]_{1\leq i,j\leq n} \in \mathbb{R}^{n\times n}$, 
we denote by $\|\bm{W}\|_{\infty} \coloneqq \max_{1\leq i,j\leq n} |W_{i,j}|$ its entrywise infinity norm, 
and $\|\bm{W}\|_{1} \coloneqq \sum_{1\leq i,j\leq n} |W_{i,j}|$ its entrywise $\ell_1$ norm. 
Let $\mathbb{R}^{n\times n}_+$ be the set of all $n\times n$ matrices with non-negative entries. 
For any matrix $\bm{F}\in \mathbb{R}^{n\times n}_+$, 
let $\mathsf{row}(\bm{F}) \coloneqq \bm{F}\bm{1}$ (resp.~$\mathsf{col}(\bm{F}) \coloneqq \bm{F}^{\top}\bm{1}$)  represent an $n$-dimensional vector consisting of all row sums (resp.~column sums) of $\bm{F}$, 
and let $\mathsf{row}_i(\bm{F})$ (resp.~$\mathsf{col}_i(\bm{F})$) be the sum of the $i$-th row (resp.~column) of $\bm{F}$. 
For any vector $\bm{x}=[x_i]_{1\leq i\leq n}\in \mathbb{R}^n$, 
we denote by $\mathsf{diag}(\bm{x})\in \mathbb{R}^{n\times n}$ a diagonal matrix whose diagonals contain the entries of $\bm{x}$.

\section{Algorithm and main results}
\label{sec:algorithm}

In this section, we present our algorithm design, followed by its convergence guarantees. 
Before continuing, let us introduce several more notation that facilitates our discussion. 
Let $\bm{w}_{i}\in \mathbb{R}^n$ represent the $i$-th row of $\bm{W}$;
for any $\bm{P}\in \mathbb{R}^{n\times n}_+$ obeying $\bm{P}\bm{1}=\bm{r}$,  
introduce a collection of probability vectors $\{\bm{p}_i\in \Delta_n\}$ such that 
$r_i\bm{p}_i$ represents the $i$-th row of $\bm{P}$; 
similarly, we denote by $\bm{P}^{\star}$ a solution to the problem~\eqref{eq:original}, 
and employ $r_i \bm{p}_i^{\star}$ to represent the $i$-th row of $\bm{P}^{\star}$ (so that $\bm{p}_i^{\star}$ is a probability vector). 
In other words, we can write  
\begin{equation}
\bm{W}=\left[\begin{array}{c}
	\bm{w}_{1}^{\top}\\
\vdots\\
\bm{w}_{n}^{\top}
\end{array}\right],
\qquad
\bm{P}=\left[\begin{array}{c}
r_{1}\bm{p}_{1}^{\top}\\
\vdots\\
r_{n}\bm{p}_{n}^{\top}
\end{array}\right],
\qquad\text{and}\qquad
\bm{P}^{\star}=\left[\begin{array}{c}
r_{1}\bm{p}_{1}^{^{\star}\top}\\
\vdots\\
r_{n}\bm{p}_{n}^{^{\star}\top}
\end{array}\right].
	\label{eq:defn-li-pi}
\end{equation}
Armed with the above notation, 
we can readily reformulate~\eqref{eq:original} as follows
\begin{align}
	\begin{array}{cl}
		\mathop{\text{minimize}}
		\limits_{\{\bm{p}_i \}_{i = 1}^n} 
		\qquad &\sum_{i = 1}^n r_i\langle \bm{w}_i, \bm{p}_i\rangle \\
		\text{subject to}\qquad & \bm{p}_i \in \Delta_n ~(1\leq i\leq n), ~ \sum_{i = 1}^n r_i \bm{p}_i = \bm{c},  
	\end{array}
	\label{eq:original-new}
\end{align}
for which $\{\bm{p}_i^{\star}\}_{i=1}^n$ stands for an optimal solution.

\subsection{Approximate transportation solution of a penalized variant}

The equivalent formulation \eqref{eq:original-new}  motivates one to look at a related $\ell_1$-penalized problem:
%
\begin{align}
	\mathop{\text{minimize}}\limits_{\{\bm{p}_i \in \Delta_n\}_{i = 1}^n} 
	\qquad f_{\ell_1}(\{\bm{p}_i\})\coloneqq \frac{1}{2}\sum_{i = 1}^n r_i\langle \bm{w}_i, \bm{p}_i\rangle +  
	\|\bm{W}\|_{\infty}
	\bigg\|\sum_{i = 1}^n r_i \bm{p}_{i} - \bm{c}\bigg\|_1 , \label{eq:unconstrained}
\end{align}
a trick that has been inspired by ideas in prior optimization literature; see, e.g., \cite{jambulapati2019direct}. 
Evidently, if we are able to compute an $\varepsilon$-optimal solution $\{\widehat{\bm{p}}_i\}_{i = 1}^n$ to~\eqref{eq:unconstrained} 
(in the sense that $f_{\ell_1}(\{\widehat{\bm{p}}_i\})\leq \min_{\{\bm{p}_i \in \Delta_n\}_{i = 1}^n}f_{\ell_1}(\{\bm{p}_i\})$, 
then one  has
\begin{align}
	0.5\sum\nolimits_{i=1}^{n}r_{i}\langle\bm{w}_{i},\widehat{\bm{p}}_{i}\rangle
	& \le 0.5\sum\nolimits_{i=1}^{n}r_{i}\langle\bm{w}_{i},\widehat{\bm{p}}_{i}\rangle+ \|\bm{W}\|_{\infty} \Big\| r_{i}\sum\nolimits_{i=1}^{n}\widehat{\bm{p}}_{i}-\bm{c}\Big\|_{1} \notag\\
	& = f_{\ell_1}(\{\widehat{\bm{p}}_i\})\leq \min_{\{\bm{p}_i \in \Delta_n\}_{i = 1}^n}f_{\ell_1}(\{\bm{p}_i\}) + \varepsilon 
	\leq f_{\ell_1}(\{\bm{p}^{\star}_i\}) + \varepsilon \notag\\
	&\le 0.5\sum\nolimits_{i=1}^{n}r_{i}\langle\bm{w}_{i},\bm{p}_{i}^{\star}\rangle+ \|\bm{W}\|_{\infty} \Big\| r_{i}\sum\nolimits_{i=1}^{n} \bm{p}^{\star}_{i}-\bm{c}\Big\|_{1} + \varepsilon \notag\\
	& 
	=0.5\sum\nolimits_{i=1}^{n}r_{i}\langle\bm{w}_{i},\bm{p}_{i}^{\star}\rangle+ \varepsilon.  
	\label{eq:condition}
\end{align}
%


In general, however, 
the solution to \eqref{eq:unconstrained}  does not satisfy the feasibility constraints of the optimal transport problem, 
and one still needs to convert it into a feasible transportation plan.  
This can be  accomplished via the following result \cite[Lemma 7]{altschuler2017near}. 
\begin{lemma} \label{lem:projection}
	For any non-negative matrix $\widehat{\bm{P}} \in \mathbb{R}^{n\times n}_+$, 
	there exists a fast algorithm (see \cite[Algorithm~2]{altschuler2017near}) 
	that is able to find a probability matrix $\widetilde{\bm{P}}\in \Delta_{n\times n}$ 
	with $O(n^2)$ computation complexity such that
\begin{align}
	\widetilde{\bm{P}}\bm{1} = \bm{r}, \quad \widetilde{\bm{P}}^{\top}\bm{1} = \bm{c}, \quad \text{and} \quad
	\|\widehat{\bm{P}} - \widetilde{\bm{P}}\|_1 \le 2\big(\| \widehat{\bm{P}} \bm{1} - \bm{r}\|_1 + \|\widehat{\bm{P}}^{\top}\bm{1} - \bm{c}\|_1 \big).
\end{align}
\end{lemma}
As a consequence, it boils down to finding a near-optimal solution $\{\widehat{\bm{p}}_i\}_{1\leq i\leq n}$ to \eqref{eq:unconstrained} in a computationally efficient manner, while  ensuring sufficiently small 
$\| \widehat{\bm{P}} \bm{1} - \bm{r}\|_1 + \|\widehat{\bm{P}}^{\top}\bm{1} - \bm{c}\|_1$, 
with $\widehat{\bm{P}}$ defined as 
{\small $ \widehat{\bm{P}}=\left[\begin{array}{c}
	r_{1}\widehat{\bm{p}}_{1}^{\top}\\
\vdots\\
	r_{n}\widehat{\bm{p}_{n}}^{\top}
\end{array}\right].$}
Given that $\widehat{\bm{p}}_{i}^{\top}\bm{1}=1$ in this case, this quantity $\| \widehat{\bm{P}} \bm{1} - \bm{r}\|_1 + \|\widehat{\bm{P}}^{\top}\bm{1} - \bm{c}\|_1$ that needs to be controlled is 
 equivalent to $\|\widehat{\bm{P}}^{\top}\bm{1}-\bm{c}\|_{1}+\|\widehat{\bm{P}}\bm{1}-\bm{r}\|_{1}
 = \|\sum_{i=1}^{n}r_{i}\widehat{\bm{p}}_{i}-\bm{c}\|_{1}+\sum_{i=1}^{n}\big|r_{i}\widehat{\bm{p}}_{i}^{\top}\bm{1}-r_{i}\big|=\| \sum_{i=1}^n r_i \widehat{\bm{p}}_i - \bm{c} \|_1$.

\subsection{Entropy-regularized extragradient methods}

In this subsection, we propose a method for solving the $\ell_1$-penalized problem \eqref{eq:unconstrained}. 
To streamline the presentation, we assume without loss of generality that $\|\bm{W}\|_{\infty} = 1$; 
this can be implemented by running $\bm{W} \leftarrow \bm{W} / \|\bm{W}\|_{\infty}$ at the very beginning of the algorithm. 

\paragraph{An equivalent minimax problem and entropy regularization.}  
The first step of our algorithm lies in converting the objective function of \eqref{eq:unconstrained} into a bilinear function, 
for which the key lies in handling the $\ell_1$ penalty term. 
Towards this end, we introduce a set of auxiliary 2-dimensional probability vectors $\bm{\mu}_j = [\mu_{j, +}, \mu_{j, -}] \in \Delta_2$ $(1\leq j\leq n)$. 
As can be easily verified, 
this allows us to recast the objective of \eqref{eq:unconstrained} as follows:
\begin{equation}
\frac{1}{2}\sum_{i=1}^{n}r_{i}\langle\bm{w}_{i},\bm{p}_{i}\rangle+ \bigg\|\sum_{i=1}^{n}r_{i}\bm{p}_{i}-\bm{c}\bigg\|_{1}
	=\max_{\bm{\mu}_{j}\in\Delta_{2},\forall j}f\big(\{\bm{p}_{i}\}_{i=1}^{n},\{\bm{\mu}_{j}\}_{j=1}^{n}\big),
	\label{eq:equiv-objective-f-246}
\end{equation}
where we define 
\begin{align}
f\big(\{\bm{p}_{i}\}_{i=1}^{n},\{\bm{\mu}_{j}\}_{j=1}^{n}\big) & \coloneqq\frac{1}{2}\sum_{i=1}^{n}r_{i}\langle\bm{w}_{i},\bm{p}_{i}\rangle 
	+ \sum_{j=1}^{n}\big(\mu_{j,+}-\mu_{j,-}\big)\bigg(\sum_{i=1}^{n}r_{i}p_{i,j}-c_{j}\bigg)\nonumber \\
 & =\frac{1}{2}\sum_{i=1}^{n}r_{i}\langle\bm{w}_{i},\bm{p}_{i}\rangle-\sum_{j=1}^{n}(\mu_{j,+}-\mu_{j,-})c_{j}+\sum_{i=1}^{n}\sum_{j=1}^{n}r_{i}(\mu_{j,+}-\mu_{j,-})p_{i,j}.
	\label{eq:defn-f-function}
\end{align}
Armed with this function, one can recast \eqref{eq:unconstrained} as the following minimax problem: 
\begin{equation}
\mathop{\text{minimize}}_{\bm{p}_{i}\in\Delta_{n},\forall i}~\max_{\bm{\mu}_{j}\in\Delta_{2},\forall j}f\big(\{\bm{p}_{i}\}_{i=1}^{n},\{\bm{\mu}_{j}\}_{j=1}^{n}\big),\label{eq:equiv-minimax-f}
\end{equation}
or equivalently (by virtue of von Neumann's minimax theorem \cite{v1928theorie}), 
\begin{equation}
\mathop{\text{maximize}}_{\bm{\mu}_{j}\in\Delta_{2},\forall j}\min_{\bm{p}_{i}\in\Delta_{n},\forall i}f\big(\{\bm{p}_{i}\}_{i=1}^{n},\{\bm{\mu}_{j}\}_{j=1}^{n}\big). \label{eq:equiv-maximin-f}
\end{equation}

Given that the bilinear objective function is convex-concave but not strongly-convex-strongly-concave, 
one strategy for accelerating the optimization procedure is to augment the objective function with entropy regularization terms. 
This leads to the following entropy-regularized minimax problem: 
\begin{align}
	\mathop{\text{maximize}}_{\bm{\mu}_{j}\in\Delta_{2},\forall j}\min_{\bm{p}_{i}\in\Delta_{n},\forall i}&
	F\big(\{\bm{p}_{i}\}_{i=1}^{n},\{\bm{\mu}_{j}\}_{j=1}^{n}\big)\coloneqq 
	\notag\\
	& f\big(\{\bm{p}_{i}\}_{i=1}^{n},\{\bm{\mu}_{j}\}_{j=1}^{n}\big)+\sum\nolimits_{j=1}^{n}\tau_{\mu,j}\mathcal{H}(\bm{\mu}_{j})-\sum\nolimits_{i=1}^{n}\tau_{p,i}\mathcal{H}(\bm{p}_{i}),
	\label{eq:game-entropy}
\end{align}
where $\mathcal{H}(\cdot)$ denotes the entropy (which is a strongly concave and non-negative function), and  $\{\tau_{\mu,j}\}_{1\leq j\leq n}$ and $\{\tau_{p,i}\}_{1\leq i\leq n}$ are a set of {\em positive} regularization parameters that we shall specify momentarily. 
The remainder of this subsection is dedicated to solving \eqref{eq:game-entropy} in an efficient fashion.



\paragraph{An extragradient method for solving \eqref{eq:game-entropy}.} 
%
%
The family of extragradient methods has proven effective for solving convex-concave minimax problems \cite{korpelevich1976extragradient,tseng1995linear,harker1990finite,mokhtari2020unified}. 
Inspired by a recent development \cite{cen2021fast}, 
we propose to solve \eqref{eq:game-entropy} by means of a variant of extragradient methods.

Let us begin by introducing a basic operation. 
Suppose the current iterate is $\big(\{\bm{p}_{i}^{\mathsf{current}}\}_{i=1}^{n},\{\bm{\mu}_{j}^{\mathsf{current}}\}_{j=1}^{n}\big)$.  
One step of mirror descent (with the KL divergence chosen to monitor the displacement) takes the following form: for $1\leq i, j\leq n$,
{\small
\begin{subequations}
\label{eq:basic-MD-step}
\begin{align}
\bm{\mu}_{j}^{\mathsf{next}} & =\arg\max_{\bm{\mu}_{j}\in\Delta_{2}}\left\{ \Big\langle\nabla_{\bm{\mu}_{j}}F\big(\{\bm{p}_{i}^{\mathsf{grad}}\}_{i=1}^{n},\{\bm{\mu}_{j}^{\mathsf{grad}}\}_{j=1}^{n}\big),\,\bm{\mu}_{j}\Big\rangle-\frac{1}{\eta_{\mu,j}}\mathsf{KL}\big(\bm{\mu}_{j}\parallel\bm{\mu}_{j}^{\mathsf{current}}\big)\right\} , 
	\label{eq:basic-MD-step-mu} \\
	\bm{p}_{i}^{\mathsf{next}} & =\arg\min_{\bm{p}_{i}\in\Delta_{n}}\left\{ \Big\langle\nabla_{\bm{p}_{i}}F\big(\{\bm{p}_{i}^{\mathsf{grad}}\}_{i=1}^{n},\{\bm{\mu}_{j}^{\mathsf{grad}}\}_{j=1}^{n}\big),\,\bm{p}_{i}\Big\rangle+\frac{1}{\eta_{p,i}}\mathsf{KL}\big(\bm{p}_{i}\parallel\bm{p}_{i}^{\mathsf{current}}\big)\right\} , 
	\label{eq:basic-MD-step-p}
\end{align}
\end{subequations}
}
or equivalently, 
{\small
\begin{subequations}
\label{eq:basic-MD-step-equiv}
\begin{align}
\mu_{j,s}^{\mathsf{next}} & \propto\big(\mu_{j,s}^{\mathsf{current}}\big)^{1-\eta}\exp\bigg(\eta_{\mu,j}s\Big(\sum_{i=1}^{n}r_{i}p_{i,j}^{\mathsf{grad}}-c_{j}\Big)\bigg),\qquad 
	&& s\in\{+,-\} \label{eq:basic-MD-step-mu-equiv} \\	
p_{i,l}^{\mathsf{next}} & \propto\big(p_{i,l}^{\mathsf{current}}\big)^{1-\eta}\exp\Big(-\eta_{p,i}r_{i}\big(0.5w_{i,l}+\mu_{l,+}^{\mathsf{grad}}-\mu_{l,-}^{\mathsf{grad}}\big)\Big),\qquad 
	&& l=1,\cdots,n \label{eq:basic-MD-step-p-equiv} 
\end{align}
\end{subequations}
}
for all $1\leq i,j\leq n$, 
where $\{\eta_{\mu,j}\}$ and $\{\eta_{p,i}\}$ are two collections of positive learning rates obeying 
$\eta_{\mu,j}\tau_{\mu,j}=\eta$ and $\eta_{p,i}\tau_{p,i}=\eta$ for some quantity $0<\eta\leq 1$ ($\forall i,j$). 
Here, we allow the gradient $\nabla F$ to be evaluated at a point $\big(\{\bm{p}_{i}^{\mathsf{grad}}\}_{i=1}^{n},\{\bm{\mu}_{j}^{\mathsf{grad}}\}_{j=1}^{n}\big)$ deviating 
from the current iterate $\big(\{\bm{p}_{i}^{\mathsf{current}}\}_{i=1}^{n},\{\bm{\mu}_{j}^{\mathsf{current}}\}_{j=1}^{n}\big)$, 
which plays a crucial role in describing the extragradient update rule.

We are now ready to present the proposed method, which maintains several sequences of the iterates; for each iteration $t$, we maintain/update the following iterates:  
\begin{itemize}

	\item {\em Updates w.r.t.~the variables $\{\bm{p}_i\}_{i=1}^n$:} 
		main sequence $\{ \bm{p}_i^t \in \Delta_n \}_{i=1}^n$; 
		midpoints $\{ \overline{\bm{p}}_i^t \in \Delta_n \}_{i=1}^n$. 

	\item {\em Updates w.r.t.~the variables $\{\bm{\mu}_j\}_{j=1}^n$:} 
		main sequence $\{ \bm{\mu}_j^t = [\mu_{j,+}^t, \mu_{j,-}^t] \in \Delta_2 \}_{j=1}^n$; 
		midpoints $\{ \overline{\bm{\mu}}_j^t = [\overline{\mu}_{j,+}^t, \overline{\mu}_{j,-}^t] \in \Delta_2 \}_{j=1}^n$; 
		adjusted main sequence $\{ \bm{\mu}_j^{t,\trunc} = [\mu_{j,+}^{t,\trunc}, \mu_{j,-}^{t,\trunc}] \in \Delta_2 \}_{j=1}^n$.

\end{itemize}
In the $t$-th iteration, the proposed algorithm performs the following three sets of updates, 
with the first two embodying the extragradient idea.  

\begin{itemize}

	\item[1)] {\em Computing the midpoints:} for each $1\leq j\leq n$, 
\begin{subequations}
		\label{eq:update-midpoints}
		{\small
		\begin{equation}
			\overline{\mu}_{j,s}^{t+1}  \propto\big(\mu_{j,s}^{t,\trunc}\big)^{1-\eta}\exp\Big(\eta_{\mu,j}s\Big(\sum\nolimits_{i=1}^{n}r_{i}p_{i,j}^{t}-c_{j}\Big)\Big),\quad s\in\{+,-\}, 
			\label{eq:update-midpoints-123}
		\end{equation}
		}
		and for each $1\leq i\leq n$,
		{\small
		\begin{equation}
			\overline{p}_{i,j}^{t+1}  \propto\big(p_{i,j}^{t}\big)^{1-\eta}\exp\Big(-\eta_{p,i}r_{i}\big(0.5w_{i,j}+\mu_{j,+}^{t,\trunc}-\mu_{j,-}^{t,\trunc}\big)\Big),
			\quad 1\leq j\leq  n, 
		\end{equation}
		}
		with the learning rates $\{\eta_{\mu,j}\}$ and $\{\eta_{p,i}\}$ to be specified shortly. 
		In words, this constitutes one step of mirror descent (cf.~\eqref{eq:basic-MD-step-equiv}) from the point $\big(\{\bm{p}_{i}^{t}\}_{i=1}^{n},\{\bm{\mu}_{j}^{t,\trunc}\}_{j=1}^{n}\big)$, 
		with the gradient evaluated at the same point. 
\end{subequations}

	\item[2)] {\em Updating the main sequence:} for each $1\leq j\leq n$,
\begin{subequations}
		\label{eq:update-main}
		{\small
		\begin{equation}
			\mu_{j,s}^{t+1}  \propto\big(\mu_{j,s}^{t,\trunc}\big)^{1-\eta}\exp\bigg(\eta_{\mu,j}s\Big(\sum\nolimits_{i=1}^{n}r_{i}\overline{p}_{i,j}^{t+1}-c_{j}\Big)\bigg),\qquad s\in\{+,-\}; 
			\label{eq:update-main-135}
		\end{equation}
		}
		and for each $1\leq i\leq n$,
		{\small
		\begin{equation}
			p_{i,j}^{t+1}  \propto\big(p_{i,j}^{t}\big)^{1-\eta}\exp\Big(-\eta_{p,i}r_{i}\big(0.5w_{i,j}+\overline{\mu}_{j,+}^{t+1}-\overline{\mu}_{j,-}^{t+1}\big)\Big),
			\qquad 1\leq j\leq  n.
		\end{equation}
		}
		This implements another step of mirror descent (cf.~\eqref{eq:basic-MD-step-equiv}) from the same point $\big(\{\bm{p}_{i}^{t}\}_{i=1}^{n},\{\bm{\mu}_{j}^{t,\trunc}\}_{j=1}^{n}\big)$ as above, 
		albeit using a gradient evaluated at the midpoint $\big(\{\overline{\bm{p}}_{i}^{t+1}\}_{i=1}^{n},\{ \overline{\bm{\mu}}_{j}^{t+1}\}_{j=1}^{n}\big)$. 
		In a nutshell, the midpoint computed in  the previous step assists in predicting a better search direction.

	\item[3)] {\em Adjusting the current iterates:} for each $1\leq j\leq n$,
%
		{\small
		\begin{equation}
			\mu_{j,s}^{t+1,\trunc}\propto\max\Big\{\mu_{j,s}^{t+1},\,e^{-B}\max\big\{\mu_{j,+}^{t+1},\mu_{j,-}^{t+1} \big\}\Big\},\qquad s\in\{+,-\},
			\label{eq:update-adjust}
		\end{equation}
		}
		%
		where $B>0$ is some parameter to be specified momentarily. 
		This operation prevents the ratio $\frac{\max\{ \mu_{j,+}^{t+1},\mu_{j,-}^{t+1} \} }{\min\{ \mu_{j,+}^{t+1},\mu_{j,-}^{t+1} \}}$ 
		from being exponentially large (i.e., it is no larger than $e^{B}$), a condition that helps facilitate analysis. 

%
\end{subequations}

\end{itemize}

\noindent
After running the above updates for $t_{\max}$ iterations, we reach a probability matrix taking the following form: 
\begin{align}
	\widehat{\bm{P}} = \big[r_1 \bm{p}_1^{t_{\max}}, \ldots, r_n \bm{p}_n^{t_{\max}}\big]^{\top}, \label{eq:output}
\end{align}
which can be converted into a feasible transportation plan $\widetilde{\bm{P}}$ by invoking \cite[Algorithm~2]{altschuler2017near}.
The whole procedure is summarized in Algorithm~\ref{alg:main}.

\begin{algorithm}[t]
	\DontPrintSemicolon
	\SetNoFillComment
	\vspace{-0.3ex}
	{\small
	\textbf{Input}: cost matrix $\bm{W}\in \mathbb{R}_+^{n\times n}$,  probability vectors $\bm{r}=[r_i]_{1\leq i\leq n}, \bm{c}=[c_i]_{1\leq i\leq n} \in \Delta_n$, target accuracy level $\varepsilon$, 
	number of iterations $t_{\max}$. 

	\tcp{Initialization} 
	$\mu_{j,+}^{0} = \mu_{j,-}^{0} = \mu_{j,+}^{0,\trunc} = \mu_{j,-}^{0,\trunc} = 1/2$ for all $1\leq j\leq n$;  $\bm{p}_i^{0} =[1/n,\ldots,1/n]$ for all $1\leq i\leq n$.  \label{line:initialization-main}\\
	$\bm{W}\leftarrow \bm{W}/\|\bm{W}\|_{\infty}$; $\varepsilon \leftarrow \varepsilon / \|\bm{W}\|_{\infty}$. \tcc{normalization}

	\tcp{Main loop}
	\For{$ t = 0$ \KwTo $t_{\max}-1$}{
		\For{$ j = 1$ \KwTo $n$}{
			$\overline{\mu}_{j,s}^{t+1}  \leftarrow \big(\mu_{j,s}^{t,\trunc}\big)^{1-\eta}\exp\Big(\eta_{\mu,j}s\big(\sum_{i=1}^{n}r_{i}p_{i,j}^{t}-c_{j}\big)\Big),\qquad \hfill s\in\{+,-\};$ \\
			$\overline{\bm{\mu}}_{j}^{t+1}\leftarrow \mathtt{Normalize}(\overline{\bm{\mu}}_{j}^{t+1})$. \tcc{Call Algorithm~\ref{alg:function-normalize}}
		}
		\For{$ i = 1$ \KwTo $n$}{
			$\overline{p}_{i,j}^{t+1}  \leftarrow \big(p_{i,j}^{t}\big)^{1-\eta}\exp\Big(-\eta_{p,i}r_{i}\big(0.5w_{i,j}+\mu_{j,+}^{t,\trunc}-\mu_{j,-}^{t,\trunc}\big)\Big),
			\qquad\hfill j=1,\cdots, n$; \\
			$\overline{\bm{p}}_{i}^{t+1}\leftarrow \mathtt{Normalize}(\overline{\bm{p}}_{i}^{t+1})$. \tcc{Call Algorithm~\ref{alg:function-normalize}}
		}

		\For{$ j = 1$ \KwTo $n$}{
			$\mu_{j,s}^{t+1}  \leftarrow \big(\mu_{j,s}^{t,\trunc}\big)^{1-\eta}\exp\bigg(\eta_{\mu,j}s\Big(\sum_{i=1}^{n}r_{i}\overline{p}_{i,j}^{t+1}-c_{j}\Big)\bigg),\qquad\hfill s\in\{+,-\}$; \\
			$\bm{\mu}_{j}^{t+1}\leftarrow \mathtt{Normalize}(\bm{\mu}_{j}^{t+1})$. \tcc{Call Algorithm~\ref{alg:function-normalize}}
		}
		\For{$ i = 1$ \KwTo $n$}{
			$p_{i,j}^{t+1}  \leftarrow \big(p_{i,j}^{t}\big)^{1-\eta}\exp\Big(-\eta_{p,i}r_{i}\big(0.5w_{i,j}+\overline{\mu}_{j,+}^{t+1}-\overline{\mu}_{j,-}^{t+1}\big)\Big),
			\qquad\hfill j=1,\cdots, n$; \\
			$\bm{p}_{i}^{t+1}\leftarrow \mathtt{Normalize}(\bm{p}_{i}^{t+1})$. \tcc{Call Algorithm~\ref{alg:function-normalize}}
		}

		\For{$ j = 1$ \KwTo $n$}{
			$\mu_{j,s}^{t+1,\trunc}\leftarrow \max\big\{\mu_{j,s}^{t+1},\,e^{-B}\max\big\{\mu_{j,+}^{t+1},\mu_{j,-}^{t+1} \big\}\big\},\qquad \hfill s\in\{+,-\}$; \\
			$\bm{\mu}_{j}^{t+1,\trunc}\leftarrow \mathtt{Normalize}(\bm{\mu}_{j}^{t+1,\trunc})$. \tcc{Call Algorithm~\ref{alg:function-normalize}}
		}

	}

	Set $\widehat{\bm{P}}= [r_1 \bm{p}_1^{t_{\max}}, \ldots, r_n \bm{p}_n^{t_{\max}} ]^{\top}$.  \tcc{Solution of regularized problem \eqref{eq:unconstrained}}
	\tcp{Convert an almost-transportation plan $\widehat{\bm{P}}$ to a feasible  $\widetilde{\bm{P}}$ }

	\textbf{Output }a feasible $\widetilde{\bm{P}}$, obtained by  invoking \cite[Algorithm~2]{altschuler2017near} with input $\widehat{\bm{P}}$.

	}
	\caption{Our entropy-regularized extragradient method for optimal transport.\label{alg:main}}
\end{algorithm}

\begin{algorithm}[t]
	\DontPrintSemicolon
	\SetNoFillComment
	\vspace{-0.3ex}
		 
		 {\small
	\textbf{Input:} $\bm{x}= [x_i]_{1\leq i\leq d}$

	\textbf{Output }$\bm{y}=[y_i]_{1\leq i\leq d}$, where $y_i = \frac{x_i}{\sum_j x_j}$.

	}
	\caption{$\mathtt{Normalize}(\bm{x})$.\label{alg:function-normalize}}
\end{algorithm}

\paragraph{Choice of algorithmic parameters.} 
Thus far, we have not yet discussed the choices of multiple parameters required to run Algorithm~\ref{alg:main}. 
Let us begin by looking at the regularization parameters $\{\tau_{\mu,j}\}$ and $\{\tau_{p,j}\}$, 
which cannot be taken to be too large. 
Evidently, if the regularization parameters are chosen such that
\begin{align}
	\sum\nolimits_{j = 1}^n \tau_{\mu,j}\log 2 + \sum\nolimits_{i = 1}^n \tau_{p,i}\log n \le \varepsilon, 
\end{align}
then it follows from elementary properties of the entropy that: for any $\{\bm{p}_{i}\in \Delta_n\}_{i=1}^{n}$ and $\{\bm{\mu}_{j}\in \Delta_2\}_{i=1}^{n}$,
{\small
\begin{equation}
\Big|F\big(\{\bm{p}_{i}\}_{i=1}^{n},\{\bm{\mu}_{j}\}_{i=1}^{n}\big)-f\big(\{\bm{p}_{i}\}_{i=1}^{n},\{\bm{\mu}_{j}\}_{i=1}^{n}\big)\Big|\leq\sum_{j=1}^{n}\tau_{\mu,j}\log2+\sum_{i=1}^{n}\tau_{p,i}\log n\leq\varepsilon .
	\label{eq:gap-regularized-original}
\end{equation}
}
Consequently, any $\varepsilon$-approximate solution to \eqref{eq:game-entropy} --- namely, a point whose resulting duality gap is no larger than $\varepsilon$ --- is an $2\varepsilon$-approximate solution to \eqref{eq:unconstrained}. 
Moverover, the theory developed in \cite{cen2021fast} for matrix games suggests that a feasible learning rate can be chosen to be inversely proportional to the regularization parameter. 
As a result, we take, for some quantity $\eta>0$, 
\begin{equation}
	\eta_{\mu,j} = \eta / \tau_{\mu,j} 
	\qquad \text{and} \qquad 
	\eta_{p,i} = \eta/\tau_{p,i} ,
	\qquad \forall 1\leq i,j\leq n
	\label{eqn:learning-rates-eta}
\end{equation}

With the above considerations in mind, we recommend the parameter choices: 
\begin{align}
	B = C_1\log\frac{n}{\varepsilon},\qquad
	\eta = \frac{C_2^2\varepsilon}{\sqrt{B}\log n}, \qquad
	\eta_{\mu,j} = \frac{15C_2\sqrt{B}}{c_j + C_3/n},\qquad 
	\eta_{p,i} = \frac{C_2}{\sqrt{B}r_i},
	 \label{eq:parameter}
\end{align}
for $1\leq i,j\leq n$ with $C_1>0, C_2>0, 0<C_3 \leq 1$ some suitable universal constants,  which correspond to 
\begin{align}
	\tau_{\mu,j}=\frac{\eta}{\eta_{\mu,j}}=\frac{C_{2}(c_{j} + C_3/n)\varepsilon}{15C_{1} (\log n) \big(\log\frac{n}{\varepsilon} \big)}
	\qquad\text{and}\qquad
	\tau_{p,i}=\frac{\eta}{\eta_{p,i}}=\frac{C_{2}r_{i}\varepsilon}{\log n}
	\label{eq:equiv-tau-mup}
\end{align}
for all $1\leq i,j\leq n$. Note that if $r_i=0$, then we simply take $\eta_{p,i}r_i=C_2/\sqrt{B}$, as the proposed algorithm only relies on the product $\eta_{p,i}r_i$. 
Three remarks are in order.  Firstly, the learning rate $\eta_{p,i}$ (resp.~$\eta_{\mu,j}$) is chosen adaptively to be inversely proportional to the row sum $r_i$ (resp.~column sum $c_j$), which is crucial in achieving our advertised convergence rate; in contrast, fixing the learing rates across all $i$ (resp.~$j$) as in prior works results in slow convergence particularly when the $r_i$'s (resp.~$c_j$'s) are far from uniform. 
Secondly, $\eta_{\mu,j}(c_j+O(1/n))$ is chosen to be larger than $\eta_{p,i}r_i$, 
in the hope that $\{\bm{\mu}_j^t\}$ converges more rapidly than $\{\bm{p}_i^t\}$. 
Furthermore, if $C_1$ is large enough and $C_2$  small enough,  
the above regularization parameters obey  
\begin{equation}
	\sum_{j=1}^{n}\tau_{\mu,j}\log2+\sum_{i=1}^{n}\tau_{p,i}\log n=\frac{C_{2}(1+C_3)\varepsilon\log2}{15C_{1}\log n\log\frac{n}{\varepsilon}}+C_{2}\varepsilon\leq\varepsilon / 4
	\label{eq:sum-tau-UB1}
\end{equation}
given that $\sum_j  c_j = \sum_i r_i=1$, 
thus satisfying \eqref{eq:gap-regularized-original}.

\subsection{Theoretical guarantees}
\label{sec:theory}

Our theoretical analysis delivers intriguing news about the convergence properties of the proposed algorithm, as stated below.

\begin{theorem}
\label{thm:theory-OT}
Consider any $0<\varepsilon <\|\bm{W}\|_{\infty}$. Algorithm~\ref{alg:main} with the parameters \eqref{eq:parameter} returns a probability matrix $\widetilde{\bm{P}}\in \Delta_{n\times n}$ obeying 
\begin{align}
	\widetilde{\bm{P}} \bm{1} = \bm{r}, \qquad \widetilde{\bm{P}}^{\top}\bm{1}=\bm{c}, \qquad 
	\text{and} \qquad
	\langle \bm{W}, \widetilde{\bm{P}} \rangle  \le \langle \bm{W}, \bm{P}^{\star}\rangle + \varepsilon, \label{eq:main}
\end{align}
provided that $C_1>0$ is large enough, $C_2>0$ is small enough, $C_2\sqrt{C_1}$ is large enough, $0<C_3\leq 1$, $C_2^2/C_3$ is small enough, and 
\begin{equation}
	t_{\max} \geq C_4 \frac{\|\bm{W}\|_{\infty}}{\eta}\log \Big( \frac{n\|\bm{W}\|_{\infty}}{\varepsilon} \Big)
\end{equation}
for some large enough constant $C_4>0$, with $\eta$  defined in \eqref{eq:parameter}.
\end{theorem}
\begin{remark}
For instance, our analysis reveals that this theorem holds under the following particular values of the constants: 
	$C_1 = 124, C_2 = 0.024, C_3 = 1, C_4 = 2$. Caution needs to be exercised, however, that these values are derived primarily for theoretical purposes and could be overly conservative.
\end{remark}
%

%
Assuming without loss of generality that $\|\bm{W}\|_{\infty}=1$, Theorem~\ref{thm:theory-OT} in conjunction with the choice of $\eta$ in \eqref{eq:parameter} asserts that the iteration complexity of our algorithm is 
\begin{equation}
	\text{(iteration complexity)} \qquad \widetilde{O}\big( 1 / \varepsilon
	\big).
	\label{eq:iteration-complexity-OT}
\end{equation}
Given that each iteration can be implemented in $O(n^2)$ time, 
the total computational complexity of Algorithm~\ref{alg:main} is no larger than
\begin{align}
	\text{(computation complexity)} \qquad
	\widetilde{O}\big(
		n^2 / \varepsilon
		\big).
	\label{eq:computation-complexity-OT}
\end{align}
This matches the state-of-the-art theory \cite{jambulapati2019direct} 
among all first-order methods tailored to the optimal transport problem; 
we will demonstrate the practical efficacy of our algorithm shortly.  
Regarding the memory complexity, all computation in our algorithm only involves matrices of dimension no larger than $n \times n$.  
Compared to \cite{blanchet2018towards,quanrud2018approximating,van2020bipartite}, 
our algorithm is easy-to-implement and amenable to parallelism, without the need of calling any unimplementable blackbox subroutine that is mainly of theoretical interest.

Finally, while our analysis is inspired by the prior work \cite{cen2021fast}, a direct application of their analysis framework can only lead to highly suboptimal iteration complexity. 
Particularly, they focus on the $\ell_{\infty}$ type bound, which can ensure $\|\widehat{\bm{P}}^{\top}\bm{1}-\bm{c}\|_{\infty} \le \varepsilon$ within $\widetilde{O}\big(\frac{1}{\varepsilon}\big)$ iterations.
However, we need $\|\widehat{\bm{P}}^{\top}\bm{1}-\bm{c}\|_{1} \le \varepsilon$ when paired with Lemma~\ref{lem:projection} to get our desired result, which in turn requires $\widetilde{O}\big(\frac{n^3}{\varepsilon}\big)$ computation complexity if naively adopting the $\ell_\infty$ type bound in \cite{cen2021fast}.
Novel algorithmic and analysis ideas (e.g., how to exploit the use of adaptive learning rates) tailored to the optimal transport problem play a central role in establishing the desired performance guarantees.

\section{Related works}
\label{sec:related-work}

Let us discuss a broader set of related past works.  

\paragraph{Entropy regularization.} 
The advantages of entropy regularization have been exploited in a diverse array of optimization problems over probability distributions, 
with prominent examples including equilibrium computation in game theory \cite{ao2022asynchronous,mckelvey1995quantal,savas2019entropy,mertikopoulos2016learning,cen2021fast,cen2022independent} and policy optimization in reinforcement learning \cite{geist2019theory,neu2017unified,mei2020global,cen2022faster,cen2022fast,lan2022policy,zhan2021policy}. 
The idea of employing entropy regularization to speed up convergence in optimal transport 
has been studied for multiple decades (e.g., \cite{knight2008sinkhorn,kalantari2008complexity,chakrabarty2021better,altschuler2022flows}) and recently popularized by \cite{cuturi2013sinkhorn}. 
By adding a reasonably small entropy penalty term (so that it does not bias the objective function by much), 
the optimal solution to the entropy-regularized problem exhibits a special form $\bm{D}_r\exp(-\eta\bm{W})\bm{D}_c$, where $\bm{D}_r$ and $\bm{D}_c$ are certain diagonal matrices and the $\exp(\cdot)$ operator is applied in an entrywise manner \cite{sinkhorn1967diagonal}. 
This special structure motivates one to alternate between row and column rescaling until convergence, the key idea behind the Sinkhorn algorithm. 
Note that the use of entropy regularization 
links optimal transport with another fundamental problem called  matrix scaling \cite{idel2016review,allen2017much}; 
see \cite{altschuler2022flows} for an exposition of such connections.

\paragraph{Extragradient methods.} Dating back to \cite{korpelevich1976extragradient,tseng1995linear}, extensive research efforts have been put forth towards understanding extragradient methods for saddle-point optimization, where a clever step of extrapolation is leveraged to accelerate convergence; partial examples include the optimistic gradient descent ascent (OGDA) method \cite{mertikopoulos2018cycles,mertikopoulos2018optimistic,rakhlin2013optimization,wei2020linear},  the implicit update method \cite{liang2019interaction}, and their stochastic variants \cite{hsieh2019convergence}. \cite{mokhtari2020convergence} analyzed the convergence of extragradient methods for unconstrained smooth convex-concave saddle-point problems under the Euclidean metric, with \cite{wei2020linear} focusing on constrained saddle-point problems. While earlier works analyzed primarily average-iterate or ergodic convergence \cite{nemirovski2004prox}, significant emphasis was put on achieving last-iterate convergence motivated by machine learning applications \cite{mertikopoulos2018cycles,wei2020linear}. By using entropy regularization, \cite{cen2021fast} demonstrated fast last-iterate convergence of extragradient methods for matrix games, under weaker assumptions than those needed for solving the unregularized games directly \cite{daskalakis2018last}.

 \paragraph{Prior algorithms for the optimal transport problem.} 
Earlier effort towards computing the optimal transport include the development of the Hungarian algorithm, 
which is not a linear-time algorithm due to its complexity $\widetilde{O}(n^3)$ \cite{kuhn1956variants,munkres1957algorithms};  
this algorithm has recently been revisited by \cite{xie2022solving}, which further came up with a variant that runs faster in a special class of problem instances. 
In comparison, Sinkhorn iteration and its variants have achieved widespread adoption in practice since \cite{cuturi2013sinkhorn}.  
\cite{altschuler2017near} developed the first theory uncovering the linear-time feature of Sinkhorn iteration with computational complexity $\widetilde{O}(n^2/\varepsilon^3)$, 
and inspired a recent strand of works (e.g., \cite{dvurechensky2018computational,feydy2019interpolating}) that strengthened the runtime for Sinkhorn-type algorithms  to $\widetilde{O}(n^2/\varepsilon^2)$
(including a fast greedy variant called the Greenkhorn algorithm \cite{altschuler2017near,lin2022efficiency}). 
Additionally, Sinkhorn and Greenkhorn iterations have been shown to converge linearly 
to the solution of the entropy-regularized optimization transport problem \cite{carlier2022linear,kostic2022batch}, 
although their iteration complexity still depend upon the regularization parameter (which needs to be chosen based on $\varepsilon$). Another recent work \cite{ghosal2022convergence} derived sublinear convergence rates for Sinkhorn, where the iteration complexity depends polynomially in the regularization parameter.

First-order methods and their stochastic variants have received much recent attention, 
including but not limited to accelerated gradient descent \cite{dvurechensky2018computational}, stochastic gradient descent (SGD) \cite{genevay2016stochastic}, 
Nesterov's smoothing \cite{an2022efficient}, 
and accelerated primal-dual methods \cite{lin2022efficiency,xie2022accelerated,chambolle2022accelerated}.   
The convergence guarantees of these algorithms remain suboptimal in terms of the dependency on either $n$ or $\varepsilon$. 
As mentioned previously, the algorithms designed by \cite{blanchet2018towards,quanrud2018approximating}, while achieving an appealing $\widetilde{O}(n^2/\varepsilon)$ runtime, 
 rely heavily on reduction to blackbox methods developed in theoretical computer science (e.g., positive linear programming \cite{allen2015nearly}), 
which hinder practical realization and do not yet admit fast parallelization. 
Inspired by \cite{nesterov2007dual,sherman2017area}, \cite{jambulapati2019direct} leveraged the concept of area convexity when designing and analyzing the dual extrapolation algorithm, 
which has become the best-performing (in theory) first-order method in the previous literature. 
It is also worth noting that \cite{jambulapati2019direct} also reformulated the problem into a minimax form, 
albeit using different constraint sets (for instance, the decision variables therein are not all probability vectors). 
Another recent work \cite{mai2022a} proposed an algorithm based on the Douglas-Rachford splitting, 
which enjoys competitive numerical performance compared with Sinkhorn and is amenable to GPU acceleration.  
While this algorithm has been shown to enjoy  an iteration complexity of $O\big( 1/(\rho\varepsilon ) \big)$ (with $\rho$ some suitable penalty parameter) 
and per-iteration cost $O(n^2)$, the penalty parameter $\rho$ is taken to be on the order of $1/n$ (see \cite[Section~4]{mai2022a}), 
resulting in a total computational complexity far exceeding $n^2/\varepsilon$.   
Another recent work \cite{lahn2019graph} tackled this problem via combinatorial algorithms, yielding a runtime as fast as $\widetilde{O}(n^2/\varepsilon + n/\varepsilon^2)$.




%

 


\section{Numerical experiments}
\label{sec:experiments}

This section provides empirical results that validate our theoretical studies and confirm the efficacy of 
the proposed method.
In particular, we compare the numerical performance of Algorithm~\ref{alg:main} with the classical Sinkhorn method, its greedy variant called Greenkhorn \cite{altschuler2017near}, the recently proposed dual extrapolation method \cite[Algorithm~3]{jambulapati2019direct} and the DROT method \cite[Algorithm 1]{mai2022a}, as well as cost-free acceleration strategies like the overrelaxation variant of Sinkhorn \cite{Lehmann2021note} and the batched version of Greenkhorn \cite{kostic2022batch};
in addition, we validate the theoretical runtime $\widetilde{O}(n^2 / \varepsilon)$ for achieving an $\varepsilon$-accurate solution.
All algorithms are implemented and tested in MATLAB R2020a on an iMac with a 3 GHz 6-Core Intel Core i5 processor and 24 GB memory\footnote{Implementation with hardware acceleration (e.g.~using GPUs), like the one in \cite{mai2022a}, is beyond the scope of our current paper.}.
In our experiments, each problem instance is generated in one of the following three ways.
\begin{itemize}
    \item[(i)] {\em ``Synthetic''.}
    We first produce two $m \times m$ images as follows: each image has a randomly placed square foreground that accounts for $50\%$ of the pixels; the foreground and background have pixel values uniformly sampled from $[0, 10]$ and $[0,1]$, respectively. 
    With two images in place, we flatten and normalize them to obtain $\bm{r}, \bm{c} \in \real^n$, where $n = m^2$. 
    In addition, we let each entry of the cost matrix $\bm{W} \in \real^{n \times n}$ be the $\ell_1$ distance between each pair of pixels in an $m \times m$ image.
    \item[(ii)] {\em ``MNIST''.}
    First, a pair of images are randomly selected from the MNIST dataset\footnote{\url{http://yann.lecun.com/exdb/mnist/}}, and downsampled to size $m \times m$; then, a value of $0.01$ is added to all pixels. 
    The remaining steps for obtaining $\bm{r}, \bm{c}$ and $\bm{W}$ are the same as the ``Synthetic'' setting.
    \item[(iii)] {\em ``Point clouds''.} 
    We first generate a pair of point clouds, each with $n$ points randomly sampled from a 2-dimensional Gaussian distribution.
    Then, the marginal distributions $\bm{r}$ (resp.~$\bm{c}$) stands for the uniform distribution supported on the first (resp.~second) point cloud, i.e.~an average of Dirac distributions; moreover, each entry of  $\bm{W}$ is defined by the Euclidean distance between two points.
\end{itemize}
For each optimal transport instance, we also assign a target accuracy level $\varepsilon$, which will be employed to set parameters of the algorithms.

\subsection{Comparisons with Sinkhorn and Greenkhorn}

To demonstrate the practical applicability of the proposed algorithm, 
we first compare its empirical performance with that of the Sinkhorn algorithm and its greedy variant called Greenkhorn, which are still among the most widely used baselines for solving optimal transport.

\paragraph{Setup} 

Each algorithm is implemented with varying parameters. 
For the Sinkhorn and Greenkhorn algorithms, recall from \cite{altschuler2017near} that $\eta^{-1}$ is the strength of entropic regularization; 
in our experiments, we consider the theoretical choice $\eta = 4\varepsilon^{-1} \log n  $, as well as less conservative options $\eta \in \{10, 100, 500\}$.
Regarding the proposed extragradient method, recall that Algorithm~\ref{alg:main} requires parameters $B, \eta, \{\eta_{p,i}\}, \{\eta_{\mu,j}\}$.
We let $\eta_{p,i} = C / (\sqrt{B} r_i), \eta_{\mu, j} = C \sqrt{B} / (c_j + C_3 / n)$, and consider two options:
(1)~the theoretical choice according to~(\ref{eq:parameter}), with $B=\log(n/\varepsilon), \eta=\varepsilon / (\sqrt{B} \log n)$ and $C=1, C_3=1$; 
(2)~a fine-tuned option, with $B=1, \eta=0, C=1, C_3 = 10^{-2}$.

\paragraph{Convergence of algorithms}

\begin{figure}[tbp]
    \centering
    \includegraphics[width=0.33\textwidth]{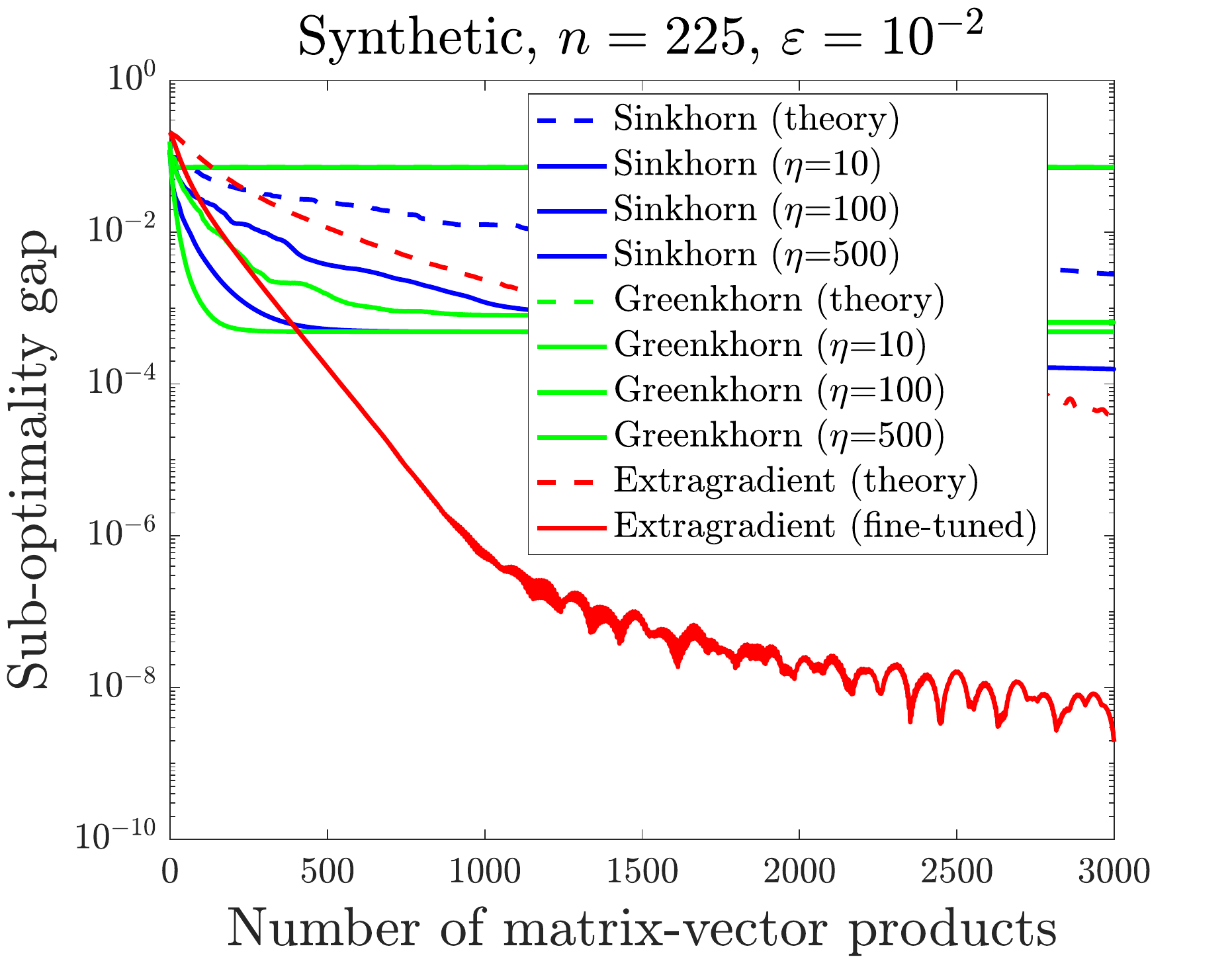}%
    \includegraphics[width=0.33\textwidth]{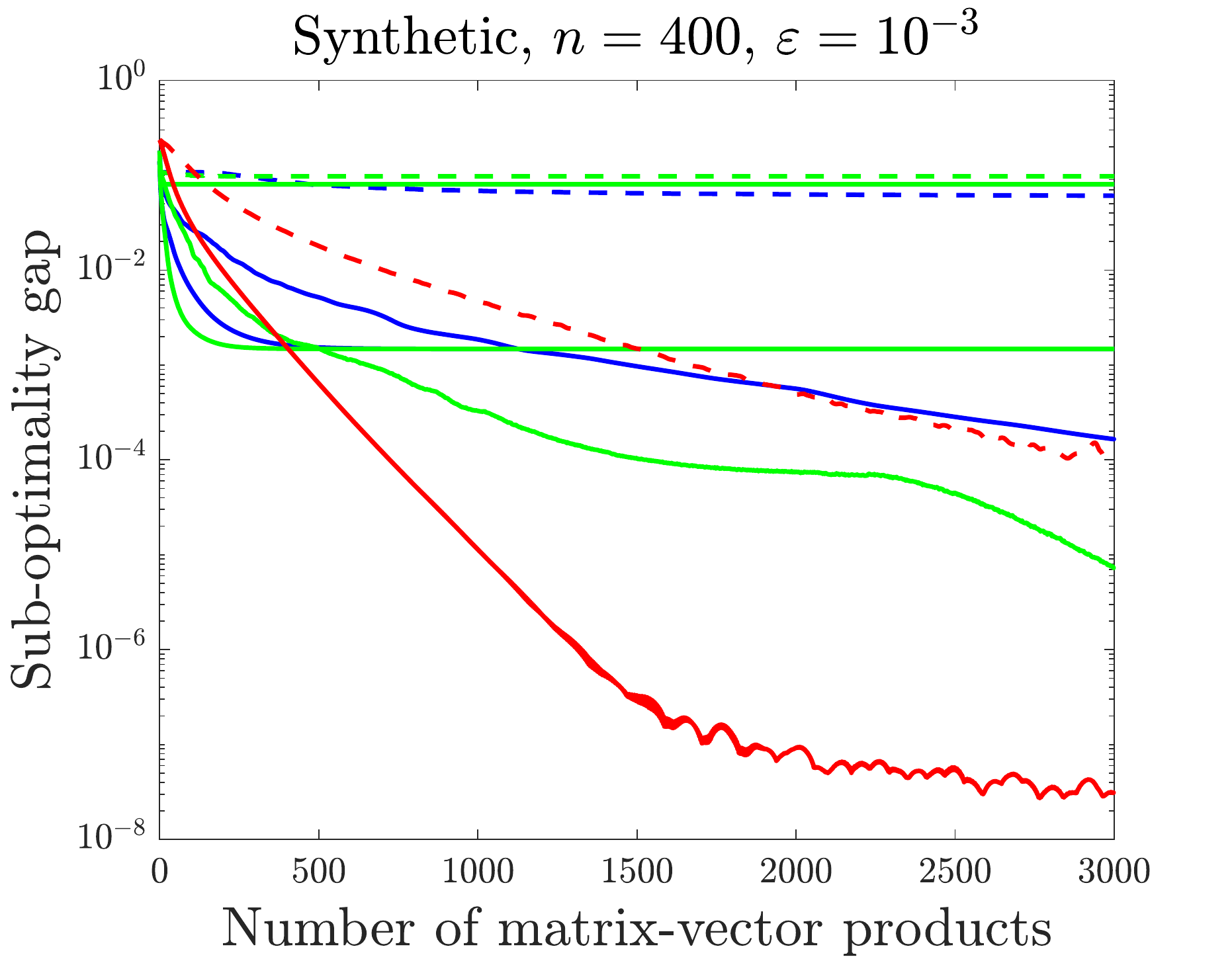}%
    \includegraphics[width=0.33\textwidth]{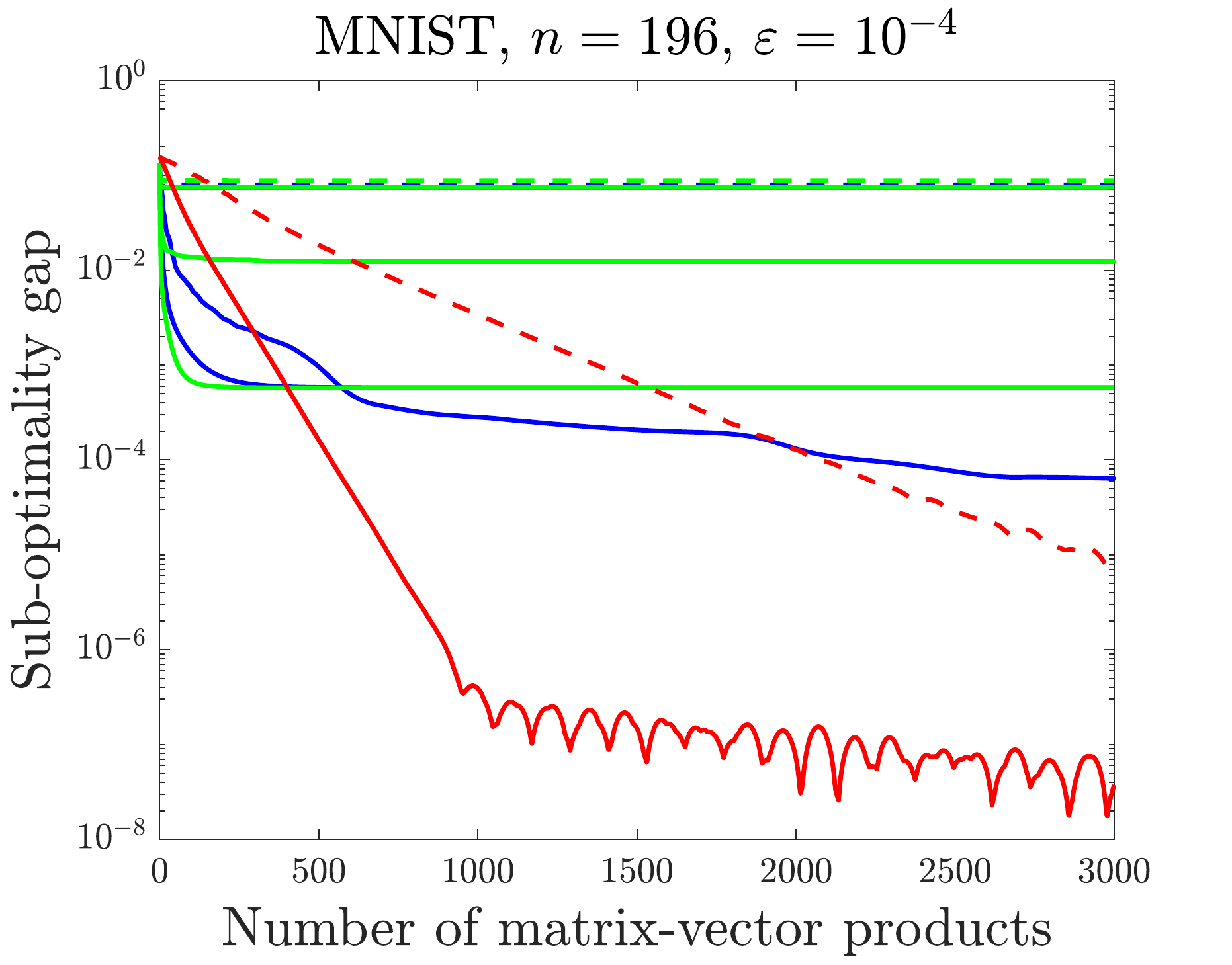}
    \includegraphics[width=0.33\textwidth]{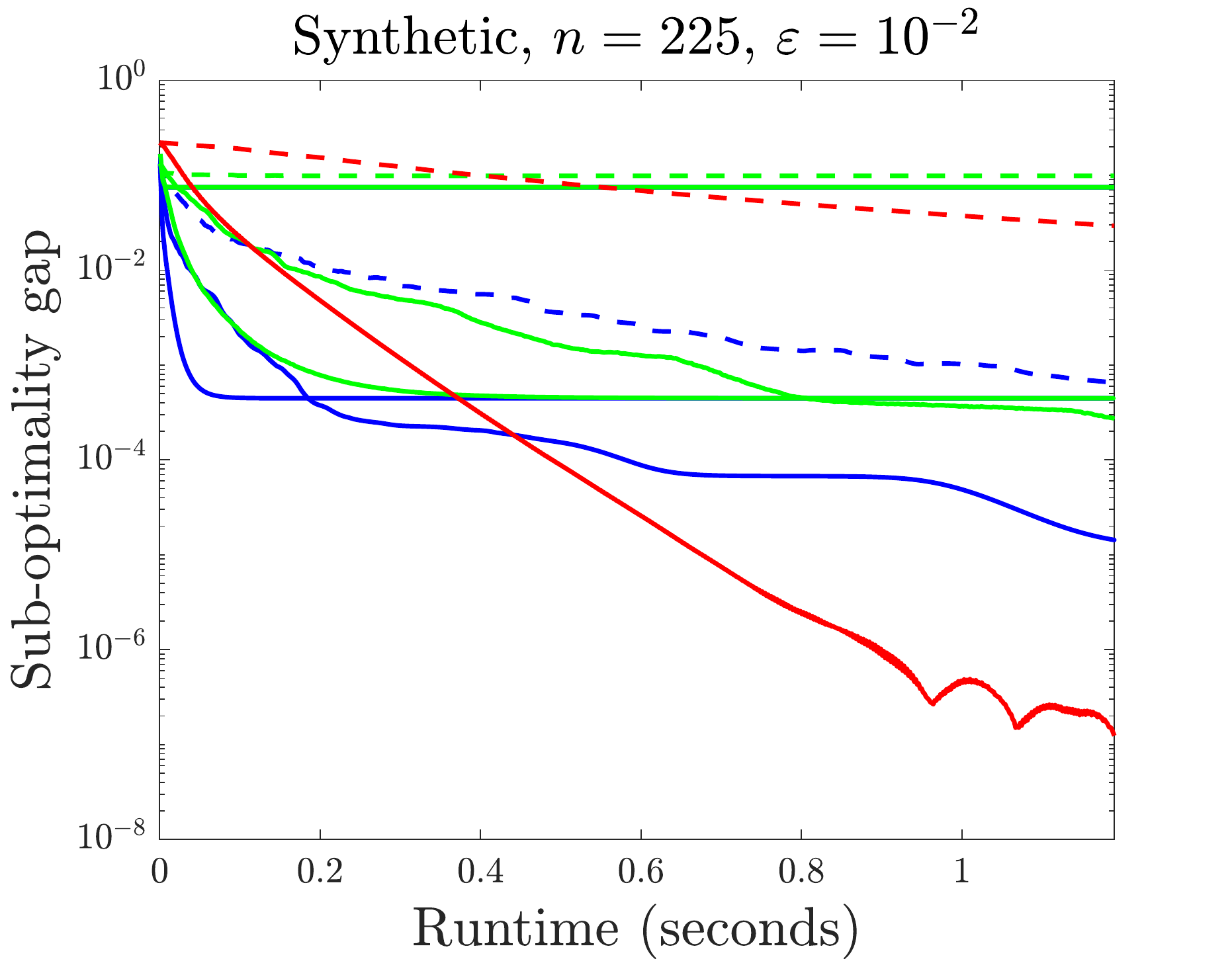}%
    \includegraphics[width=0.33\textwidth]{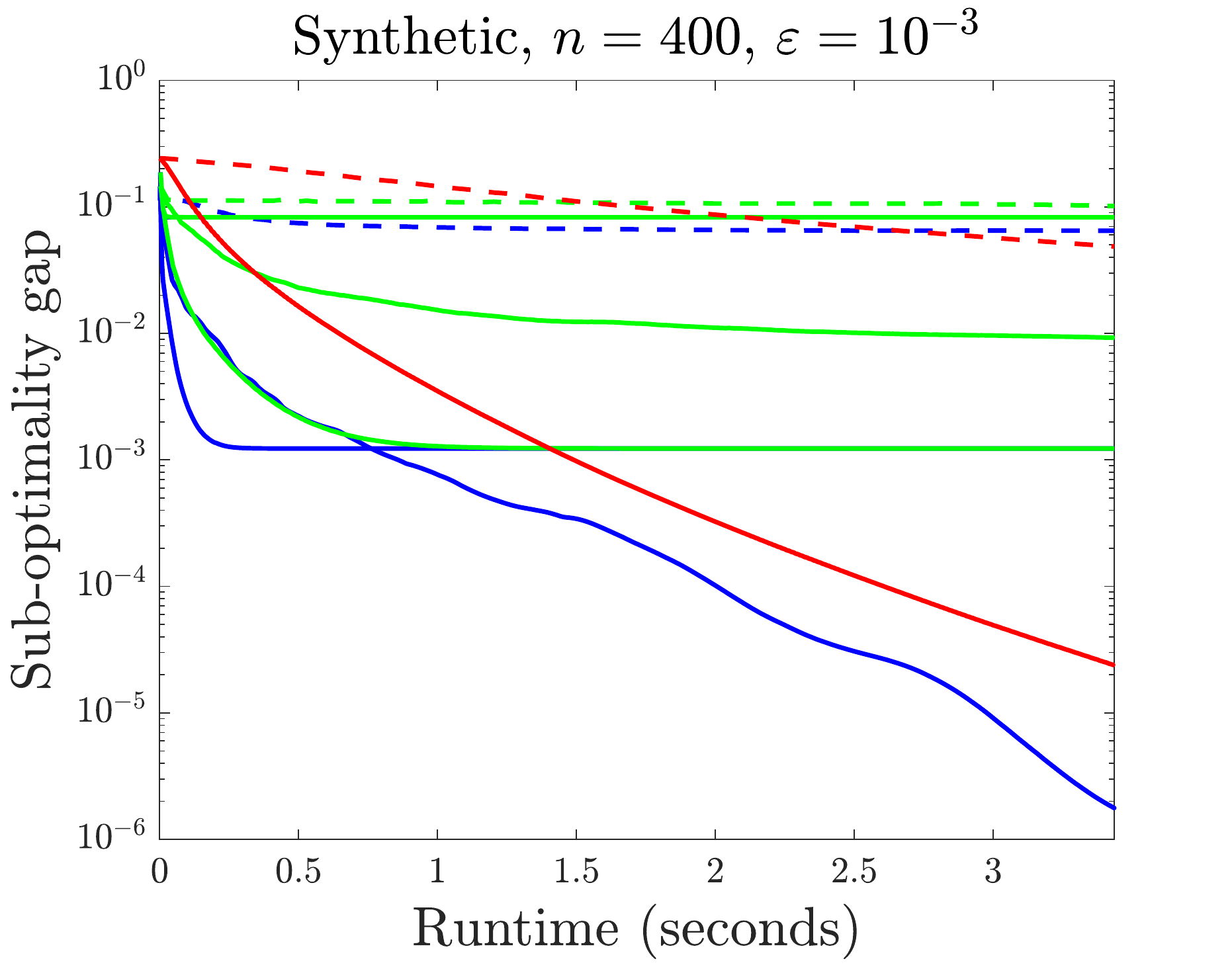}%
    \includegraphics[width=0.33\textwidth]{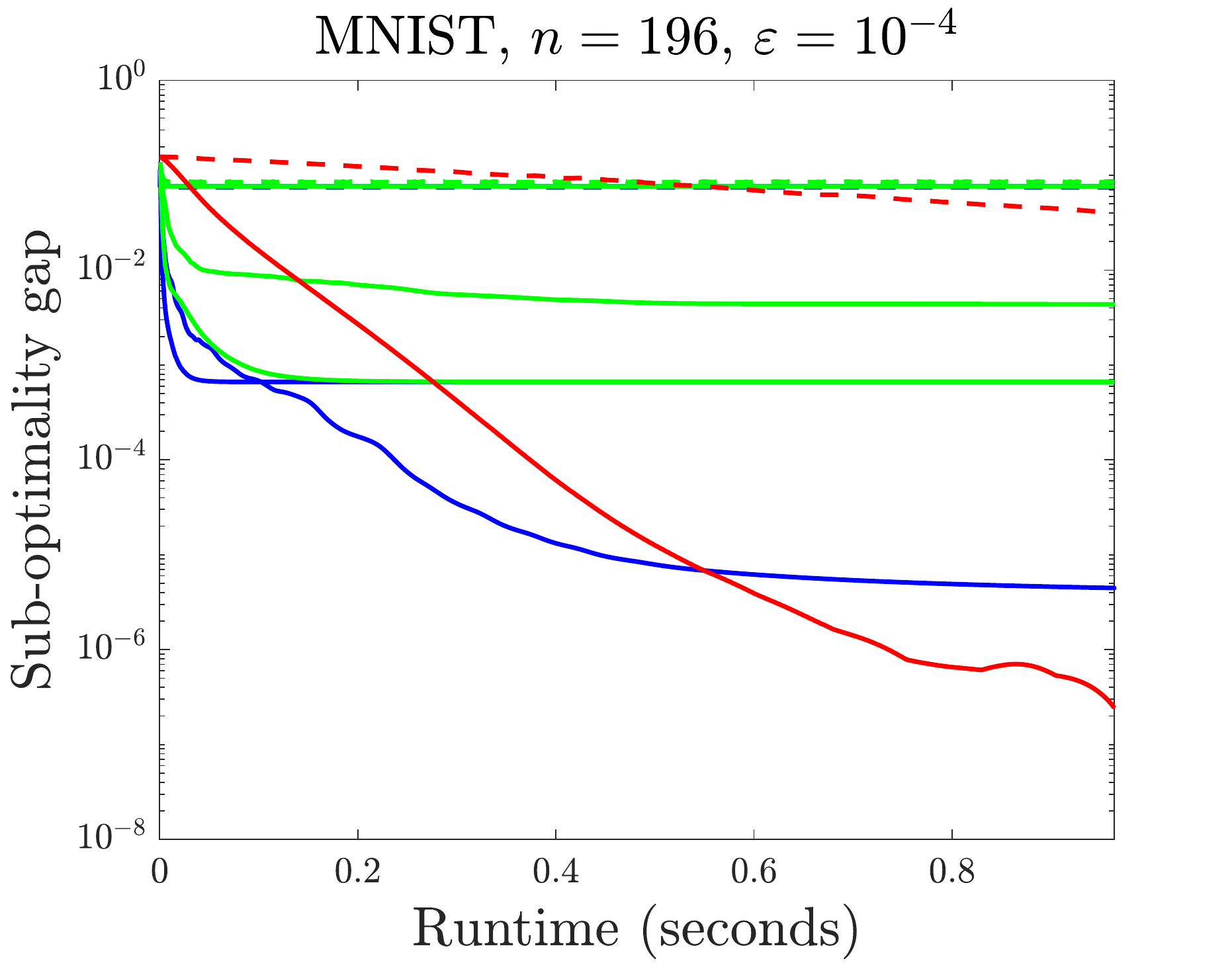}
    \includegraphics[width=0.33\textwidth]{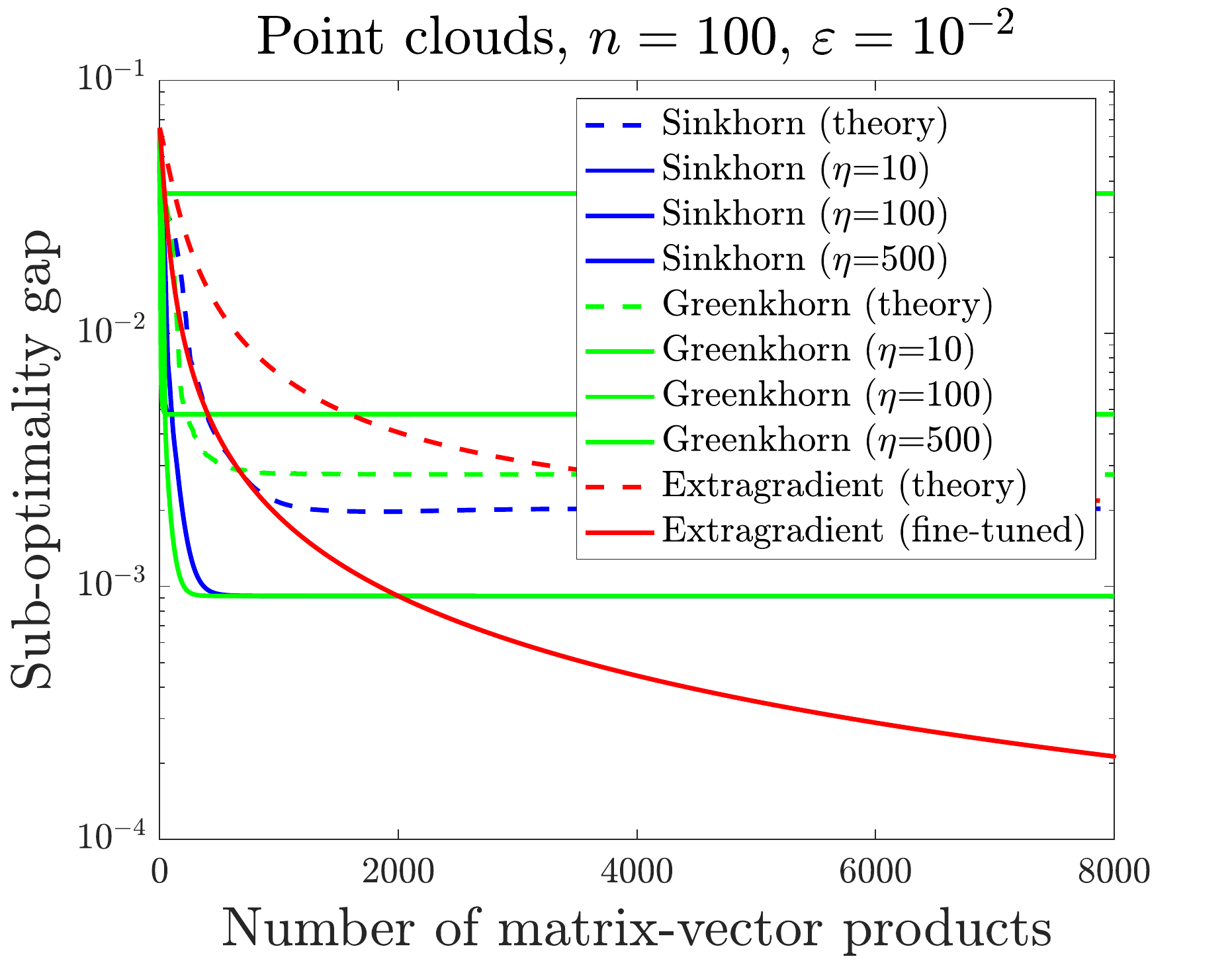}%
    \includegraphics[width=0.33\textwidth]{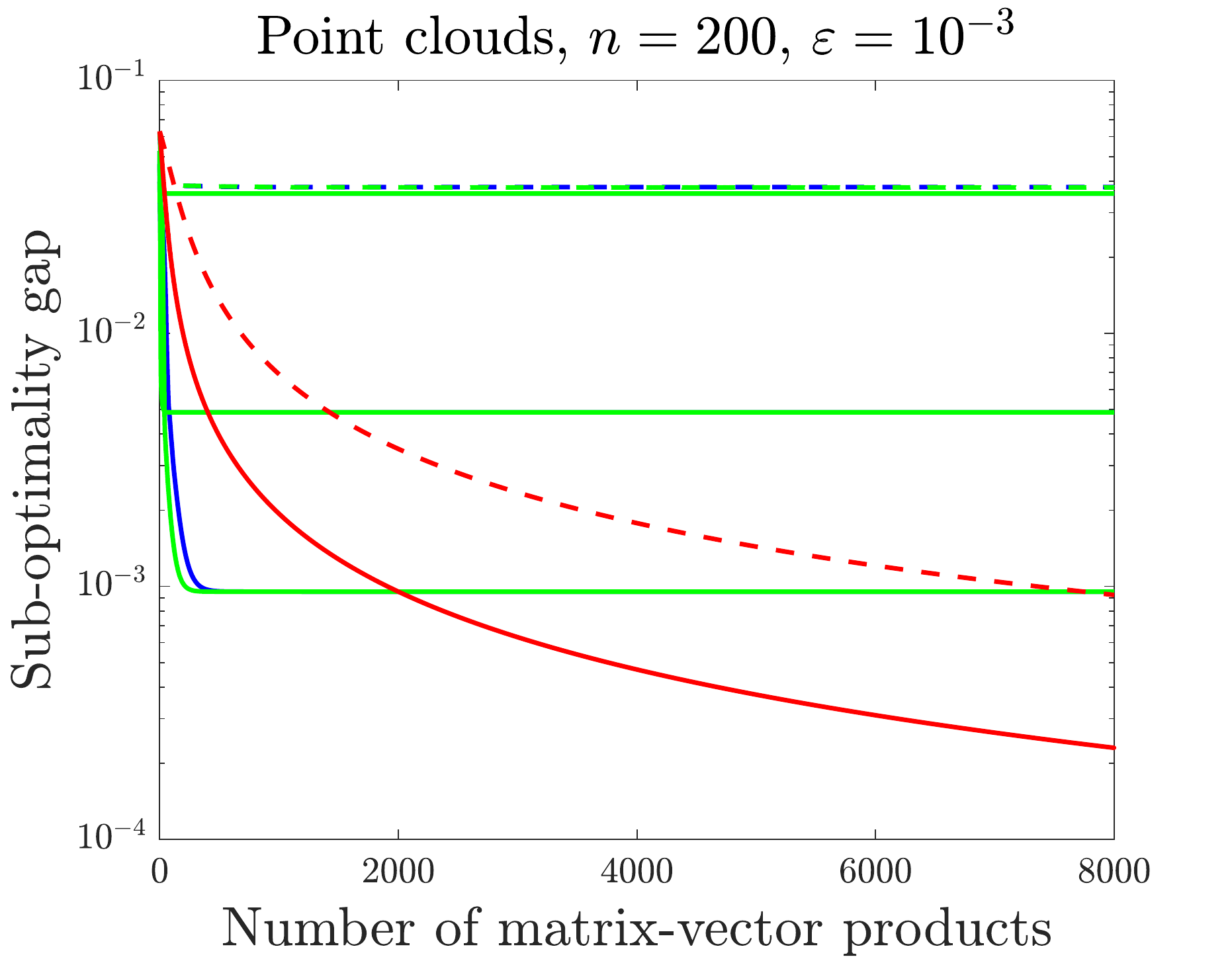}%
    \includegraphics[width=0.33\textwidth]{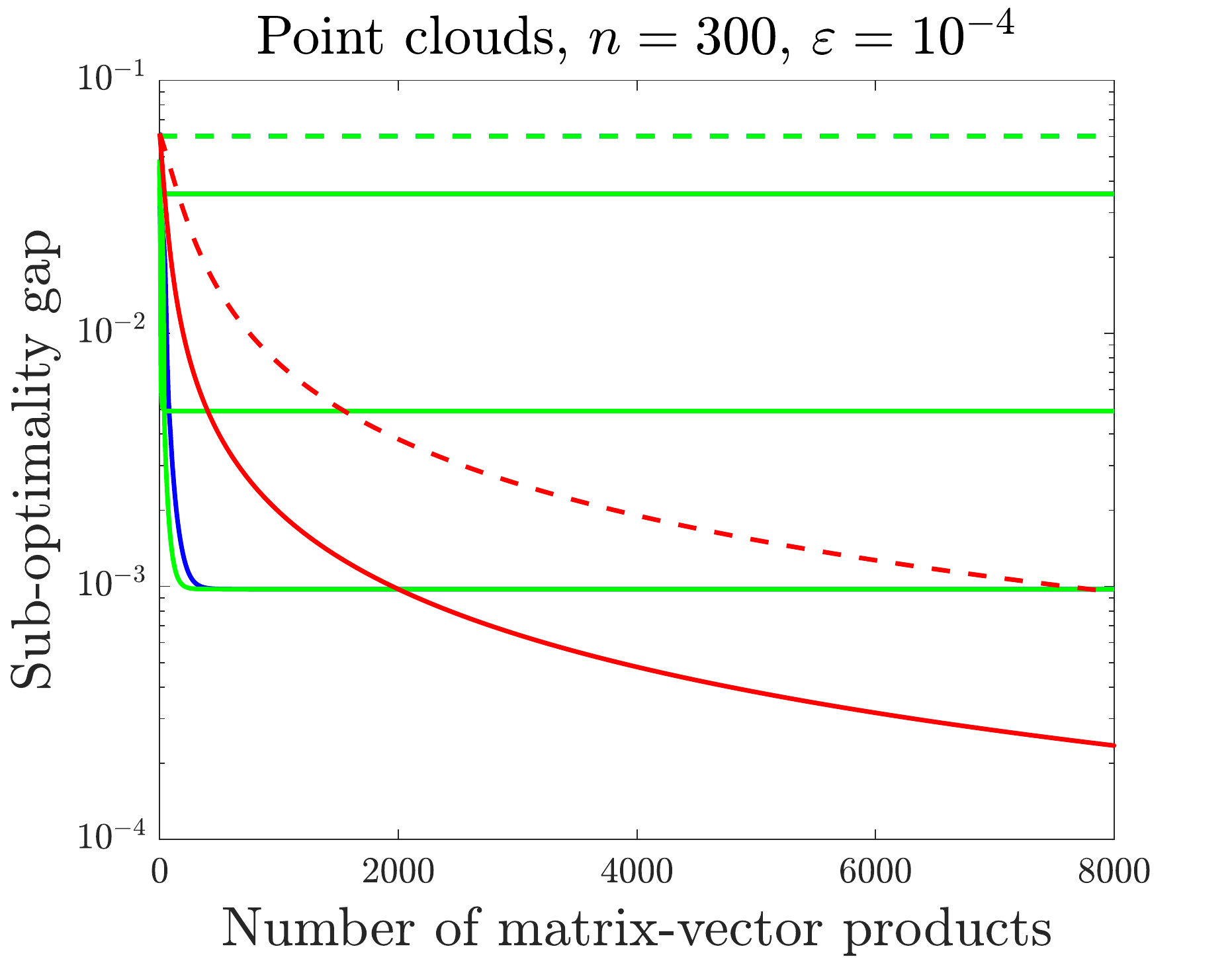}
    \includegraphics[width=0.33\textwidth]{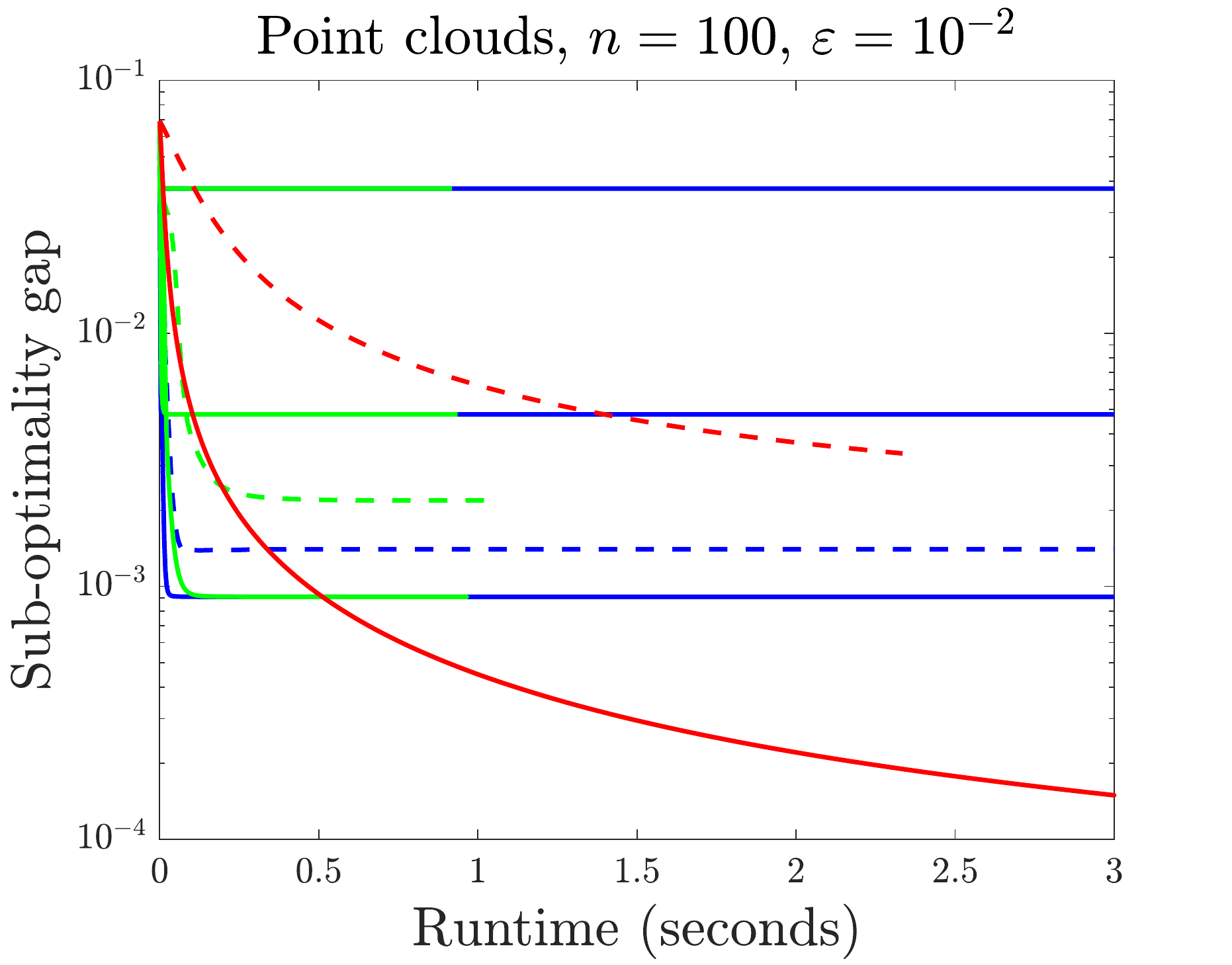}%
    \includegraphics[width=0.33\textwidth]{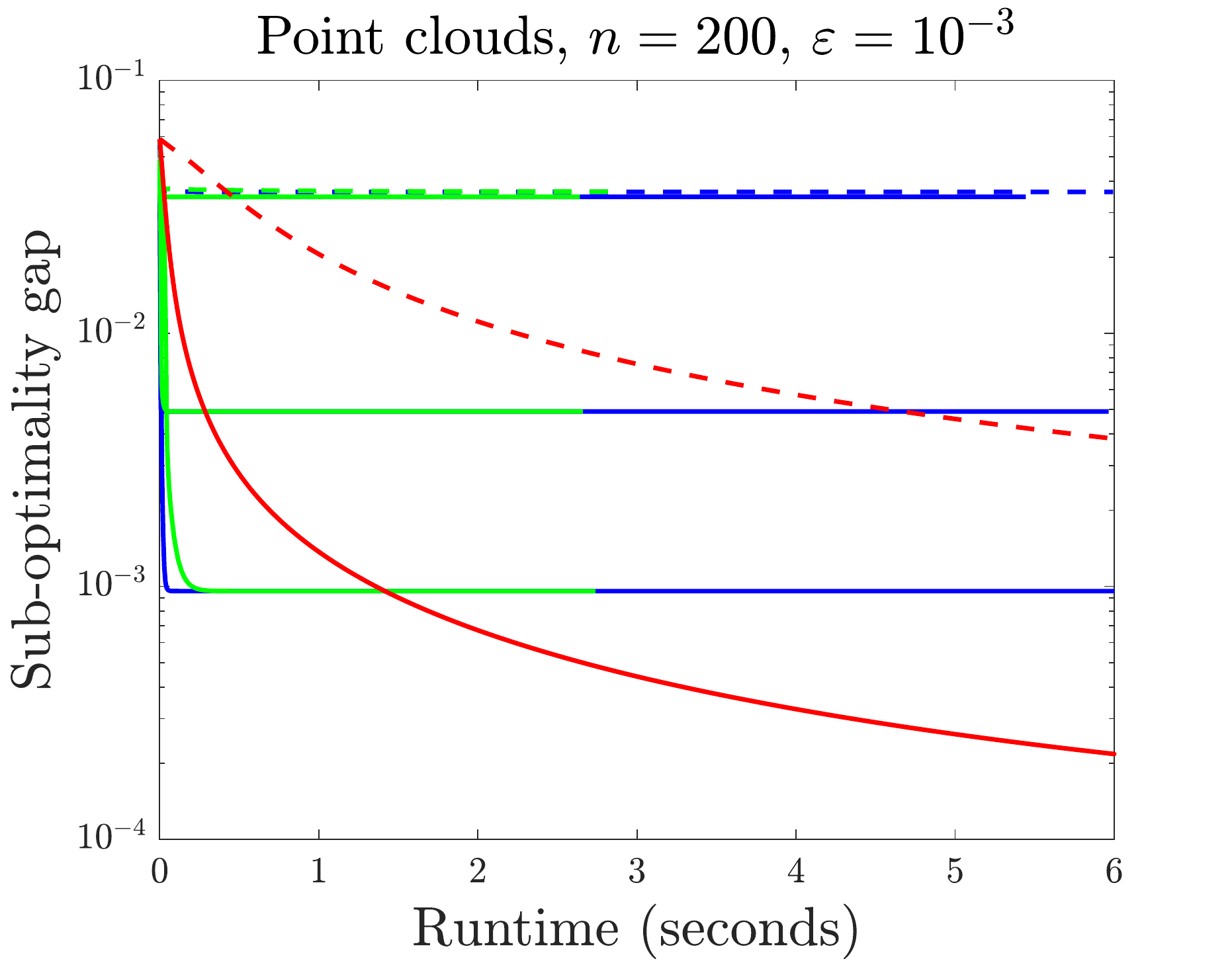}%
    \includegraphics[width=0.33\textwidth]{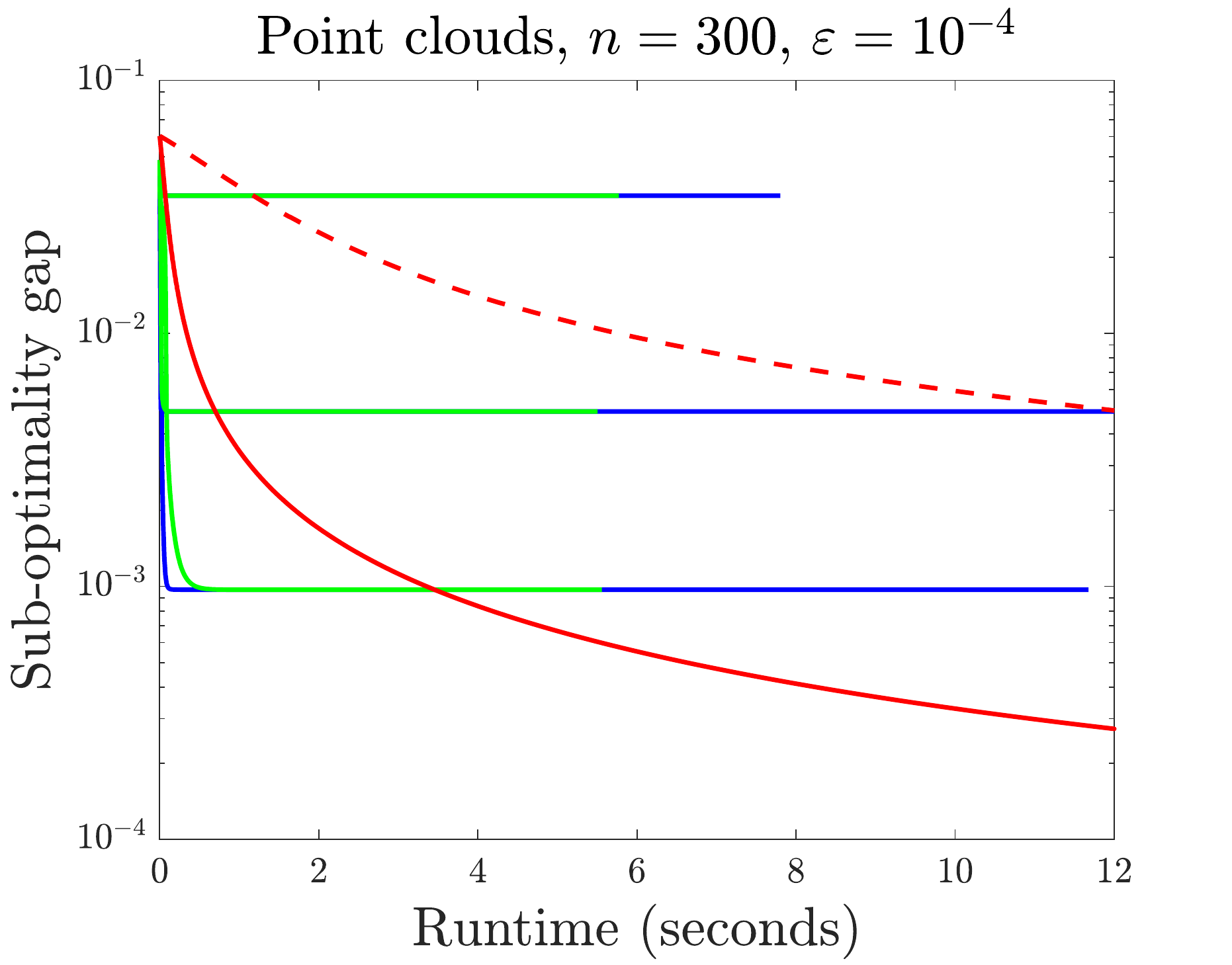}
    \caption{Empirical comparisons of various algorithms under different settings. Each curve is an average over 10 independent trials. The first and third rows use the number of matrix-vector products as a metric of computational complexities, while the second and fourth use the actual runtime. Here, we take $C_1=C_2=1$, and the Y-axis represents the sub-optimality gap $\langle \bm{W}, \bm{P} \rangle - \langle \bm{W}, \bm{P}^{\star} \rangle$.}
    \label{fig:compare_alg}
\end{figure}

The numerical results for various settings are illustrated in Figure~\ref{fig:compare_alg}.
Each subfigure represents one specific setting of optimal transport, and each curve is an average over multiple independent trials under that setting.
The $X$-axis reflects the computation cost, measured either by the total number of matrix-vector products\footnote{The number of matrix-vector products for each iteration of Sinkhorn and extragradient method is 1 and 2, respectively; for Greenkhorn, we set this value to $1/n$, based on the ratio between the numbers of row/column updates per iteration for Sinkhorn and Greenkhorn \cite{altschuler2017near}.} (akin to \cite{jambulapati2019direct}), or the actual runtime; 
the $Y$-axis reports the gap between the cost of the current iterate (rounded to the probability simplex with marginals $\bm{r}$ and $\bm{c}$) and the true optimal transport value (computed via linear programming).
The curves w.r.t.~Sinkhorn, Greenkhorn and the extragradient method are plotted in blue, green and red, respectively.  
The dashed lines stand for the theoretical choices of parameters, while solid lines correspond to the practical choices.

As illustrated in Figure~\ref{fig:compare_alg} for multiple settings, the proposed extragradient method (with fine-tuned parameters) compares favorably to both Sinkhorn and Greenkhorn, especially when the target accuracy level $\varepsilon$ is small.
The numerical results also hint at practical choices of algorithmic parameters.
\begin{itemize}
\item For example, in the presence of a small $\varepsilon$, the theoretical choice of $\eta = 4\varepsilon^{-1} \log n $ for Sinkhorn and Greenkhorn barely works in practice, since it causes numerical instability in computing $\exp(- \eta \bm{W})$ (which is a commonly known issue, cf.~\cite[Chapter~4]{peyre2019computational});
in addition, the solid lines show that within a reasonable range, reducing regularization tends to result in slower convergence but also a smaller error floor after convergence, just as expected.
\item With regards to our extragradient method, the theoretical choices of parameters also tend to be too conservative, while the fine-tuned parameters work substantially better, with the aggressive choices of  $\eta=0$ (i.e., no entropic regularization), large $C$ and small $C_3$ for the stepsizes $\{\eta_{p,i}, \eta_{\mu,j}\}$, and $B = 1$ for strong adjustment of the $\bm{\mu}$ sequences. 
Although these fine-tuned parameters were chosen based on early experiments in one or two settings, it turns out that they work uniformly well under various settings; therefore, we will mostly focus on these fine-tuned parameters in the remaining numerical results.
The superior performance under such choices might merit further theoretical studies. 
\end{itemize}
Moreover, the careful reader might remark that the numerical curves of the proposed extragradient methods exhibit non-monotonicity; 
such an oscillation behavior is common in extragradient-type methods, as they are, in general, not descent methods.

\begin{figure}[tbp]
    \centering
    \includegraphics[width=0.33\textwidth]{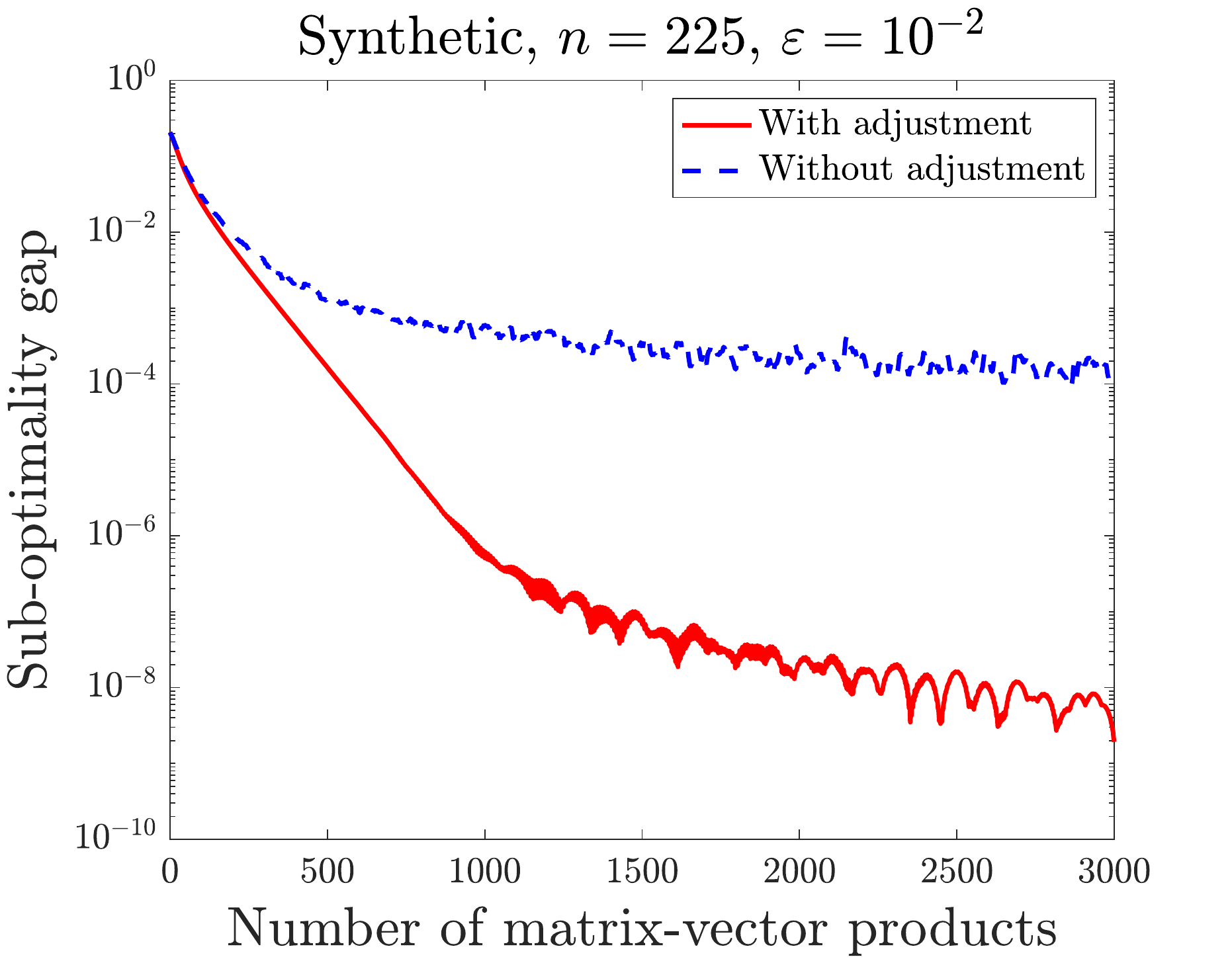}%
    \includegraphics[width=0.33\textwidth]{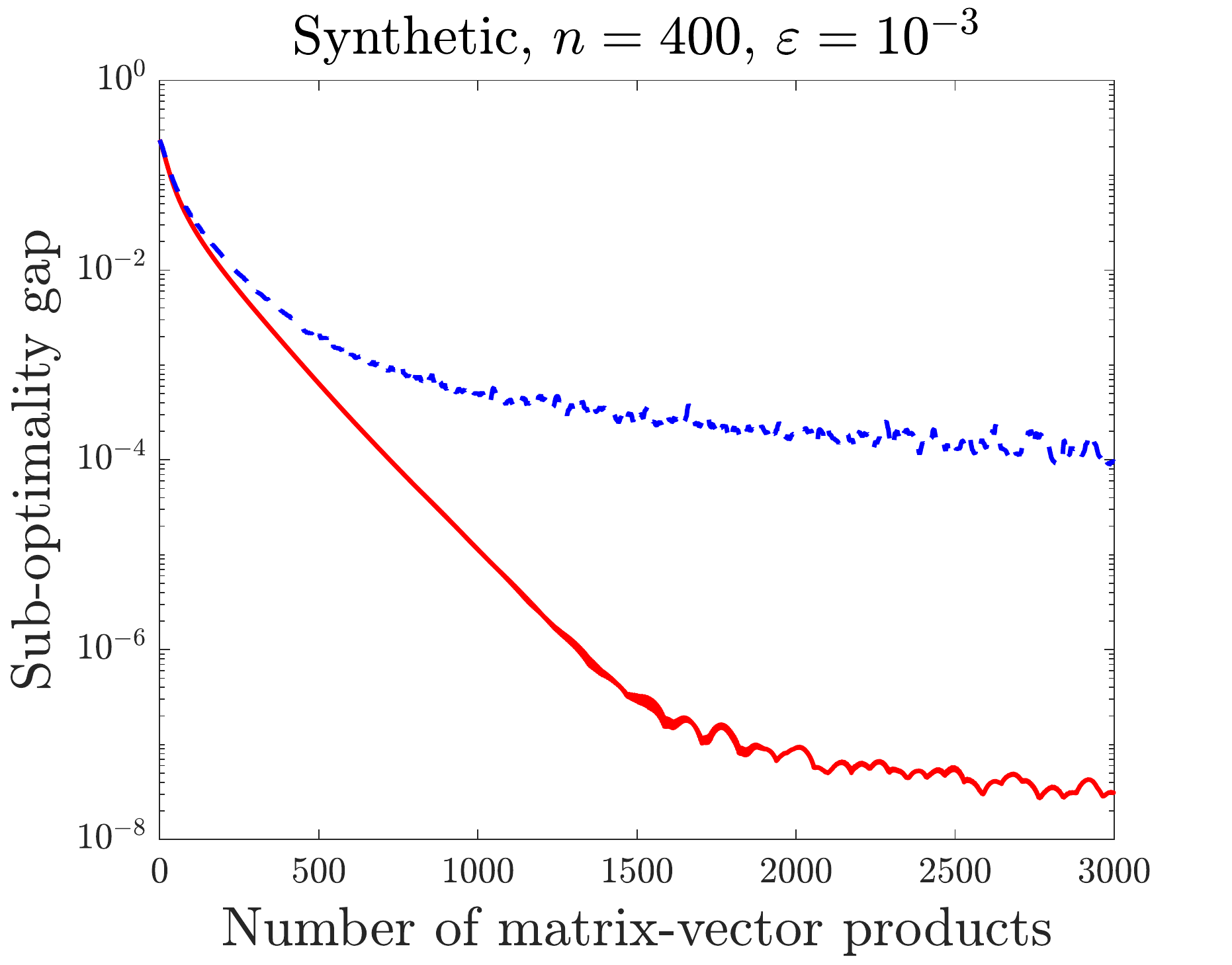}%
    \includegraphics[width=0.33\textwidth]{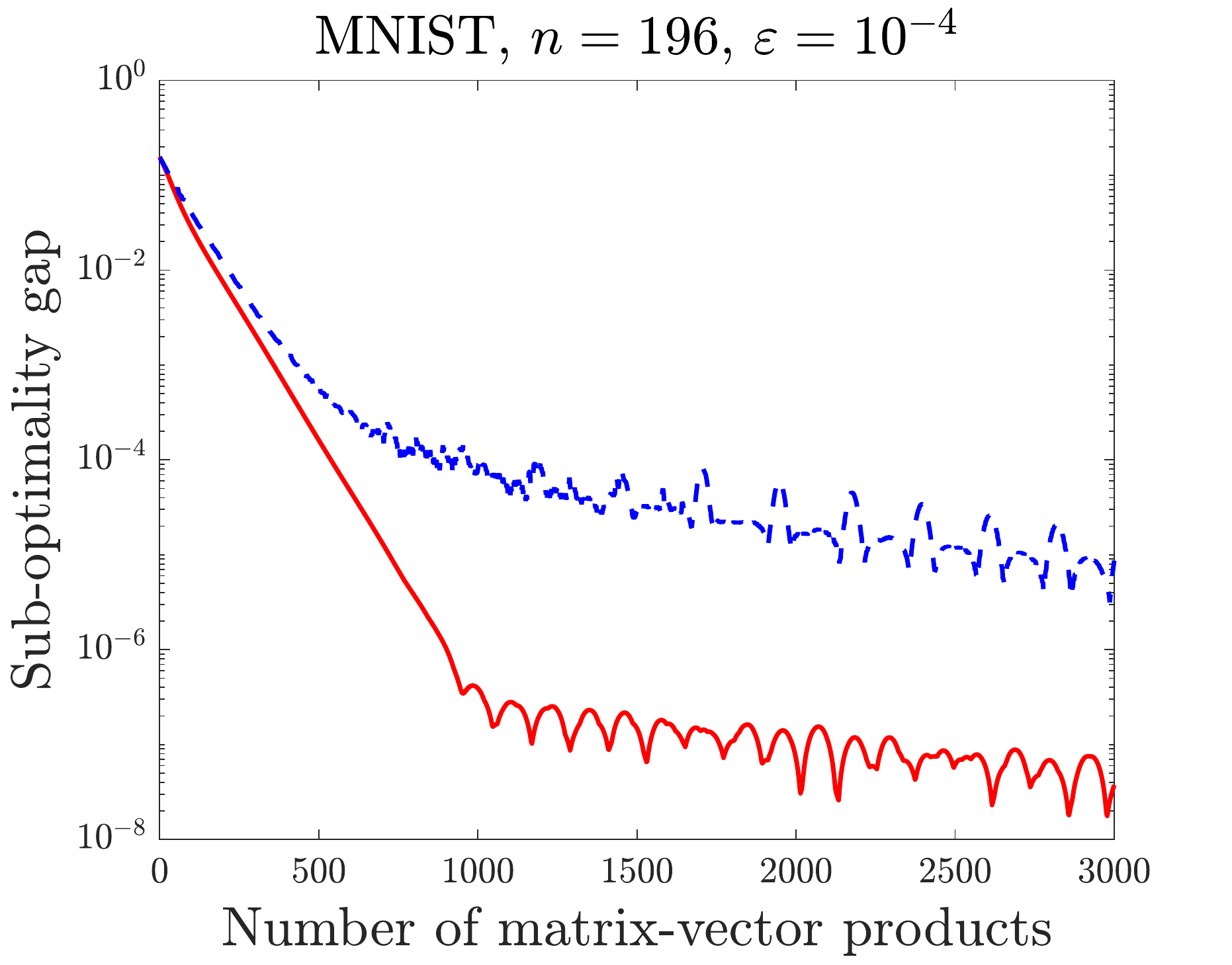}
    \includegraphics[width=0.33\textwidth]{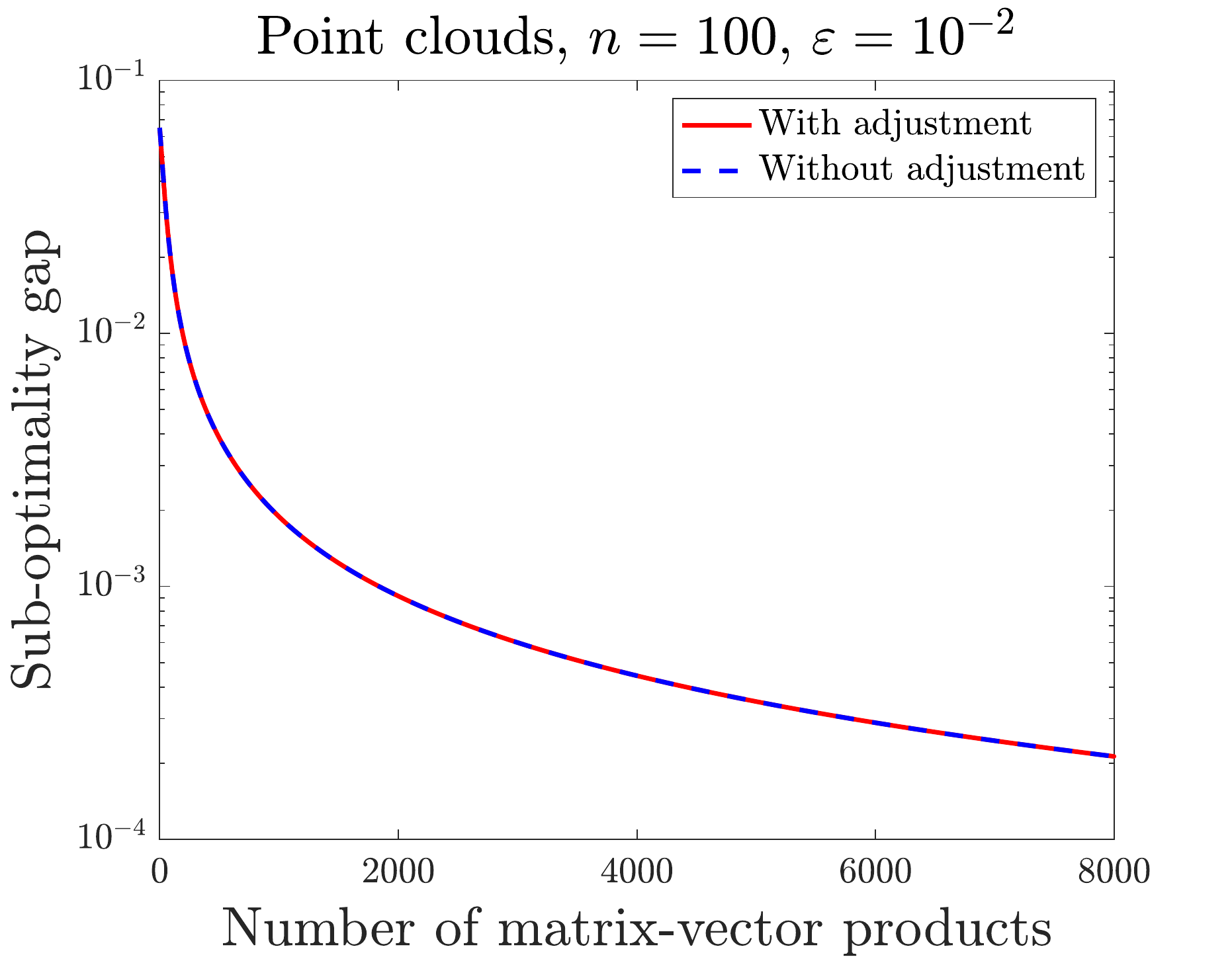}%
    \includegraphics[width=0.33\textwidth]{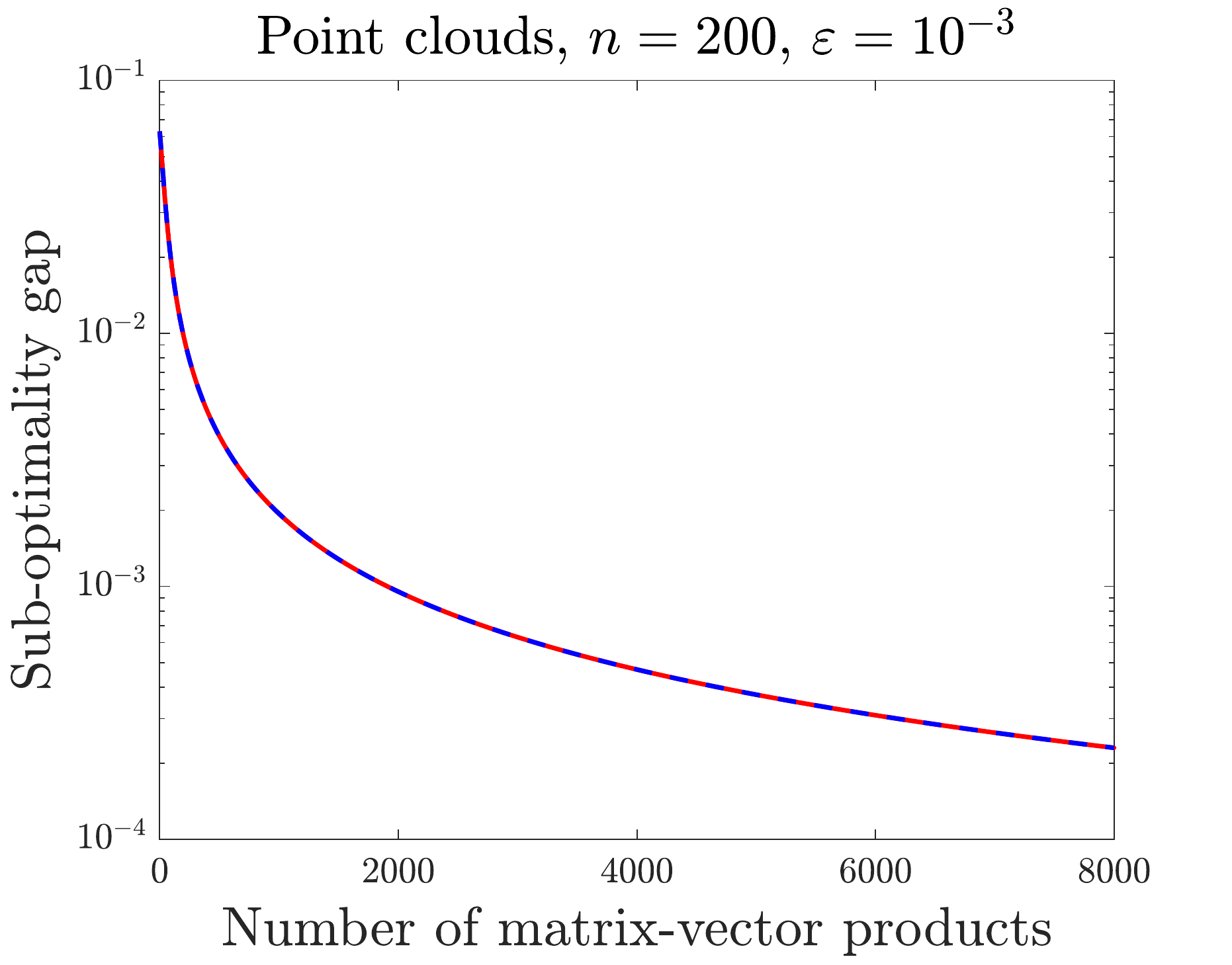}%
    \includegraphics[width=0.33\textwidth]{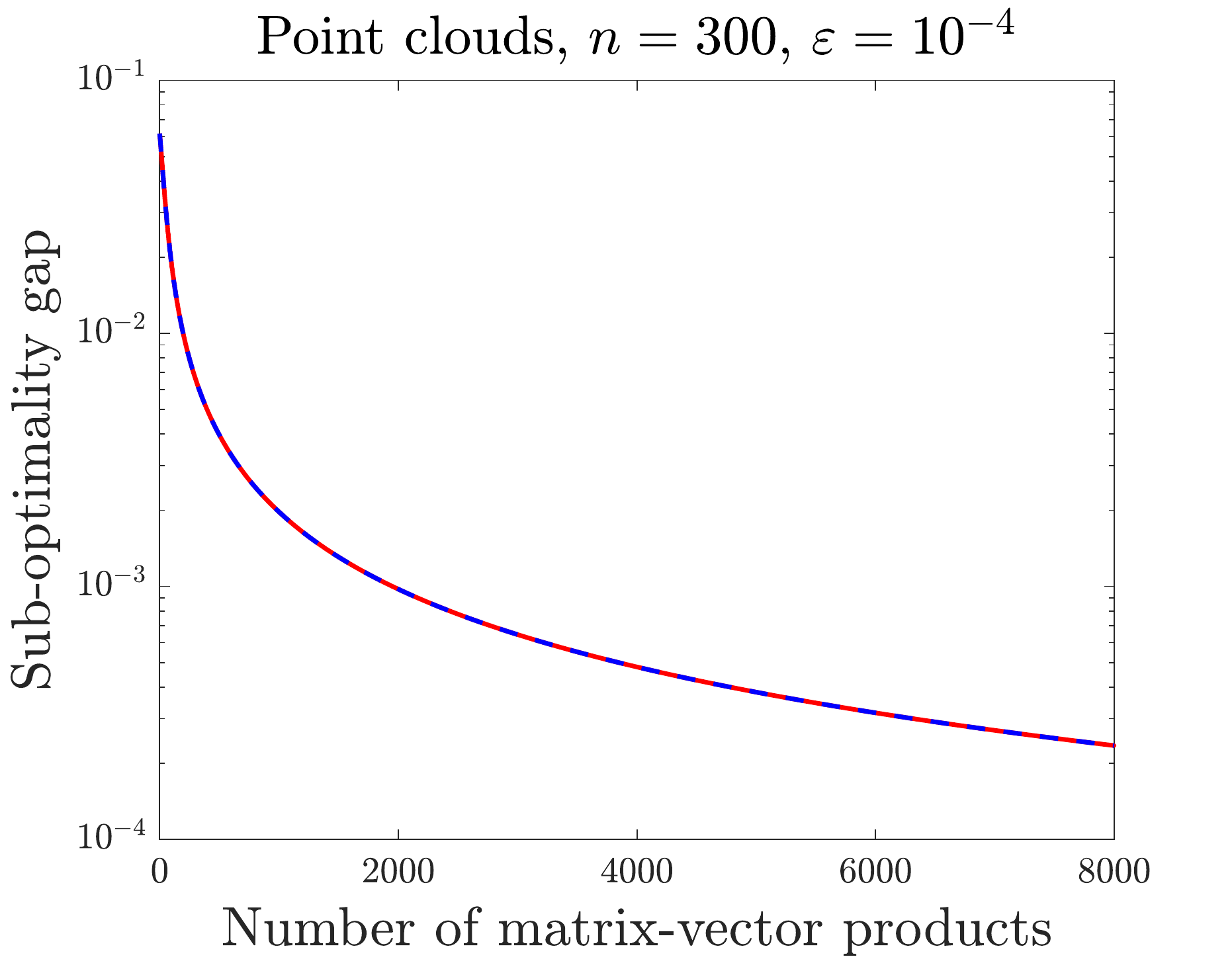}
    \caption{The proposed extragradient method (Algorithm~\ref{alg:main}) with adjustment step vs.~the version without adjustment. The problem settings are the same as those in Figure~\ref{fig:compare_alg}. }
    \label{fig:adjustment}
\end{figure}

We make a few more comments.
 Our extragradient method converges fast (almost linearly in early iterations) in the ``Synthetic'' and ``MNIST'' settings, while at a sublinear rate in the ``Point clouds'' setting.
This is likely because, in the latter case, the optimal transport plan typically lies on or close to the boundary of the polytope, where the optimization landscape is less well-conditioned and thus results in slower convergence.
 Another interesting observation is that the adjustment step (cf.~\eqref{eq:update-adjust}) in Algorithm~\ref{alg:main} turns out to have a significant impact on the practical performance under the ``Synthetic'' and ``MNIST'' settings.
This can be seen from Figure~\ref{fig:adjustment}, where the solid lines stand for the extragradient method with fine-tuned parameters (as in Figure~\ref{fig:compare_alg}), and the dashed lines correspond to the same except that the adjustment step in Algorithm~\ref{alg:main} is skipped.
These results showcase that the adjustment step can help avoid getting stuck at undesirable points, thus accelerating convergence.

\subsubsection{Experiments with squared Euclidean distances}
\begin{figure}[tbp]
    \centering 
    \includegraphics[width=.33\textwidth]{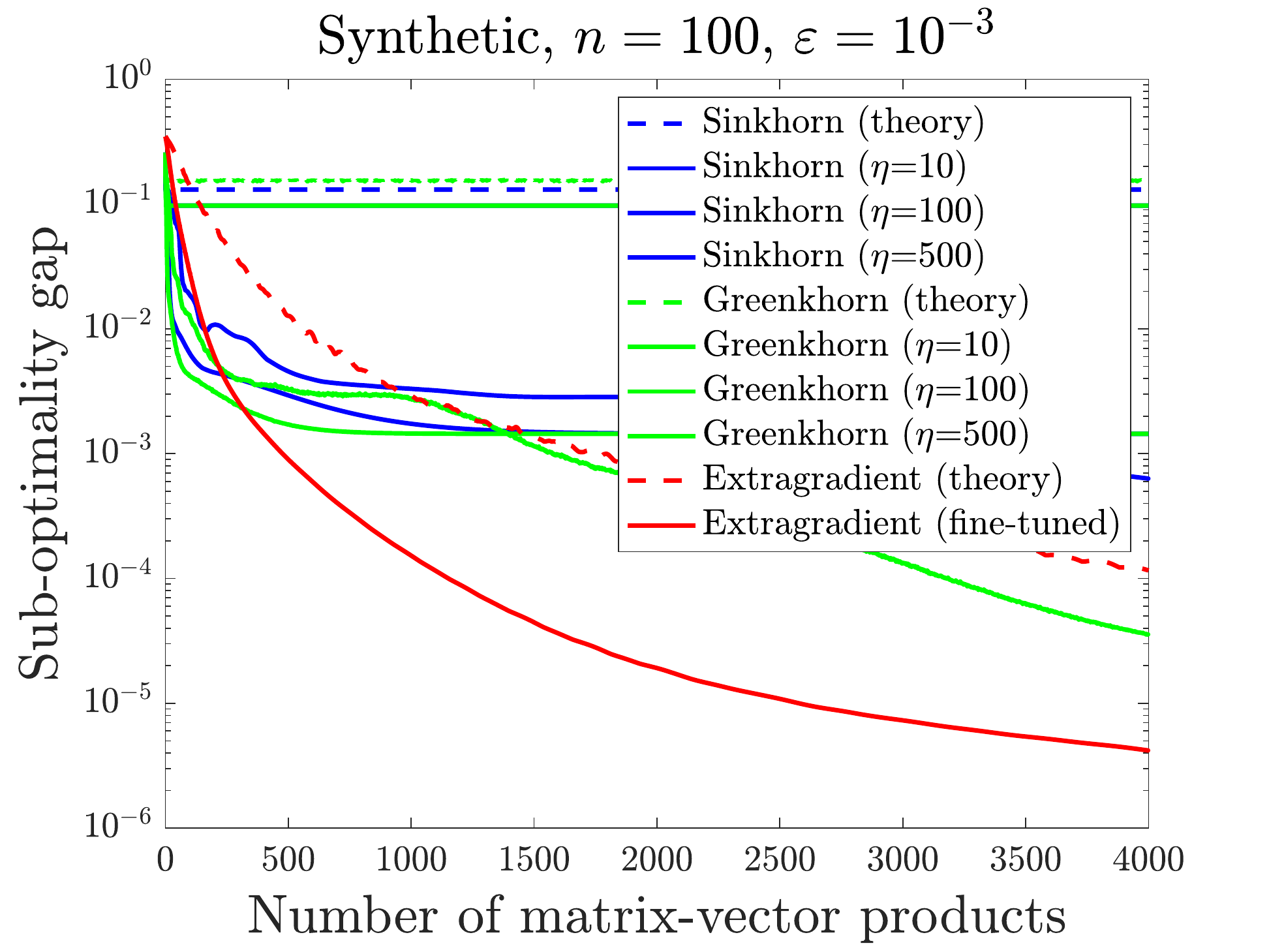}%
    \includegraphics[width=.33\textwidth]{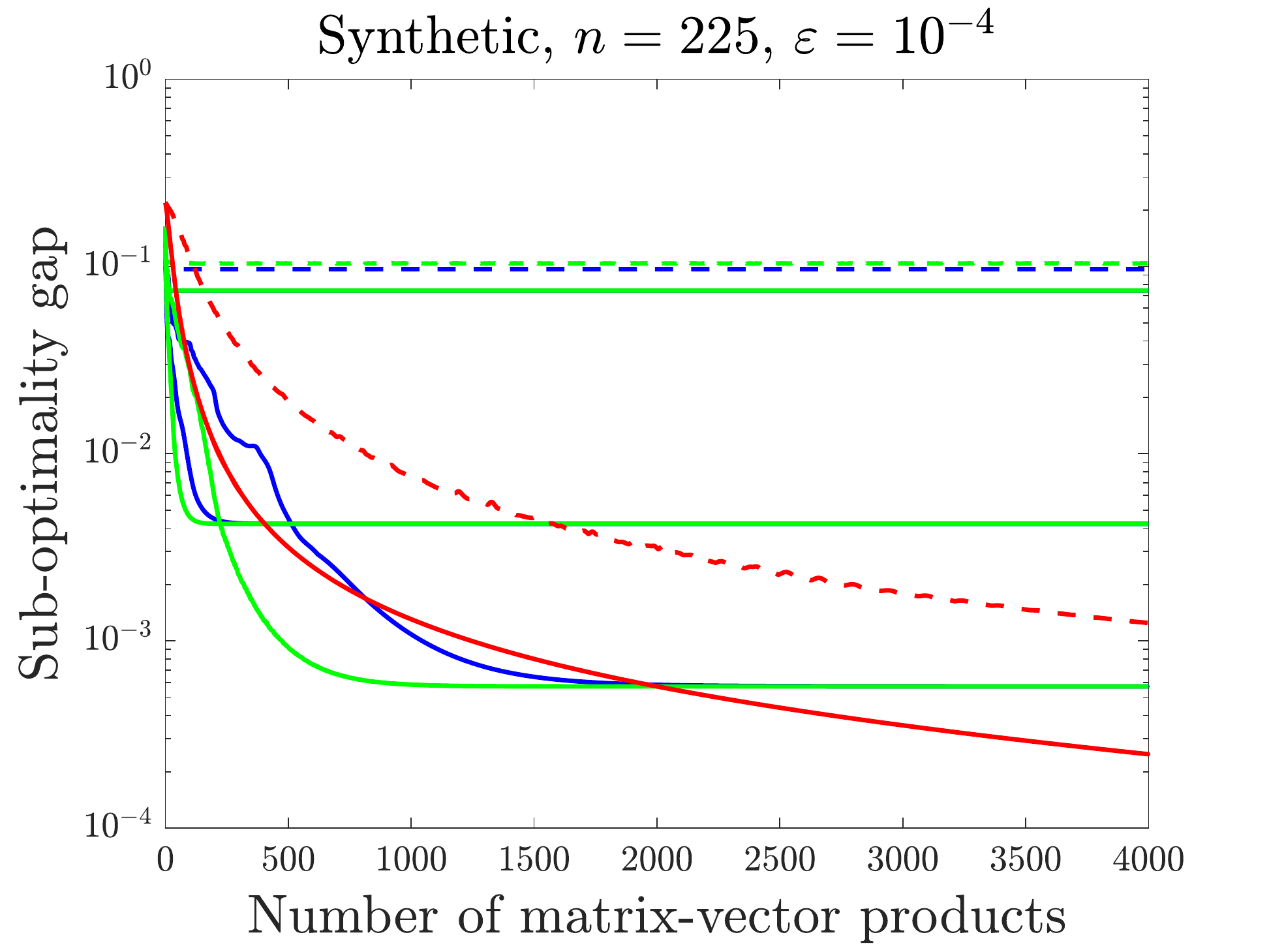}%
    \includegraphics[width=.33\textwidth]{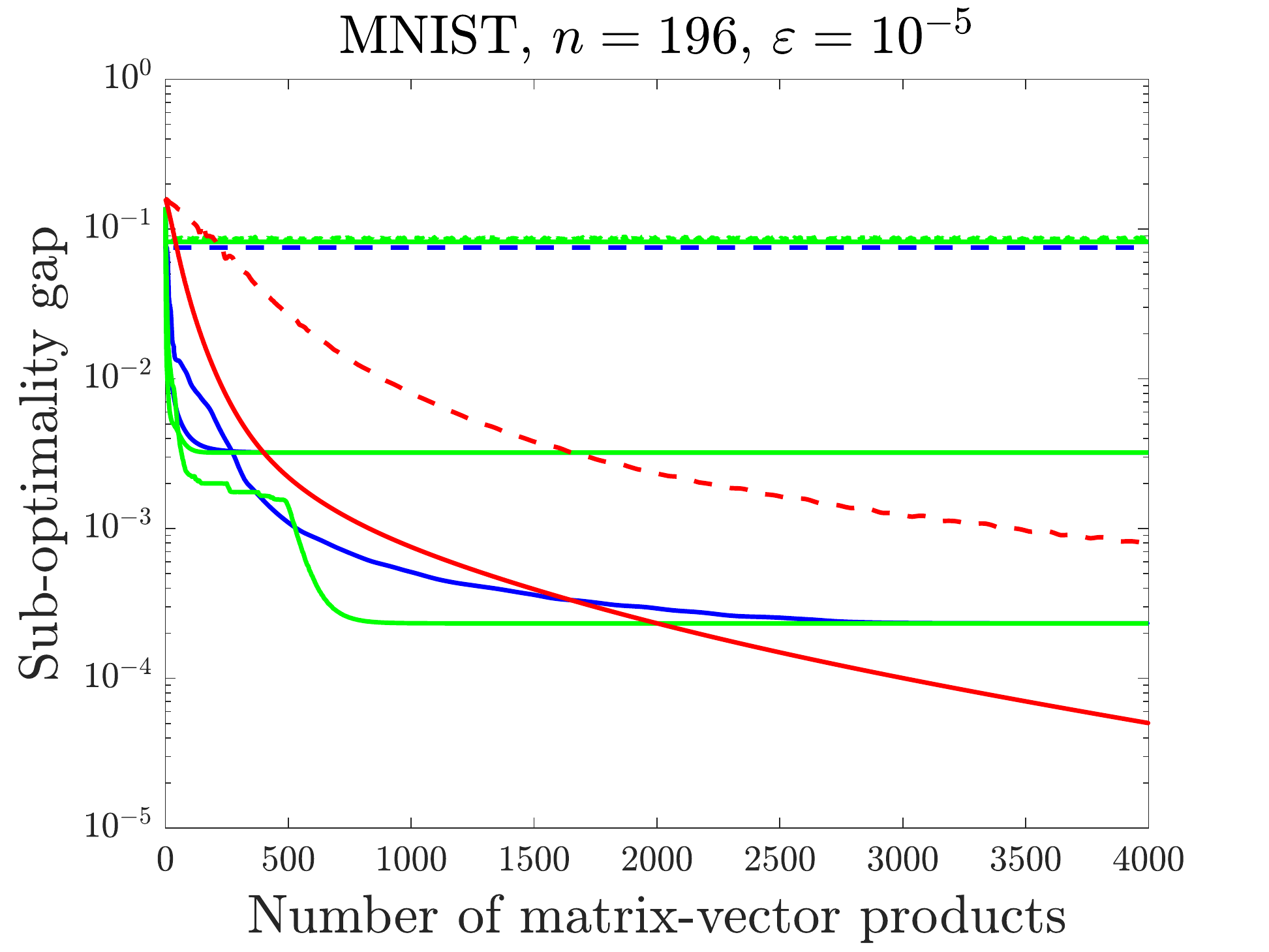}
    \includegraphics[width=.33\textwidth]{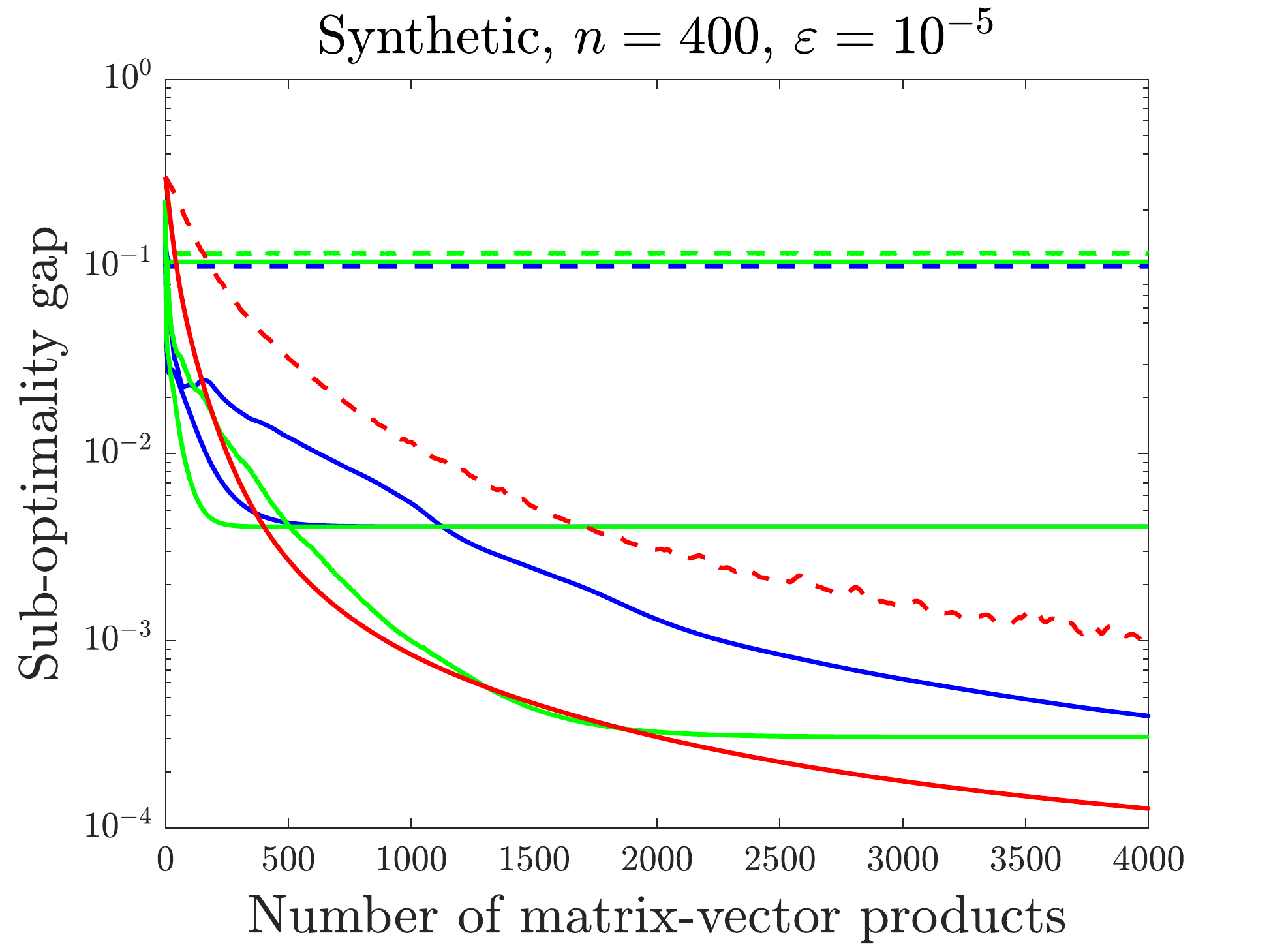}%
    \includegraphics[width=.33\textwidth]{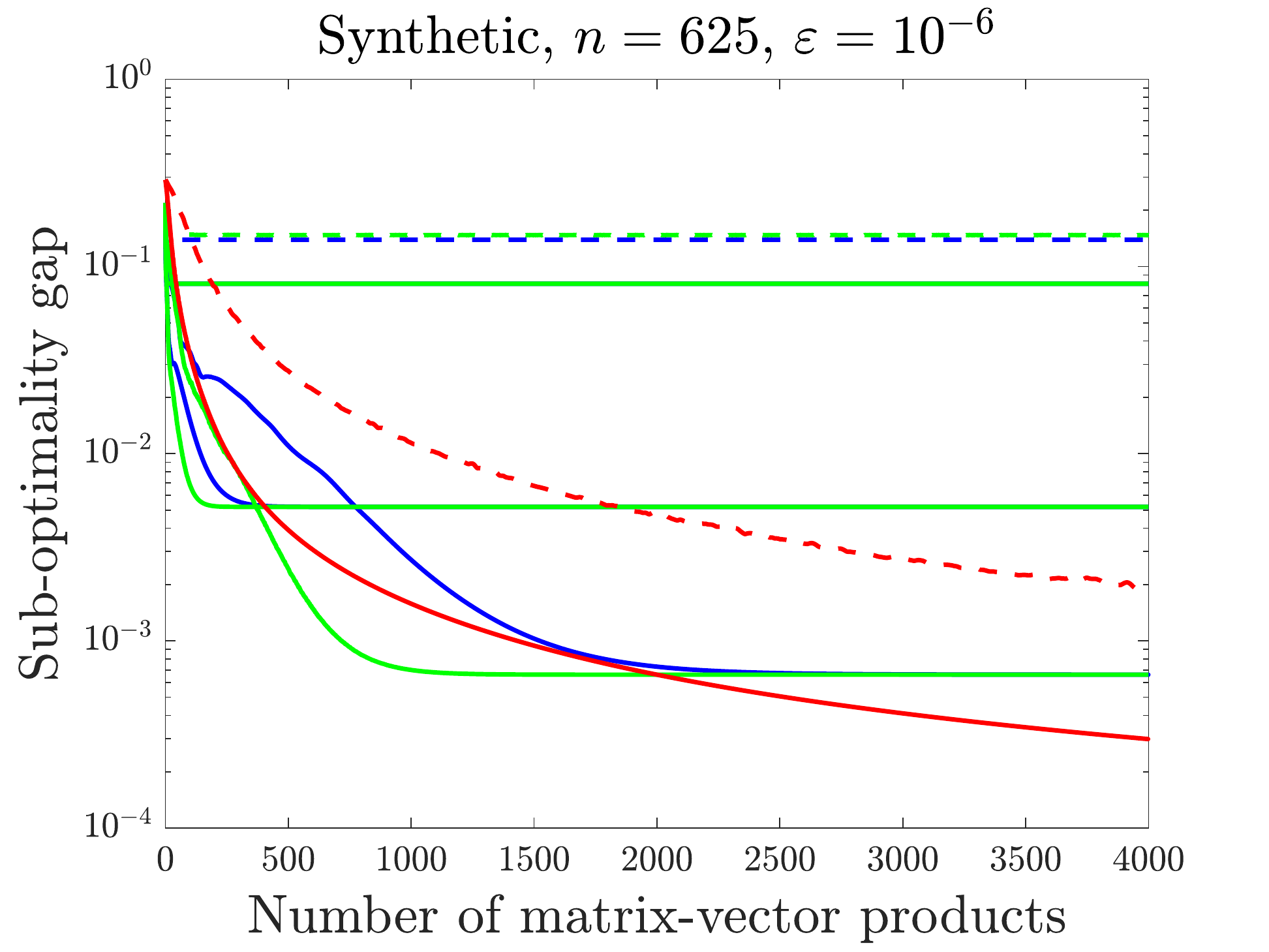}%
    \includegraphics[width=.33\textwidth]{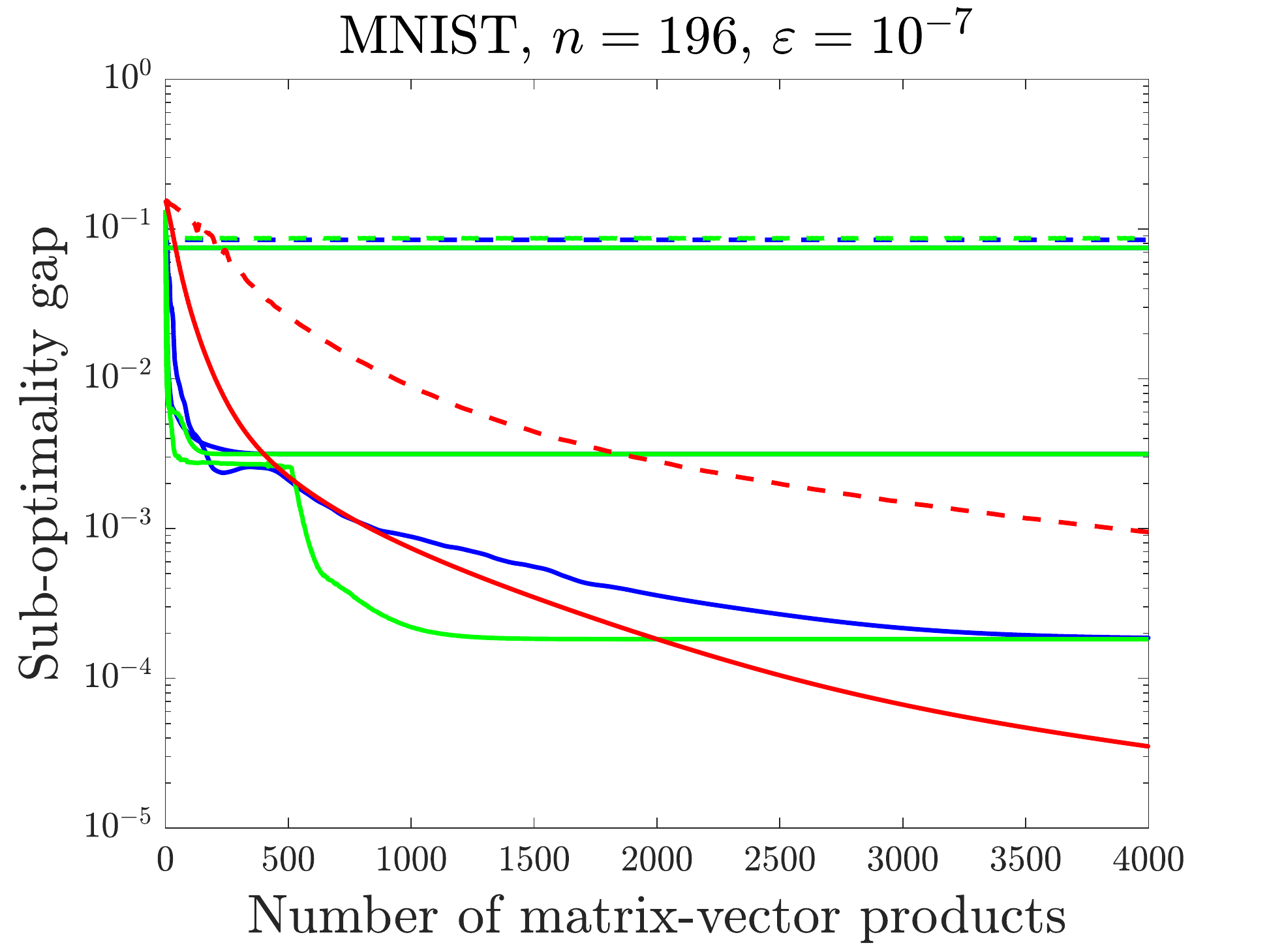}
    \caption{Empirical comparisons of algorithms using the 2-Wasserstein distance.}
    \label{fig:compare_l2}
\end{figure}

In addition to previous experiments, we have further conducted numerical experiments for the scenarios where the cost matrix is computed based on squared Euclidean distances, corresponding to the 2-Wasserstein distance. The numerical results of these experiments are presented in Figure~\ref{fig:compare_l2}. As can be seen from these plots, our proposed algorithm consistently demonstrates superior performance compared to the other tested algorithms, while exhibiting a smoother learning curve.


\subsection{Comparisons with more recent approaches}

\begin{figure}[tbp]
    \centering
    \includegraphics[width=.33\textwidth]{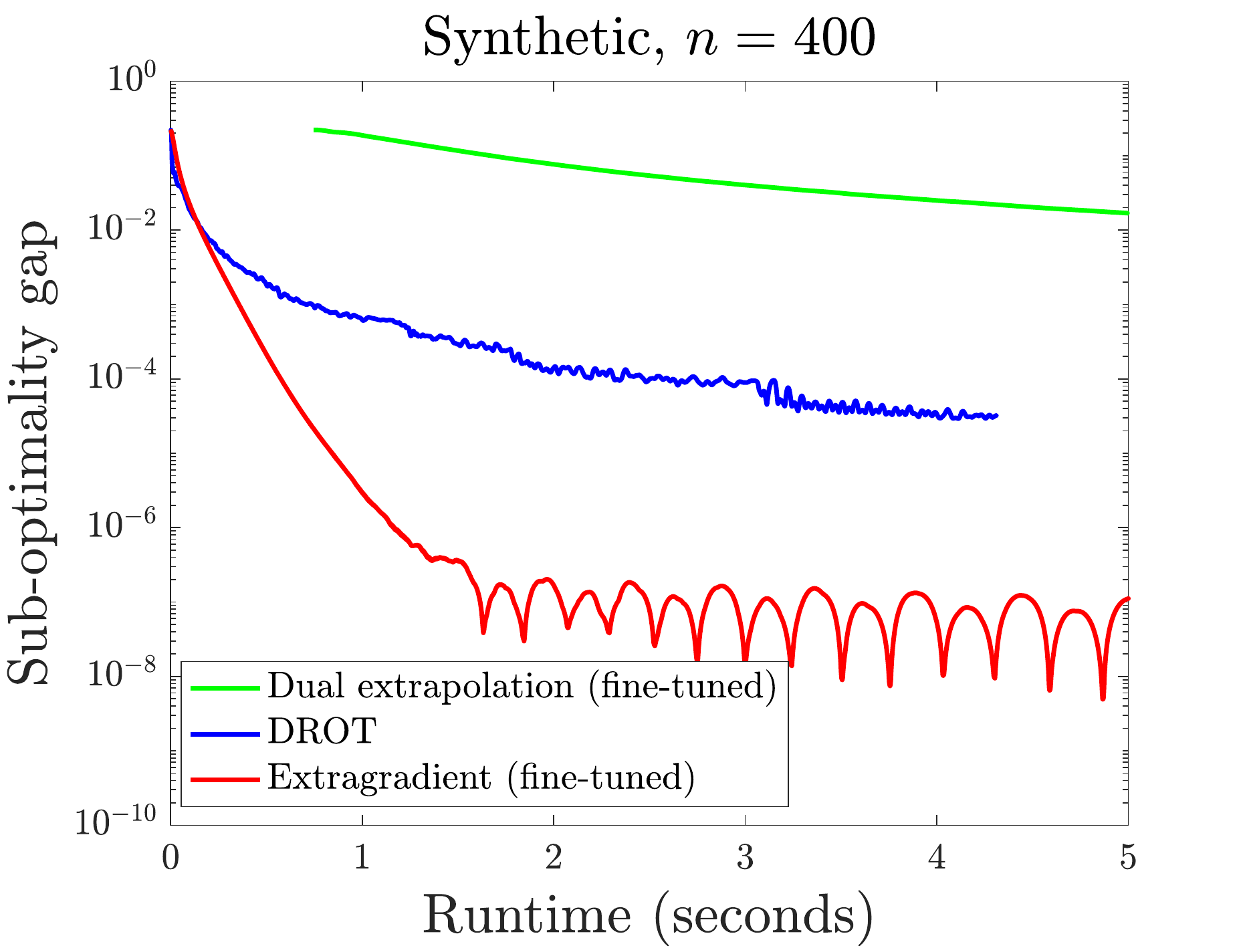}%
    \includegraphics[width=.33\textwidth]{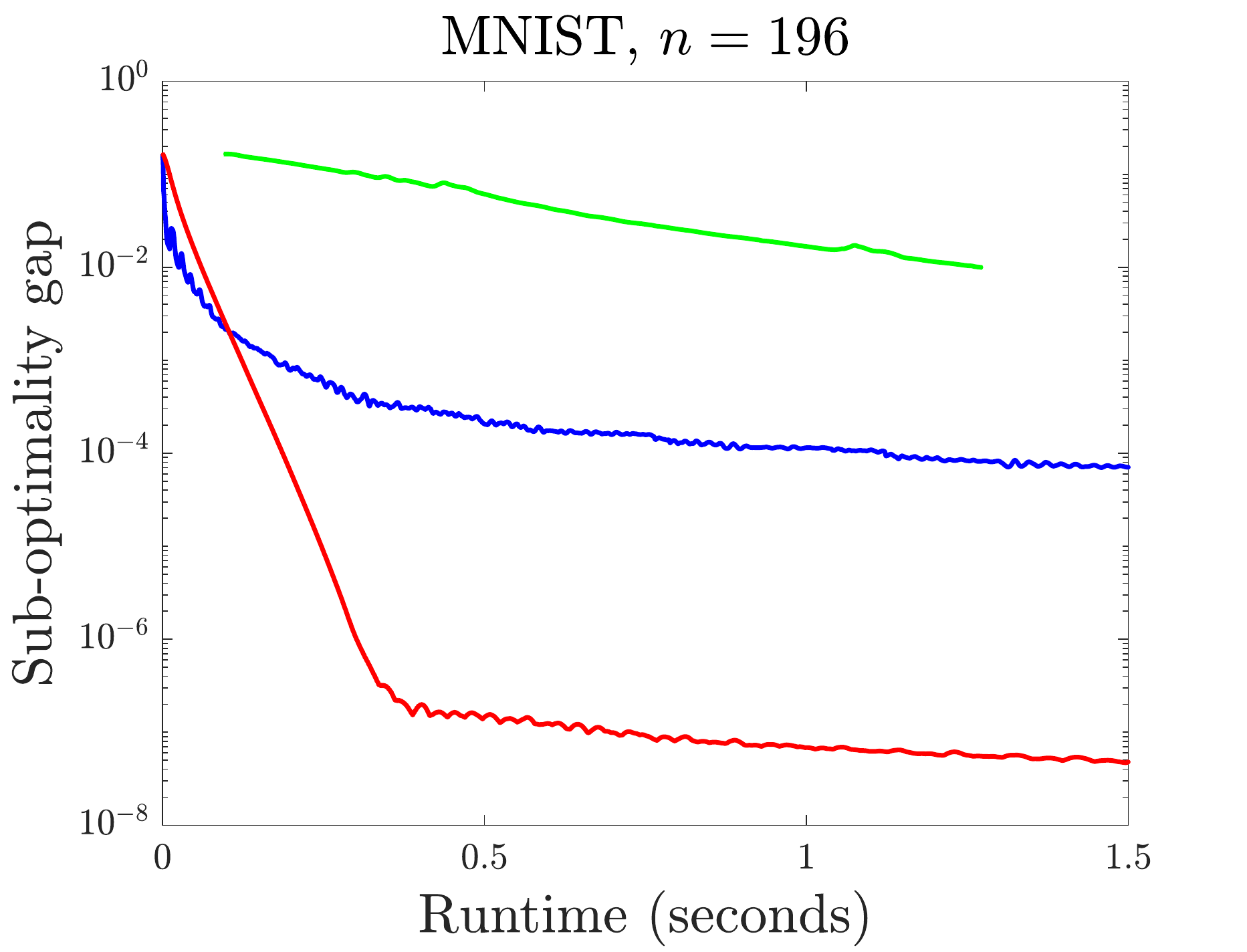}%
    \includegraphics[width=.33\textwidth]{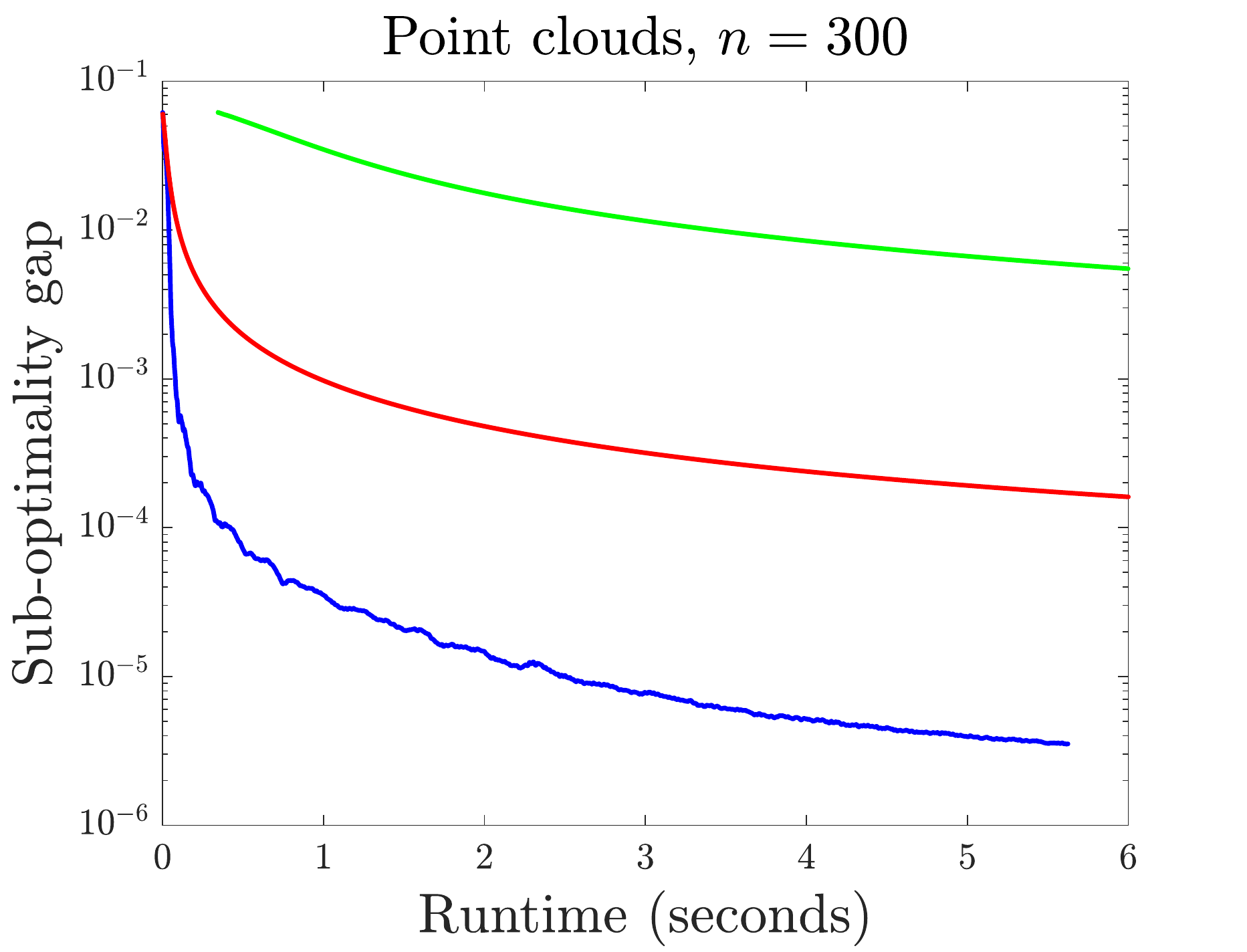}
    \caption{Empirical comparisons between our extragradient method and two recently proposed algorithms, namely the dual extrapolation method \cite[Algorithm 3]{jambulapati2019direct} and the DROT method \cite[Algorithm 1]{mai2022a}. Each curve is an average over 10 independent trials. }
    \label{fig:compare_sota}
\end{figure}

We further implement, and compare our algorithm numerically with,  two recently proposed methods: 
(a) the dual extrapolation method from \cite[Algorithm 3]{jambulapati2019direct}\footnote{As mentioned by the authors of \cite{jambulapati2019direct}, the algorithm that they actually implemented and tested numerically in their work is different from the one with theoretical guarantees (i.e.~the one that we implement and compare with); its (pseudo-)codes are not yet publicly available.}, 
and (b) the DROT method from \cite[Algorithm 1]{mai2022a}. 
For the dual extrapolation method, we try our best to fine-tune its parameters (e.g., the step sizes and the numbers of iterations for the inner loops), while for the DROT method, we follow the choices of initialization and parameters described in \cite[Section~4]{mai2022a}.
The empirical results are illustrated in Figure~\ref{fig:compare_sota}.
The dual extrapolation method is outperformed by the other approaches in all three settings.  
Our extragradient method achieves the best numerical performance in both ``Synthetic'' and ``MNIST'' settings, 
but is outperformed by the DROT algorithm in the ``Point Clouds'' setting. 
These numerical findings (together with the previous numerical results) suggest that there might not be a single algorithm that empirically dominates others in every setting, and each algorithm has its own pros and cons. 

\subsection{Comparisons with other cost-free acceleration  strategies}
\begin{figure}[tbp]
    \centering
    \includegraphics[width=.33\textwidth]{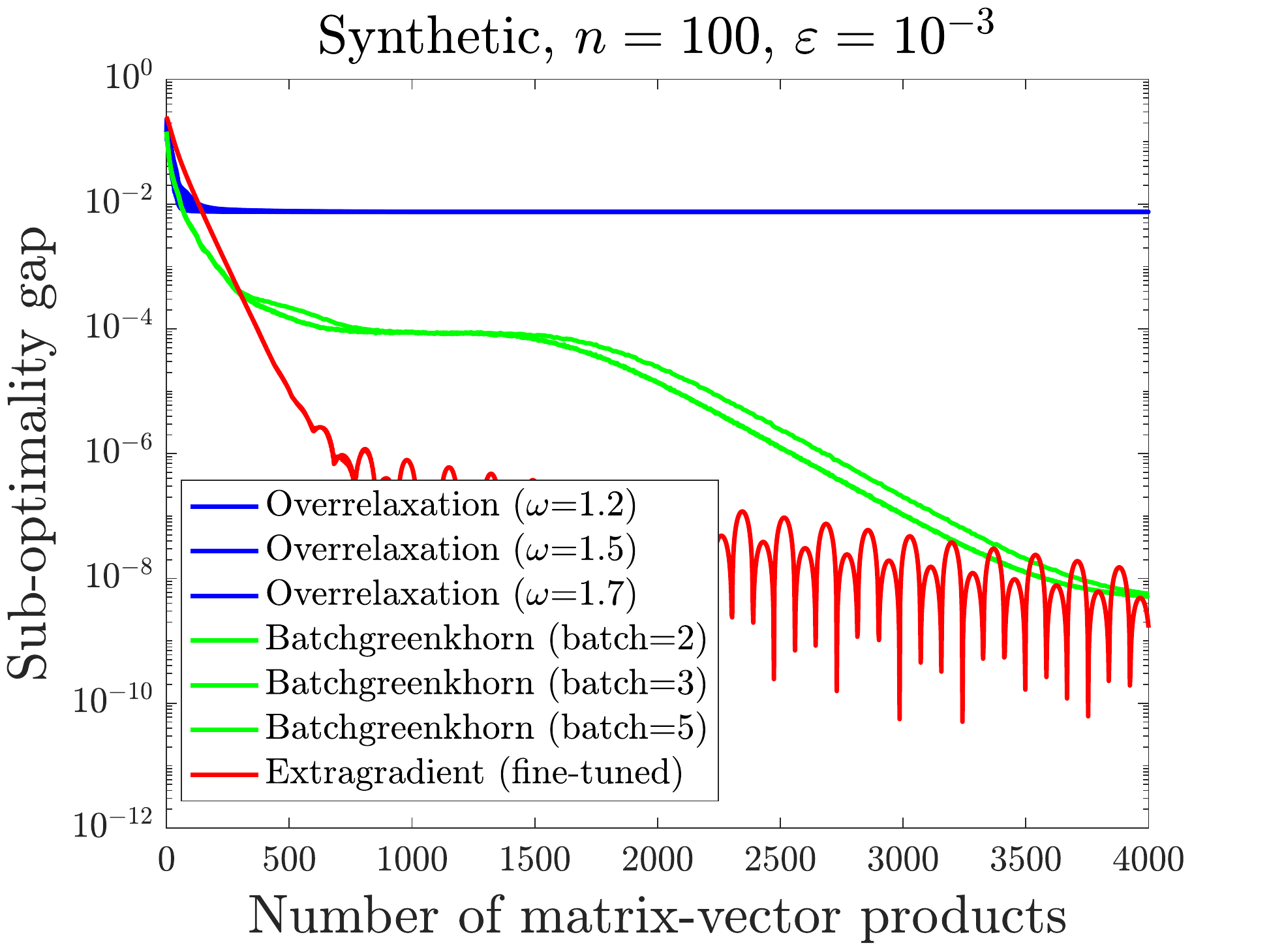}%
    \includegraphics[width=.33\textwidth]{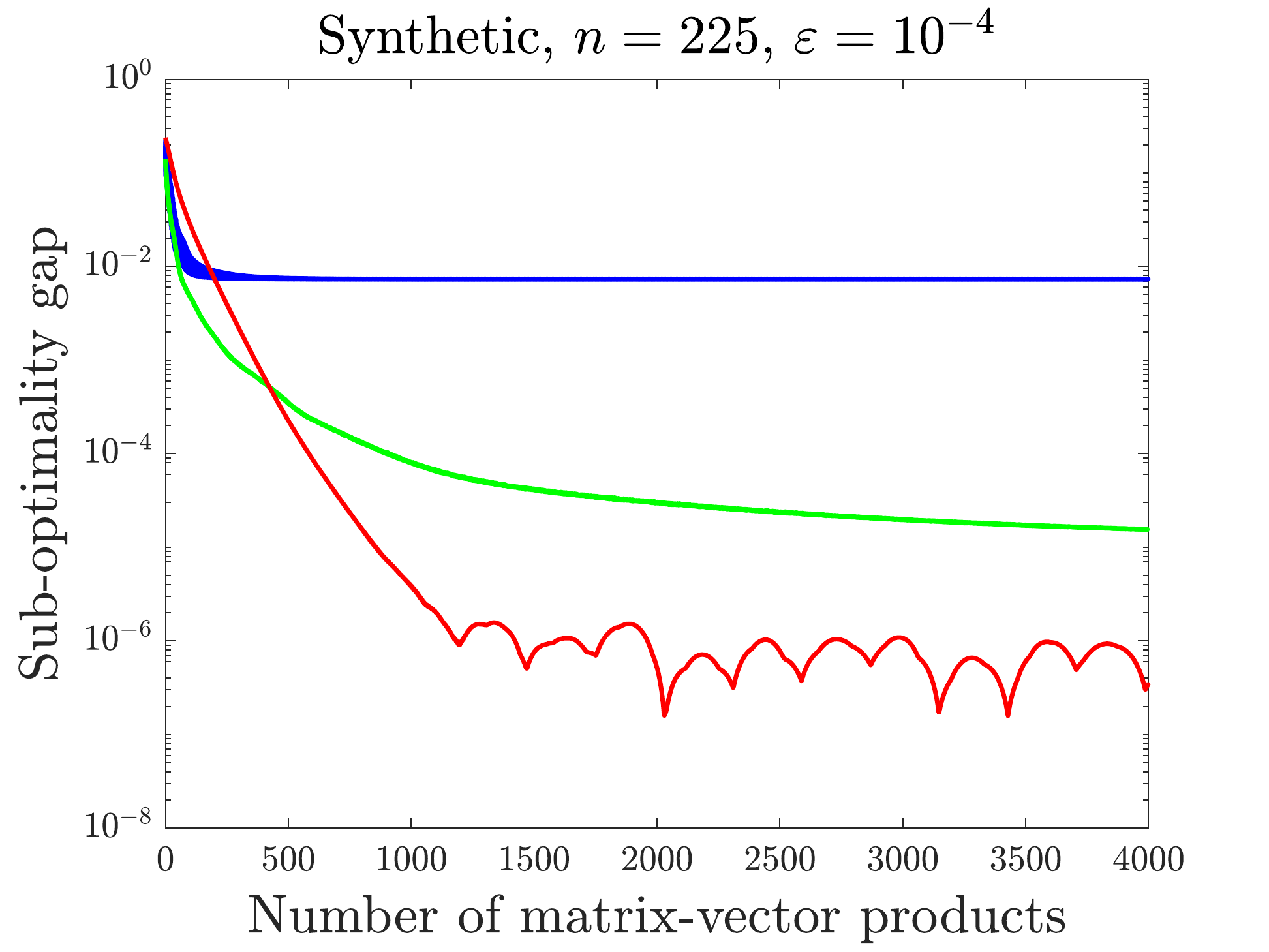}%
    \includegraphics[width=.33\textwidth]{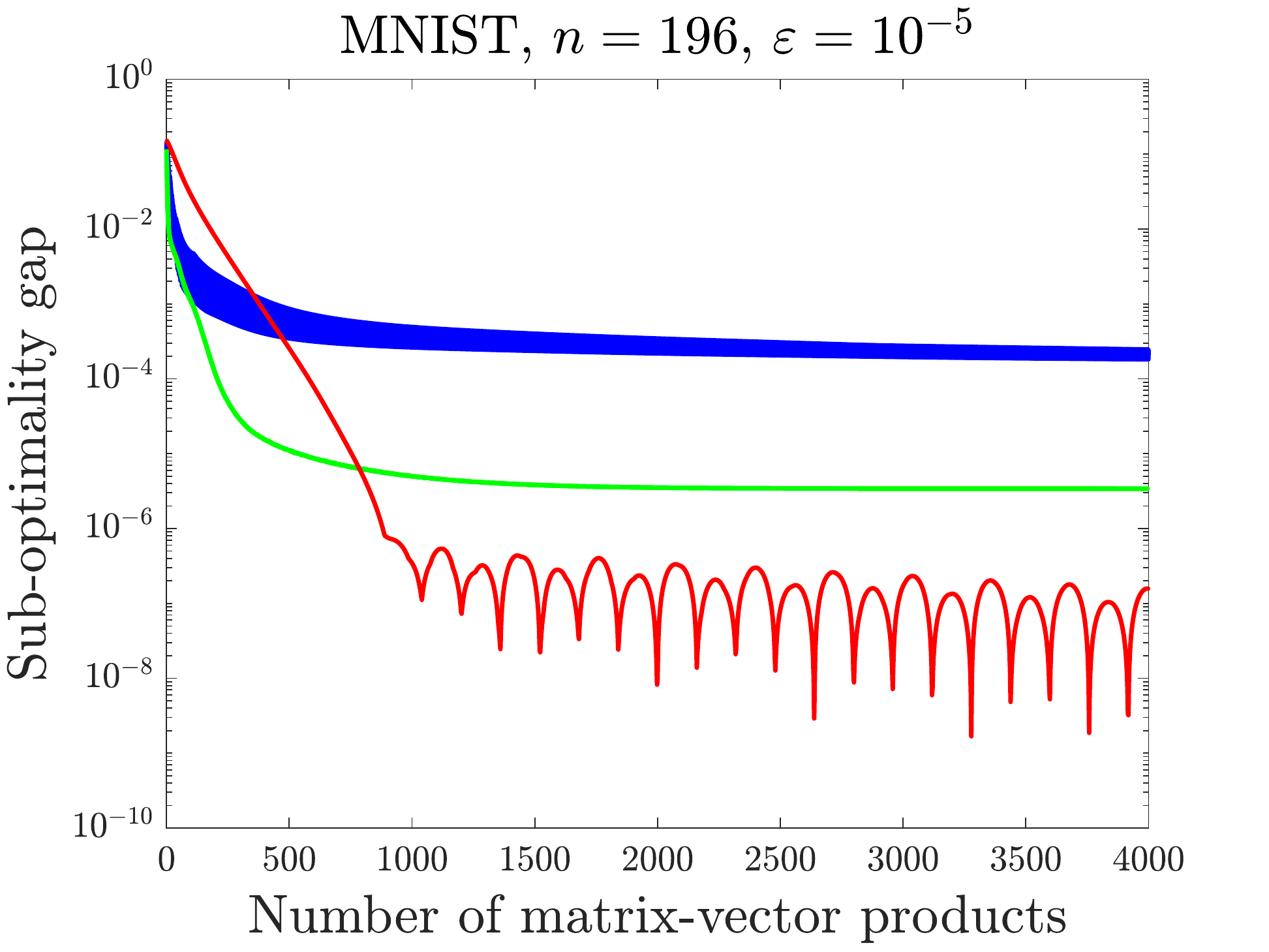}
    \includegraphics[width=.33\textwidth]{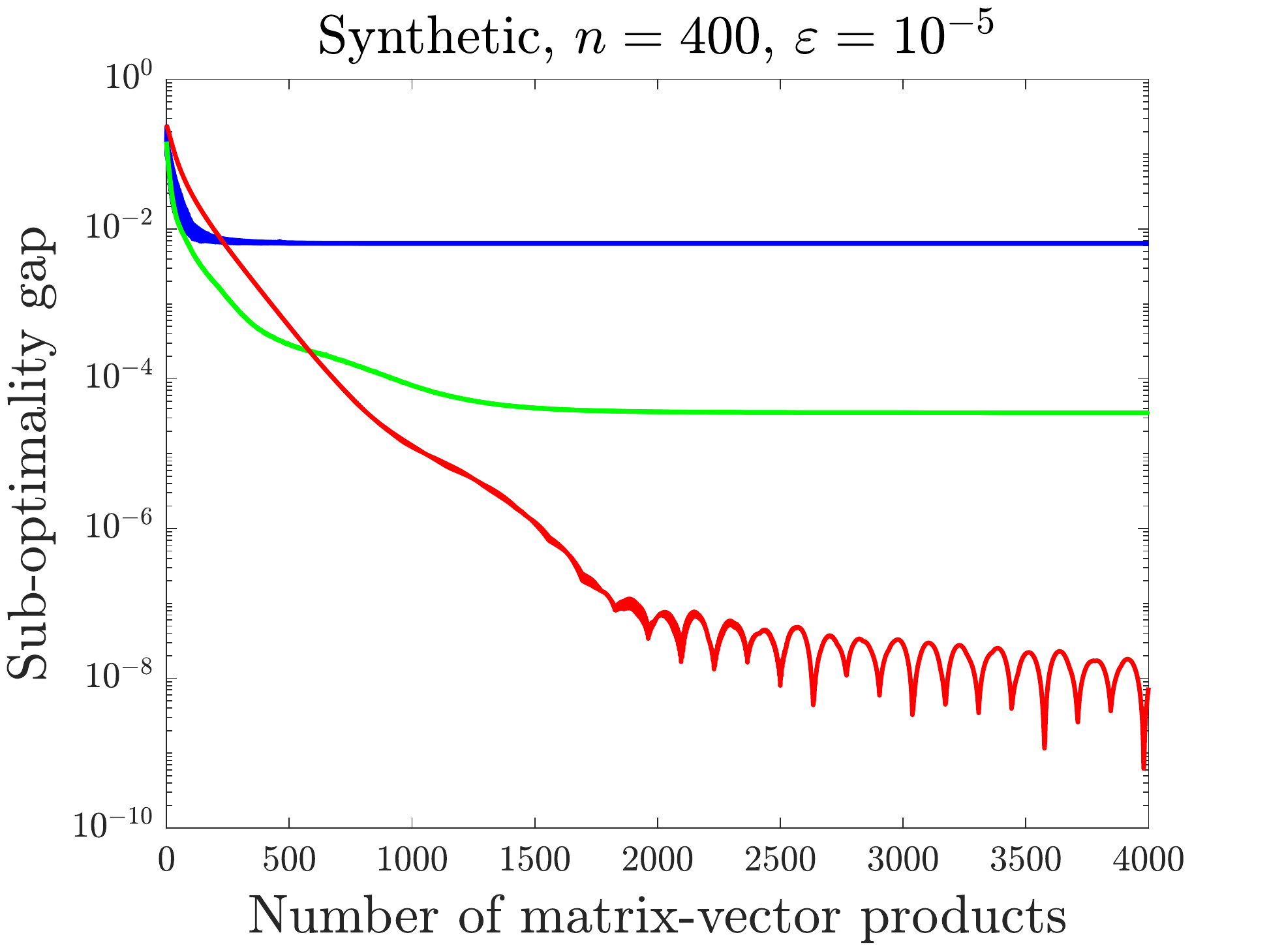}%
    \includegraphics[width=.33\textwidth]{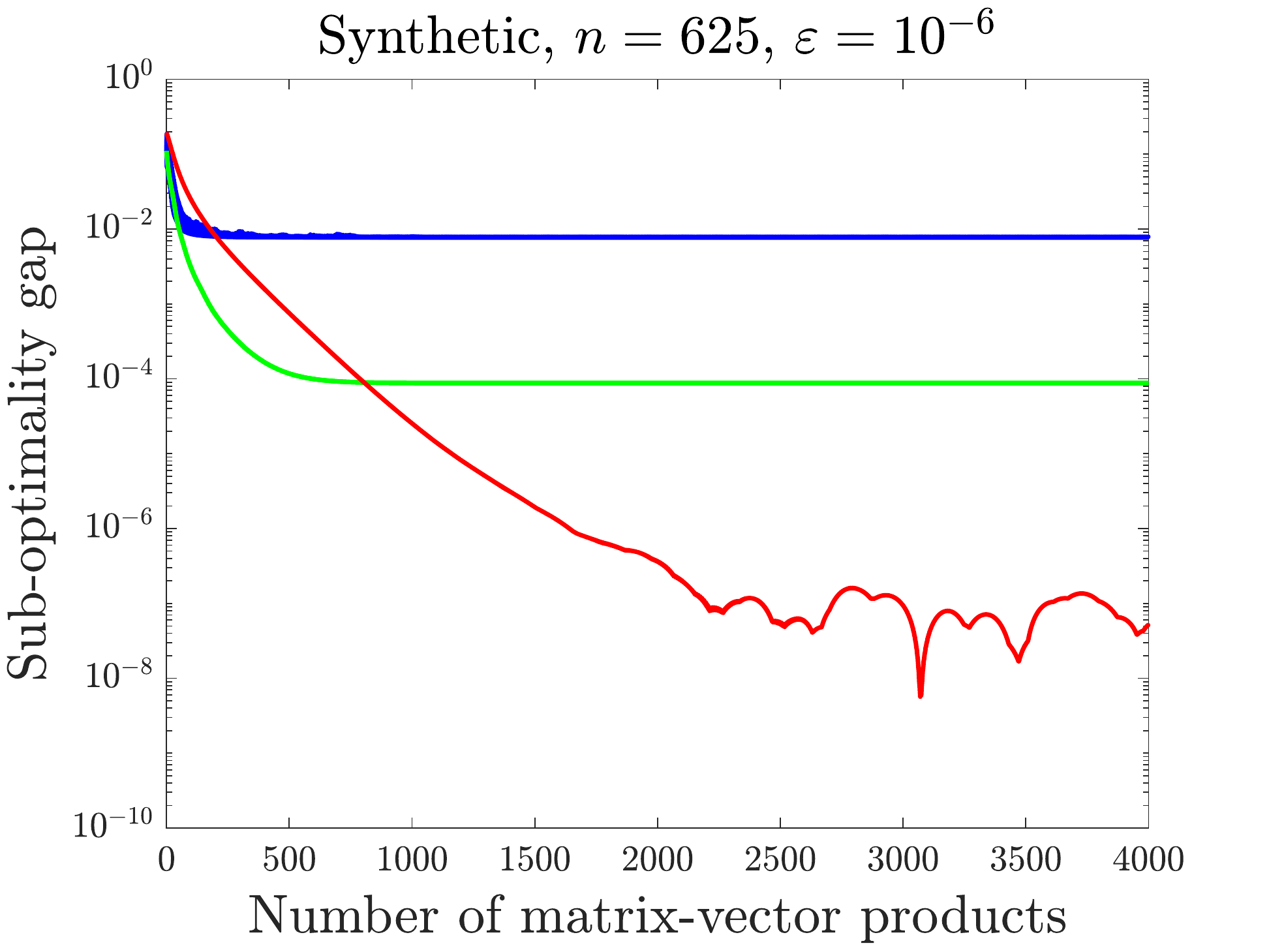}%
    \includegraphics[width=.33\textwidth]{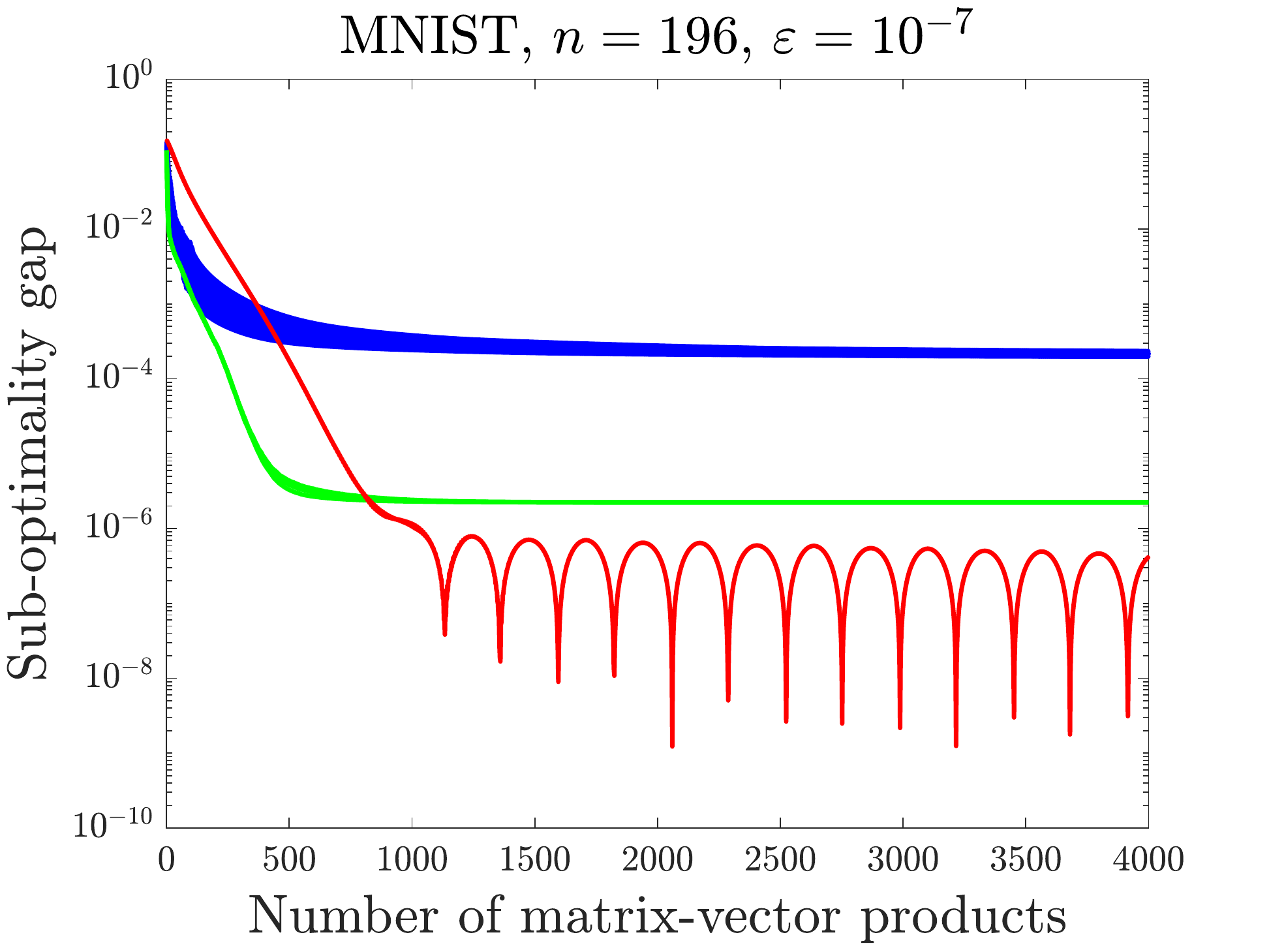}
    \caption{Empirical comparisons of our extragradient method and two cost-free acceleration strategies, namely, the overrelaxation method \cite{Lehmann2021note} and the batching Greenkhorn method \cite{kostic2022batch}. Each curve represents an average over 10 independent trials.}
    \label{fig:compare_other_acc}
\end{figure}

To further assess the efficiency of the proposed algorithm, we perform comparisons with a few other acceleration methods: the overrelaxation variant of the Sinkhorn algorithm~\cite{Lehmann2021note}, and the batched version of the Greenkhorn algorithm~\cite{kostic2022batch}. For the overrelaxation technique, we have tested multiple relaxation parameters, specifically $\omega \in \{1.2, 1.5, 1.7\}$. In the case of the batching method, we have adjusted the batch size settings to $\{2, 3, 5\}$. These numerical results are depicted in Figure~\ref{fig:compare_other_acc}. Our results demonstrate that the proposed extragradient method consistently outperforms both the overrelaxation and batching methods across various setups, in both the ``Synthetic" and ``MNIST" datasets. This numerical superiority is evident across all tested hyperparameter configurations.


\section{Analysis}
\label{sec:analysis}

This section presents the proof for our main result: Theorem~\ref{thm:theory-OT}.

\subsection{Preliminary facts and additional notation}

Before proceeding, let us collect several elementary facts concerning the ``adjusting'' step in our algorithm. 
Consider any probability vector $\bm{q} = [q_i]_{1\leq i\leq d}\in \Delta_d$, and transform it into another probability vector $\bm{q}^{\trunc}= [q_i^{\trunc}]_{1\leq i\leq d} \in \Delta_d$ via the following two steps: 
\begin{align}
	q_{i}^{\trunc} & =\frac{\max\big\{ q_{i},\,e^{-B}\|\bm{q}\|_{\infty}\big\}}{\sum_{j=1}^{d}\max\big\{ q_{j},\,e^{-B}\|\bm{q}\|_{\infty}\big\}},\qquad i=1,\cdots,d	.
\end{align}
We first make note of  basic property of this transformation:
%
\begin{align}
	\max_{1\leq i\leq d}\frac{q_{i}}{q_{i}^{\trunc}}\le
	1+de^{-B}.
 	\label{eq:trunc}
\end{align}
%
%
In addition, consider the special two-dimensional case where $\bm{q} = [q_+, q_-]^{\top}\in \Delta_2$ and let $\bm{q}^{\trunc} = [q_+^{\trunc}, q_-^{\trunc}]^{\top}\in \Delta_2$. 
If $B>0$, then it holds that 
\begin{equation}
\frac{q_{+}^{\trunc}}{q_{-}^{\trunc}}=\begin{cases}
\min\big\{\frac{q_{+}}{q_{-}},e^{B}\big\}\leq e^{B}, & \text{if }q_{+}\geq q_{-},\\
\max\big\{\frac{q_{+}}{q_{-}},e^{-B}\big\}\geq e^{-B}, & \text{if }q_{+}<q_{-}.
\end{cases}\label{eq:trunc-dimension-2}
\end{equation}
The proof of \eqref{eq:trunc} and \eqref{eq:trunc-dimension-2} are straightforward and hence omitted here for brevity. 
%

Moreover, we would also like to introduce several additional convenient notation that is useful for presenting the proof. 
Define
{\small
\begin{subequations}
	\label{eq:defn-zeta-all}
\begin{align}
\bm{\zeta}^{t} & \coloneqq\bigg(\Big\{\frac{1}{\eta_{p,i}}\bm{p}_{i}^{t}\Big\}_{i=1}^{n},\Big\{\frac{1}{\eta_{\mu,j}}\bm{\mu}_{j}^{t}\Big\}_{j=1}^{n}\bigg),\label{eq:defn-zeta-t}\\
\overline{\bm{\zeta}}^{t} & \coloneqq\bigg(\Big\{\frac{1}{\eta_{p,i}}\overline{\bm{p}}_{i}^{t}\Big\}_{i=1}^{n},\Big\{\frac{1}{\eta_{\mu,j}}\overline{\bm{\mu}}_{j}^{t}\Big\}_{j=1}^{n}\bigg),\label{eq:defn-zeta-bar} \\
\bm{\zeta}^{t,\trunc} & \coloneqq\bigg(\Big\{\frac{1}{\eta_{p,i}}\bm{p}_{i}^{t}\Big\}_{i=1}^{n},\Big\{\frac{1}{\eta_{\mu,j}}\bm{\mu}_{j}^{t,\trunc}\Big\}_{j=1}^{n}\bigg),\label{eq:defn-zeta-t-trunc}\\
\bm{\zeta}^{\star} & \coloneqq\bigg(\Big\{\frac{1}{\eta_{p,i}}\bm{p}_{i}^{\star,\mathsf{reg}}\Big\}_{i=1}^{n},\Big\{\frac{1}{\eta_{\mu,j}}\bm{\mu}_{j}^{\star,\mathsf{reg}}\Big\}_{j=1}^{n}\bigg),\label{eq:defn-zeta-star}
\end{align}
\end{subequations}
}
where $\big\{\bm{p}_{i}^{\star,\mathsf{reg}}\big\}_{i=1}^{n},\big\{\bm{\mu}_{j}^{\star,\mathsf{reg}}\big\}_{j=1}^{n}$ denotes the  optimizer of the entropy-regularized minimax problem \eqref{eq:game-entropy}. 
Additionally, for any $\bm{\zeta}^{(1)}=\big(\{\frac{1}{\eta_{p,i}}\bm{p}_{i}^{(1)}\}_{i=1}^{n}, \{\frac{1}{\eta_{\mu,j}}\bm{\mu}_{j}^{(1)} \}_{j=1}^{n}\big)$ 
and $\bm{\zeta}^{(2)}=\big(\{\frac{1}{\eta_{p,i}}\bm{p}_{i}^{(2)}\}_{i=1}^{n}, \{\frac{1}{\eta_{\mu,j}}\bm{\mu}_{j}^{(2)} \}_{j=1}^{n}\big)$ 
(with the $\bm{p}_i^{(1)}$'s, $\bm{p}_i^{(2)}$'s, $\bm{\mu}_i^{(1)}$'s and $\bm{\mu}_i^{(2)}$'s being probability vectors), we introduce a weighted KL divergence metric: 
{\small
\begin{align}
	\dist\big(\bm{\zeta}^{(1)}\parallel \bm{\zeta}^{(2)}\big) \coloneqq 
	\sum_{i = 1}^{n} \frac{1}{\eta_{p,i}} \mathsf{KL}\big( \bm{p}^{(1)}_i \parallel \bm{p}^{(2)}_i \big) + 
	\sum_{j = 1}^{n} \frac{1}{\eta_{\mu,j}} \mathsf{KL}\big( \bm{\mu}^{(1)}_j \parallel \bm{\mu}^{(2)}_j \big).
	\label{eq:defn-KL-gen}
\end{align}
}

\subsection{Proof of Theorem~\ref{thm:theory-OT}}

We are now positioned to prove Theorem~\ref{thm:theory-OT}. 
To begin with, we make the observation that  
{\small
\begin{align*}
\mathsf{KL}(\bm{\mu}_{j}^{\star,\mathsf{reg}}\parallel\bm{\mu}_{j}^{t,\trunc}) & =\sum_{s\in\{+,-\}}\mu_{j,s}^{\star,\mathsf{reg}}\log\frac{\mu_{j,s}^{\star,\mathsf{reg}}}{\mu_{j,s}^{t,\trunc}}\\
&=\sum_{s\in\{+,-\}}\mu_{j,s}^{\star,\mathsf{reg}}\log\frac{\mu_{j,s}^{\star,\mathsf{reg}}}{\mu_{j,s}^{t}}+\sum_{s\in\{+,-\}}\mu_{j,s}^{\star,\mathsf{reg}}\log\frac{\mu_{j,s}^{t}}{\mu_{j,s}^{t,\trunc}}\\
 & \leq\mathsf{KL}(\bm{\mu}_{j}^{\star,\mathsf{reg}}\parallel\bm{\mu}_{j}^{t})+\sum_{s\in\{+,-\}}\mu_{j,s}^{\star,\mathsf{reg}}\log\max\left\{ \frac{\mu_{j,+}^{t}}{\mu_{j,+}^{t,\trunc}},\frac{\mu_{j,-}^{t}}{\mu_{j,-}^{t,\trunc}}\right\} \\
 & \leq\mathsf{KL}(\bm{\mu}_{j}^{\star,\mathsf{reg}}\parallel\bm{\mu}_{j}^{t})+\log\big(1+2e^{-B}\big),
\end{align*}
}
where the last inequality arises from \eqref{eq:trunc} and the fact that $\sum_{s\in\{+,-\}}\mu_{j,s}^{\star,\mathsf{reg}}=1$. 
%
%
As a result, combine the above inequality with the definitions \eqref{eq:defn-zeta-t-trunc} and \eqref{eq:defn-zeta-t} to reach 
{\small
\begin{align}
	& \dist\big(\bm{\zeta}^{\star}\parallel\bm{\zeta}^{t,\trunc}\big)  =\sum_{i=1}^{n}\frac{1}{\eta_{p,i}}\mathsf{KL}\big(\bm{p}_{i}^{\star,\mathsf{reg}}\parallel\bm{p}_{i}^{t}\big)+\sum_{j=1}^{n}\frac{1}{\eta_{\mu,j}}\mathsf{KL}\big(\bm{\mu}_{j}^{\star,\mathsf{reg}}\parallel\bm{\mu}_{j}^{t,\trunc}\big) \notag\\
 & \qquad \leq\sum_{i=1}^{n}\frac{1}{\eta_{p,i}}\mathsf{KL}\big(\bm{p}_{i}^{\star,\mathsf{reg}}\parallel\bm{p}_{i}^{t}\big)+\sum_{j=1}^{n}\frac{1}{\eta_{\mu,j}}\mathsf{KL}\big(\bm{\mu}_{j}^{\star,\mathsf{reg}}\parallel\bm{\mu}_{j}^{t}\big) 
	+\log\big(1+2e^{-B}\big) \sum_{j=1}^{n}\frac{1}{\eta_{\mu,j}} \notag\\
	&\qquad\quad =\dist\big(\bm{\zeta}^{\star}\parallel\bm{\zeta}^{t}\big)+\log\big(1+2e^{-B}\big) \cdot \sum_{j=1}^{n}  \frac{c_{j}+ C_3/n}{15C_{2}\sqrt{B}} \notag\\
	&\qquad\quad =\dist\big(\bm{\zeta}^{\star}\parallel\bm{\zeta}^{t}\big)+\frac{1+C_3}{15C_2\sqrt{C_1\log \frac{n}{\varepsilon}}}\log\big(1+2e^{-B}\big) \notag\\
	&\qquad\quad \le\dist\big(\bm{\zeta}^{\star}\parallel\bm{\zeta}^{t}\big)+e^{-B}, 
	\label{eq:KL-gen-zeta-t-zeta-trunc}
\end{align}
}
%
where the third line results from the choice \eqref{eq:parameter}, 
the penultimate line relies on $ \sum_j c_j = 1$, and the last line makes use of $\log (1+x)\leq x$ for all $x\geq 0$ and holds if $C_2\sqrt{C_1}$ is sufficiently large (recall that $C_3\leq 1$). 
In a nutshell, \eqref{eq:KL-gen-zeta-t-zeta-trunc} indicates that the KL divergence between the optimal point and the $t$-th iterate is not increased by much when $\bm{\zeta}^t$ is replaced with $\bm{\zeta}^{t, \trunc}$.

%
%

The next step consists of establishing the following result that monitors the change of KL divergence when $\bm{\zeta}^{t, \trunc}$ is further replaced with $\bm{\zeta}^{t+1}$: 
{\small
\begin{align}
	\dist\big(\bm{\zeta}^{\star}\parallel \bm{\zeta}^{t+1}\big) \le (1-\eta)\dist\big(\bm{\zeta}^{\star}\parallel \bm{\zeta}^{t, \trunc}\big) + 8ne^{-B},  
	\qquad \forall t\geq 0 .\label{eq:recursion-bound}
\end{align}
}
Crucially, a contraction factor of $1-\eta$ appears in the above claim,  revealing the progress made per iteration. 
Suppose for the moment that this claim \eqref{eq:recursion-bound} is valid. Then taking it together with \eqref{eq:KL-gen-zeta-t-zeta-trunc} gives
{\small
\begin{align}
	\dist\big(\bm{\zeta}^{\star}\parallel \bm{\zeta}^{t+1}\big) \le (1-\eta)\dist\big(\bm{\zeta}^{\star}\parallel \bm{\zeta}^{t}\big) + (8n+1)e^{-B},   
	\qquad \forall t\geq 0 .
\end{align}
}
Applying this relation recursively further implies that
{\small
\begin{align}
	\dist\big(\bm{\zeta}^{\star}\parallel\bm{\zeta}^{t_{\max}}\big) & \le(1-\eta)^{t_{\max}}\dist\big(\bm{\zeta}^{\star}\parallel\bm{\zeta}^{0}\big)+(8n+1)e^{-B} \sum_{t=0}^{t_{\max}-1}(1-\eta)^{t} \notag\\
 	& \leq  (1-\eta)^{t_{\max}}\dist\big(\bm{\zeta}^{\star}\parallel\bm{\zeta}^{0}\big)+\frac{(8n+1) e^{-B}}{\eta} .  
	\label{eq:KL-zetastar-zeta-tmax-pre}
\end{align}
}
Regarding the first term in \eqref{eq:KL-zetastar-zeta-tmax-pre}, it is observed from our initialization (cf.~line~\ref{line:initialization-main} of Algorithm~\ref{alg:main}) that
{\small
\[
\begin{cases}
\mathsf{KL}\big(\bm{p}_{i}^{\star,\mathsf{reg}}\parallel\bm{p}_{i}^{0}\big) & =-\mathcal{H}\big(\bm{p}_{i}^{\star,\mathsf{reg}}\big)+\sum_{j=1}^{n}p_{i,j}^{\star,\mathsf{reg}}\log\frac{1}{p_{i,j}^{0}}\leq\sum_{j=1}^{n}p_{i,j}^{\star,\mathsf{reg}}\log n=\log n,\\
\mathsf{KL}\big(\bm{\mu}_{j}^{\star,\mathsf{reg}}\parallel\bm{\mu}_{j}^{0}\big) & =-\mathcal{H}\big(\bm{\mu}_{j}^{\star,\mathsf{reg}}\big)+\sum_{s\in\{+,-\}}\mu_{j,s}^{\star,\mathsf{reg}}\log\frac{1}{\mu_{j,s}^{0}}\leq\sum_{s\in\{+,-\}}\mu_{j,s}^{\star,\mathsf{reg}}\log2=\log2,
\end{cases}
\]
}
and consequently,
{\small
\begin{align*}
 \dist\big(\bm{\zeta}^{\star}\parallel\bm{\zeta}^{0}\big)&\le\sum_{i=1}^{n}\frac{\log n}{\eta_{p,i}}+\sum_{j=1}^{n}\frac{\log2}{\eta_{\mu,j}}\\
 &=\frac{\sqrt{B}\log n}{C_{2}}\sum_{i=1}^{n}r_{i}+\frac{\log2}{15C_{2}\sqrt{B}}\sum_{j=1}^{n}\Big(c_{j}+\frac{C_3}{n}\Big)\le\frac{2\sqrt{B}\log n}{C_{2}}, 
\end{align*}
}
where we recall that $0<C_3\leq 1$. 
This in turn leads to $(1-\eta)^{t_{\max}}\dist\big(\bm{\zeta}^{\star}\parallel\bm{\zeta}^{0}\big)\leq \varepsilon^2 $, provided that  $t_{\max} \geq C_4 \frac{1}{\eta}\log\frac{n}{\varepsilon}$ for some large enough constant $C_4>0$. 
Turning to the second term in \eqref{eq:KL-zetastar-zeta-tmax-pre}, one utilizes $B = C_1\log\frac{n}{\varepsilon}$ to obtain
{\small
\begin{align*}
	\frac{(8n+1) e^{-B}}{\eta} &= \frac{(8n+1)e^{-B}\sqrt{B}\log n}{C_2^2\varepsilon} \le \frac{(8n+1)e^{-\frac{1}{2}B}\log n}{C_2^2\varepsilon} \\
	&= \frac{(8n+1)e^{-\frac{1}{2}C_1 \log \frac{n}{\varepsilon}}\log n}{C_2^2\varepsilon}
	\le \varepsilon^2, 
\end{align*}
}
with the proviso that $C_1>0$ is sufficiently large. Substitution into \eqref{eq:KL-zetastar-zeta-tmax-pre} thus yields
%
\begin{align}
	\dist\big(\bm{\zeta}^{\star}\parallel\bm{\zeta}^{t_{\max}}\big) & \le \varepsilon^2 + \varepsilon^2 = 2\varepsilon^2.  
	\label{eq:KL-zetastar-zeta-tmax-final}
\end{align}
%

As it turns out, it is more convenient to work with the $\ell_1$-based error. 
To convert the above bound on KL divergence into $\ell_1$-based distance, one invokes Pinsker's inequality \cite[Lemma 2.5]{tsybakov2004introduction} to reach 
{\small
\begin{align}
	\big\| \widehat{\bm{P}}-\bm{P}^{\star,\mathsf{reg}}\big\|_{1} & =\sum\nolimits_{i=1}^{n}r_{i}\big\|\bm{p}_{i}^{\star,\mathsf{reg}}-\bm{p}_{i}^{\star,t_{\max}}\big\|_{1}\leq\sum\nolimits_{i=1}^{n}r_{i}\sqrt{2\mathsf{KL}\big(\bm{p}_{i}^{\star,\mathsf{reg}}\parallel\bm{p}_{i}^{t_{\max}}\big)} \notag\\
 & \leq\left\{ \sum\nolimits_{i=1}^{n}r_{i}\right\} ^{\frac{1}{2}}\left\{ 2\sum\nolimits_{i=1}^{n}r_{i}\mathsf{KL}\big(\bm{p}_{i}^{\star,\mathsf{reg}}\parallel\bm{p}_{i}^{t_{\max}}\big)\right\} ^{\frac{1}{2}} \notag\\
 & =\left\{ 2\sum\nolimits_{i=1}^{n}r_{i}\mathsf{KL}\big(\bm{p}_{i}^{\star,\mathsf{reg}}\parallel\bm{p}_{i}^{t_{\max}}\big)\right\} ^{\frac{1}{2}} \notag\\
 & =\bigg\{ \frac{2C_{2}}{\sqrt{B}}\sum\nolimits_{i=1}^{n}\frac{1}{\eta_{p,i}}\mathsf{KL}\big(\bm{p}_{i}^{\star,\mathsf{reg}}\parallel\bm{p}_{i}^{t_{\max}}\big)\bigg\} ^{\frac{1}{2}} \notag\\
	& \leq\bigg\{ \frac{2C_{2}}{\sqrt{C_1 \log\frac{n}{\varepsilon}}}\dist\big(\bm{\zeta}^{\star}\parallel \bm{\zeta}^{t_{\max}}\big) \bigg\} ^{\frac{1}{2}}\leq \frac{\varepsilon}{6}. 
	\label{eq:P-Preg-L1}
\end{align}
}
Here, the first inequality invokes Pinsker's inequality, the second inequality results from  Cauchy-Schwarz, 
the third line holds since $\sum_i r_i=1$, the fourth line uses the choice \eqref{eqn:learning-rates-eta} of $\eta_{p,i}$, 
whereas the last line results from the definition \eqref{eq:defn-KL-gen} and the bound \eqref{eq:KL-zetastar-zeta-tmax-final} and is valid as long as $C_2$ (resp.~$C_1$) is sufficiently small (resp.~large).

Thus far, we have demonstrated fast convergence of our algorithm to the solution to the entropy-regularized problem \eqref{eq:game-entropy}.  
To finish up, we still need to show the proximity of the objective values under the regularized solution $\bm{P}^{\star,\mathsf{reg}}$ and under the true optimizer $\bm{P}^{\star}$ of \eqref{eq:original}. 
For notational convenience, define  
\begin{equation}
	\overline{\bm{\mu}}_j^{\star} \coloneqq \arg\max_{\bm{\mu}_j\in \Delta_2} f\big(\{\bm{p}_{i}^{\star,\mathsf{reg}}\}_{i=1}^{n},\{\bm{\mu}_{j}\}_{i=1}^{n}\big) ,
	\qquad 1\leq j\leq n. 
	\label{eq:defn-mu-bar-star}
\end{equation}
 Given that $\sum_{j=1}^{n}\tau_{\mu,j}\log2+\sum_{i=1}^{n}\tau_{p,i}\log n\leq \varepsilon/4$ under our choice \eqref{eq:equiv-tau-mup} (see \eqref{eq:sum-tau-UB1} as long as $C_2/C_1$ is sufficiently small),  
we obtain 
{\small
\begin{align}
\frac{1}{2}\langle\bm{W},\bm{P}^{\star,\mathsf{reg}}\rangle+\big\|\bm{P}^{\star,\mathsf{reg}}\bm{1}-\bm{c}\big\|_{1} 
& =f\big(\{\bm{p}_{i}^{\star,\mathsf{reg}}\}_{i=1}^{n},\{\overline{\bm{\mu}}_{j}^{\star}\}_{i=1}^{n}\big) \notag\\
 & \leq F\big(\{\bm{p}_{i}^{\star,\mathsf{reg}}\}_{i=1}^{n},\{\overline{\bm{\mu}}_{j}^{\star}\}_{i=1}^{n}\big)+\sum\nolimits_{i=1}^{n}\tau_{p,i}\log n \notag\\
 & \leq F\big(\{\bm{p}_{i}^{\star,\mathsf{reg}}\}_{i=1}^{n},\{\bm{\mu}_{j}^{\star,\mathsf{reg}}\}_{i=1}^{n}\big)+\sum\nolimits_{i=1}^{n}\tau_{p,i}\log n \notag\\
 & \leq F\big(\{\bm{p}_{i}^{\star}\}_{i=1}^{n},\{\bm{\mu}_{j}^{\star,\mathsf{reg}}\}_{i=1}^{n}\big)+\sum\nolimits_{i=1}^{n}\tau_{p,i}\log n \notag\\
 & \leq f\big(\{\bm{p}_{i}^{\star}\}_{i=1}^{n},\{\bm{\mu}_{j}^{\star,\mathsf{reg}}\}_{i=1}^{n}\big)+\sum\nolimits_{i=1}^{n}\tau_{p,i}\log n+\sum\nolimits_{j=1}^{n}\tau_{\mu,j}\log2 \notag\\
 & \leq\max_{\bm{\mu}_{j}\in\Delta_{2},\forall j}f\big(\{\bm{p}_{i}^{\star}\}_{i=1}^{n},\{\bm{\mu}_{j}\}_{i=1}^{n}\big)+\sum_{i=1}^{n}\tau_{p,i}\log n+\sum_{j=1}^{n}\tau_{\mu,j}\log2 \notag\\
	& = \tfrac{1}{2} \langle\bm{W},\bm{P}^{\star}\rangle+ \Big\| \sum\nolimits_{i=1}^n r_i \bm{p}_i^{\star} - \bm{c} \Big\|_1 +\sum\nolimits_{i=1}^{n}\tau_{p,i}\log n+\sum\nolimits_{j=1}^{n}\tau_{\mu,j}\log2 \notag\\
	& \leq \tfrac{1}{2} \langle\bm{W},\bm{P}^{\star}\rangle+ \varepsilon/4. 
	\label{eq:translate-f-F}
\end{align}
}
Here, the first identity comes from \eqref{eq:equiv-objective-f-246} and the definition of $\overline{\bm{\mu}}_j^{\star}$, 
the second and the fifth lines are valid since $0\leq \mathcal{H}(\bm{p}_i)\leq \log n$ for any $\bm{p}_i\in \Delta_n$  
and $0\leq \mathcal{H}(\bm{\mu}_j)\leq \log 2$ for any $\bm{\mu}_j\in \Delta_2$, 
the third and the fourth lines hold since $\big( \{\bm{p}_{i}^{\star,\mathsf{reg}}\}_{i=1}^{n},\{\bm{\mu}_{j}^{\star,\mathsf{reg}}\}_{i=1}^{n} \big)$ corresponds to the minimax solution of $F(\cdot,\cdot)$, 
the penultimate line arises from \eqref{eq:equiv-objective-f-246}, 
while the last line results from the fact $\sum_i r_i \bm{p}_i^{\star} = \bm{c}$ and the assumption $\sum_{j=1}^{n}\tau_{\mu,j}\log2+\sum_{i=1}^{n}\tau_{p,i}\log n\leq \varepsilon/4$. 
As a consequence, we are ready to conclude 
{\small
\begin{align*}
\langle\bm{W},\widetilde{\bm{P}}\rangle & \leq\langle\bm{W},\widehat{\bm{P}}\rangle+\|\bm{W}\|_{\infty}\|\widehat{\bm{P}}-\widetilde{\bm{P}}\|_{1}
  \leq\langle\bm{W},\widehat{\bm{P}}\rangle+2\big\|\widehat{\bm{P}}\bm{1}-\bm{c}\big\|_{1}\\
 & \leq\langle\bm{W},\bm{P}^{\star,\mathsf{reg}}\rangle+\|\bm{W}\|_{\infty}\big\|\widehat{\bm{P}}-\bm{P}^{\star,\mathsf{reg}}\big\|_{1}+2\big\|\bm{P}^{\star,\mathsf{reg}}\bm{1}-\bm{c}\big\|_{1}+2\big\|\widehat{\bm{P}}-\bm{P}^{\star,\mathsf{reg}}\big\|_{1} \|\bm{1}\|_{\infty}\\
 & =\langle\bm{W},\bm{P}^{\star,\mathsf{reg}}\rangle+2\big\|\bm{P}^{\star,\mathsf{reg}}\bm{1}-\bm{c}\big\|_{1}+3\big\|\widehat{\bm{P}}-\bm{P}^{\star,\mathsf{reg}}\big\|_{1}\\
 & \leq\langle\bm{W},\bm{P}^{\star}\rangle+\varepsilon/2+\varepsilon/2=\langle\bm{W},\bm{P}^{\star}\rangle+\varepsilon, 
\end{align*}
}
where the second inequality invokes Lemma~\ref{lem:projection},  the assumption $\|\bm{W}\|_{\infty}=1$, and the fact $\widehat{\bm{P}}\bm{1}=\bm{r}$, 
and the last inequality arises from \eqref{eq:P-Preg-L1} and \eqref{eq:translate-f-F}.
This establishes the advertised result in Theorem~\ref{thm:theory-OT}.

The remainder of the proof is thus dedicated to establishing the claim~\eqref{eq:recursion-bound}.

\subsection{Proof of Claim~\eqref{eq:recursion-bound}}

\subsubsection{Step 1: decomposing the KL divergence of interest}
Elementary calculation together with the definition \eqref{eq:defn-KL-gen} of $\dist(\cdot\parallel\cdot)$ reveals that
{\small
\begin{align}
 & (1-\eta)\dist\big(\bm{\zeta}^{\star}\parallel\bm{\zeta}^{t,\trunc}\big)-(1-\eta)\dist\big(\overline{\bm{\zeta}}^{t+1}\parallel\bm{\zeta}^{t,\trunc}\big)-\eta\dist\big(\overline{\bm{\zeta}}^{t+1}\parallel\bm{\zeta}^{\star}\big)-\dist\big(\bm{\zeta}^{t+1}\parallel\overline{\bm{\zeta}}^{t+1}\big)\nonumber \\
 & \qquad+\big\langle\overline{\bm{\zeta}}^{t+1}-\bm{\zeta}^{t+1},\, \log\overline{\bm{\zeta}}^{t+1}-\log\bm{\zeta}^{t+1}\big\rangle
	+ \big\langle \overline{\bm{\zeta}}^{t+1}-\bm{\zeta}^{\star}, \, \log\bm{\zeta}^{t+1}-(1-\eta)\log\bm{\zeta}^{t,\trunc}-\eta\log\bm{\zeta}^{\star} \big\rangle\nonumber \\
 & =(1-\eta)\big\langle\bm{\zeta}^{\star},\,\log\bm{\zeta}^{\star}-\log\bm{\zeta}^{t,\trunc}\big\rangle-(1-\eta)\big\langle\overline{\bm{\zeta}}^{t+1},\,\log\overline{\bm{\zeta}}^{t+1}-\log\bm{\zeta}^{t,\trunc}\big\rangle\nonumber \\
 & \qquad-\eta\big\langle\overline{\bm{\zeta}}^{t+1},\,\log\overline{\bm{\zeta}}^{t+1}-\log\bm{\zeta}^{\star}\big\rangle-\big\langle\bm{\zeta}^{t+1},\,\log\bm{\zeta}^{t+1}-\log\overline{\bm{\zeta}}^{t+1}\big\rangle\nonumber \\
 & \qquad+\big\langle\overline{\bm{\zeta}}^{t+1}-\bm{\zeta}^{t+1},\,\log\overline{\bm{\zeta}}^{t+1}-\log\bm{\zeta}^{t+1}\big\rangle+\big\langle\overline{\bm{\zeta}}^{t+1}-\bm{\zeta}^{\star},\,\log\bm{\zeta}^{t+1}-(1-\eta)\log\bm{\zeta}^{t,\trunc}-\eta\log\bm{\zeta}^{\star}\big\rangle\nonumber \\
 & =\big\langle\bm{\zeta}^{\star},\,\log\bm{\zeta}^{\star}\big\rangle-\big\langle\bm{\zeta}^{\star},\,\log\bm{\zeta}^{t+1}\big\rangle=\dist\big(\bm{\zeta}^{\star}\parallel\bm{\zeta}^{t+1}\big); 
	\label{eq:KL-ge-long-elementary}
\end{align}
}
here and throughout, the logarithmic operator in $\log \bm{\zeta}$ is applied in an entrywise manner. 
In addition, inspired by \cite[Lemma 1]{cen2021fast}, we observe that
{\small
\begin{align} \label{eq:identity}
	\big\langle \overline{\bm{\zeta}}^{t+1} - \bm{\zeta}^{\star}, \, \log \bm{\zeta}^{t+1} - (1-\eta)\log \bm{\zeta}^{t, \trunc} - \eta\log \bm{\zeta}^{\star}\big\rangle = 0; 
\end{align}
}
see Appendix~\ref{sec:proof-identity} for the proof of this relation.   
Substitution into \eqref{eq:KL-ge-long-elementary} leads to
{\small
\begin{align}
\dist\big(\bm{\zeta}^{\star}\parallel \bm{\zeta}^{t+1}\big) &= (1-\eta)\dist\big(\bm{\zeta}^{\star}\parallel \bm{\zeta}^{t, \trunc}\big) - (1-\eta)\dist\big(\overline{\bm{\zeta}}^{t+1}\parallel \bm{\zeta}^{t, \trunc}\big) - \eta \dist\big(\overline{\bm{\zeta}}^{t+1}\parallel \bm{\zeta}^{\star}\big) \notag\\
&\qquad\qquad	- \dist\big(\bm{\zeta}^{t+1}\parallel \overline{\bm{\zeta}}^{t+1}\big) 
+ \big\langle \overline{\bm{\zeta}}^{t+1} - \bm{\zeta}^{t+1}, \, \log \overline{\bm{\zeta}}^{t+1} - \log \bm{\zeta}^{t+1} \big\rangle. 
	\label{eq:recursion}
\end{align}
}
It then boils down to controlling the term 
$\big\langle \overline{\bm{\zeta}}^{t+1} - \bm{\zeta}^{t+1}, \, \log \overline{\bm{\zeta}}^{t+1} - \log \bm{\zeta}^{t+1} \big\rangle$.  

%
Towards this end, 
we first invoke the update rules~\eqref{eq:update-midpoints} and~\eqref{eq:update-main} to yield
{\small
\begin{align}
\big\langle \overline{\bm{\zeta}}^{t+1} - \bm{\zeta}^{t+1}, \, \log \overline{\bm{\zeta}}^{t+1} - \log \bm{\zeta}^{t+1} \big\rangle 
	&= \sum\nolimits_{j = 1}^n\big(\overline{\mu}_{j, +}^{t+1} - \overline{\mu}_{j, -}^{t+1} - \mu_{j, +}^{t+1} + \mu_{j, -}^{t+1}\big) \Big\{ \sum\nolimits_{i = 1}^n r_i\big(p_{i,j}^{t} - \overline{p}_{i,j}^{t+1}\big) \Big\} \nonumber\\
	&\quad+ \sum\nolimits_{j = 1}^n\big(\overline{\mu}_{j, +}^{t+1} - \overline{\mu}_{j, -}^{t+1} - \mu_{j, +}^{t, \trunc} + \mu_{j, -}^{t, \trunc}\big) \Big\{ \sum\nolimits_{i = 1}^nr_i\big(\overline{p}_{i,j}^{t+1} - p_{i,j}^{t+1}\big) \Big\},  
	\label{eq:prediction-error}
\end{align}
}
whose proof is also deferred to Appendix~\ref{sec:proof-identity}. 
To bound the two sums on the right-hand side of \eqref{eq:prediction-error}, 
we find it convenient to first introduce the following index subset: 
{\small
\begin{align}
	\mathcal{J}_t \coloneqq \bigg\{j : \sum_{i = 1}^n r_ip_{i,j}^{t} < 2e \Big(c_j + \frac{1}{n} \Big)\bigg\}.
	\label{eq:defn-Jt-1}
\end{align}
}
We then divide each sum into two parts --- $\big\{ j: j \in \mathcal{J}_t \big\}$ and $\big\{j: j \notin \mathcal{J}_t \big\}$ --- and look at them separately.

\subsubsection{Step 2: controlling terms with $j \in \mathcal{J}_t$}
We start by looking at those terms with $j \in \mathcal{J}_t$. 
Regarding the first sum on the right-hand side of~\eqref{eq:prediction-error}, 
we have
{\small
\begin{align}
 & \sum_{j\in\mathcal{J}_{t}}\big(\overline{\mu}_{j,+}^{t+1}-\overline{\mu}_{j,-}^{t+1}-\mu_{j,+}^{t+1}+\mu_{j,-}^{t+1}\big)\bigg\{\sum_{i=1}^{n}r_{i}\big(p_{i,j}^{t}-\overline{p}_{i,j}^{t+1}\big)\bigg\}\notag\\
 & \qquad\le\sum_{j\in\mathcal{J}_{t}}\frac{1}{2\eta_{\mu,j}}\big(\overline{\mu}_{j,+}^{t+1}-\overline{\mu}_{j,-}^{t+1}-\mu_{j,+}^{t+1}+\mu_{j,-}^{t+1}\big)^{2}+\sum_{j\in\mathcal{J}_{t}}\frac{\eta_{\mu,j}}{2}\bigg\{\sum_{i=1}^{n}r_{i}\big(p_{i,j}^{t}-\overline{p}_{i,j}^{t+1}\big)\bigg\}^{2}. 
	\label{eq:sum-Jt-two-terms}
\end{align}
}
Recognizing that $\overline{\bm{\mu}}_{j}^{t+1}, \bm{\mu}_{j}^{t+1} \in \Delta_2$, 
one can invoke Pinsker's inequality \cite[Lemma 2.5]{tsybakov2004introduction} to bound the first term on the right-hand side of \eqref{eq:sum-Jt-two-terms} as follows: 
{\small
\begin{align}
	\big(\overline{\mu}_{j, +}^{t+1} - \overline{\mu}_{j, -}^{t+1} - \mu_{j, +}^{t+1} + \mu_{j, -}^{t+1}\big)^2 
	& = 
	\big\|\overline{\bm{\mu}}_{j}^{t+1} - \bm{\mu}_{j}^{t+1}\big\|_1^2  
	\le 
	2\mathsf{KL}\big(\bm{\mu}_{j}^{t+1}\parallel\overline{\bm{\mu}}_{j}^{t+1} \big).
	\label{eq:first-term-mu-13579}
\end{align}
}
Here, the first identity holds since, according to the properties that $\overline{\bm{\mu}}_{j}^{t+1},\bm{\mu}_{j}^{t+1}\in \Delta_2$, 
\begin{align*}
\Big|\overline{\mu}_{j,+}^{t+1}-\overline{\mu}_{j,-}^{t+1}-\mu_{j,+}^{t+1}+\mu_{j,-}^{t+1}\Big| & =\Big|\overline{\mu}_{j,+}^{t+1}+\big(1-\overline{\mu}_{j,-}^{t+1}\big)-\mu_{j,+}^{t+1}-\big(1-\mu_{j,-}^{t+1}\big)\Big|\\
 & =\Big|2\overline{\mu}_{j,+}^{t+1}-2\mu_{j,+}^{t+1}\Big|=\Big|\overline{\mu}_{j,+}^{t+1}-\mu_{j,+}^{t+1}\Big|+\Big|1-\overline{\mu}_{j,+}^{t+1}-1+\mu_{j,+}^{t+1}\Big|\\
 & =\Big|\overline{\mu}_{j,+}^{t+1}-\mu_{j,+}^{t+1}\Big|+\Big|\overline{\mu}_{j,-}^{t+1}-\mu_{j,-}^{t+1}\Big|
	= \big\|\overline{\bm{\mu}}_{j}^{t+1} - \bm{\mu}_{j}^{t+1}\big\|_1.
\end{align*}
Regarding the second term on the right-hand side of \eqref{eq:sum-Jt-two-terms}, we see that, for $\eta < 1/2$, 
{\small
\begin{align}
	& \sum_{i} (1-\eta)\frac{1}{\eta_{p,i}}\KL\big(\overline{\bm{p}}_i^{t+1}\parallel \bm{p}_i^{t}\big) \ge \frac{1}{2}\sum_{i = 1}^n \sum_{j = 1}^n \frac{\overline{p}_{i,j}^{t+1}}{\eta_{p,i}}\log \frac{\overline{p}_{i,j}^{t+1}}{p_{i,j}^{t}} \\
&\qquad\ge \frac{1}{2\max_i \eta_{p,i} r_i} \sum_{i = 1}^n \sum_{j = 1}^n r_ip_{i,j}^{t}\big((1 + x_{ij})\log (1 + x_{i,j}) \big) \notag\\
	& \qquad \overset{\mathrm{(i)}}{=} \frac{1}{2\max_i \eta_{p,i} r_i} \sum_{j = 1}^n \sum_{i = 1}^n  r_ip_{i,j}^{t}\big((1 + x_{ij})\log (1 + x_{i,j}) - x_{i,j}\big) \notag\\
&\qquad\overset{\mathrm{(ii)}}{\ge } \frac{3}{4\max_i \eta_{p,i} r_i} \sum_{j = 1}^n \frac{1}{2\sum_{i = 1}^n r_ip_{i,j}^{t} + \sum_{i = 1}^n r_i\overline{p}_{i,j}^{t+1}}\bigg\{\sum_{i = 1}^n r_i\big(p_{i,j}^{t} - \overline{p}_{i,j}^{t+1}\big)\bigg\}^2 \notag\\
	&\qquad  \overset{\mathrm{(iii)}}{\ge } \frac{1}{20e \big(\max_i \eta_{p,i} r_i \big) \big( \max_j \eta_{\mu,j} (c_j + 1/n) \big)} \sum_{j \in \mathcal{J}_t} \eta_{\mu,j} \bigg\{\sum_{i = 1}^n r_i\big(p_{i,j}^{t} - \overline{p}_{i,j}^{t+1}\big)\bigg\}^2, 
	\label{eq:KL-lb-13579}
\end{align}
}
where we define $x_{i,j} \coloneqq \frac{\overline{p}_{i,j}^{t+1}}{p_{i,j}^{t}} - 1$.  
Here, (i) is valid due to $\sum_{j}p_{i,j}^{t}x_{i,j}=\sum_{j}(\overline{p}_{i,j}^{t}-p_{i,j}^{t})=0$;    
(ii) arises since 
{\small
\begin{align*}
 & \sum_{i=1}^{n}r_{i}p_{i,j}^{t}\big((1+x_{i,j})\log(1+x_{i,j})-x_{i,j}\big)\geq\frac{1}{2}\sum_{i=1}^{n}r_{i}p_{i,j}^{t}\frac{x_{i,j}^{2}}{1+x_{i,j}/3}=\frac{1}{2}\sum_{i=1}^{n}\frac{(r_{i}p_{i,j}^{t})^{2}x_{i,j}^{2}}{r_{i}p_{i,j}^{t}(1+x_{i,j}/3)}\\
 & \quad\ge\frac{1}{2}\frac{1}{\sum_{i=1}^{n}r_{i}p_{i,j}^{t}(1+x_{i,j}/3)}\Big(\sum_{i=1}^{n}r_{i}p_{i,j}^{t}x_{i,j}\Big)^{2}=\frac{3}{4\sum_{i=1}^{n}r_{i}p_{i,j}^{t}+2\sum_{i=1}^{n}r_{i}\overline{p}_{i,j}^{t+1}}\Big(\sum_{i=1}^{n}r_{i}\big(\overline{p}_{i,j}^{t+1}-p_{i,j}^{t}\big)\Big)^{2},	
\end{align*}
}
where the first inequality holds since $(1+z)\log (1+z) - z \geq \frac{z^2}{2(1+z/3)}$ for all $z\geq -1$, 
and the second line follows from Sedrakyan's inequality \cite[Chapter 8, Lemma 1]{sedrakyan2018algebraic}; 
(iii) comes from the definition \eqref{eq:defn-Jt-1} of $\mathcal{J}_t$, as well as the following claim (which will be proved momentarily): 
{\small
\begin{align}\label{eq:p-bound}
	\sum\nolimits_{i = 1}^n r_i\overline{p}_{i,j}^{t+1} < 5e \big( c_j + 1/n \big)
\qquad\text{and}\qquad
	\sum\nolimits_{i = 1}^n r_ip_{i,j}^{t+1} < 5e \big(c_j + 1/n \big), 
	\quad \forall j\in \mathcal{J}_t.
\end{align}  
}
Substituting \eqref{eq:first-term-mu-13579} and \eqref{eq:KL-lb-13579} into \eqref{eq:sum-Jt-two-terms} yields
{\small
\begin{align}
 & \sum_{j\in\mathcal{J}_{t}}\big(\overline{\mu}_{j,+}^{t+1}-\overline{\mu}_{j,-}^{t+1}-\mu_{j,+}^{t+1}+\mu_{j,-}^{t+1}\big)\bigg\{\sum\nolimits_{i=1}^{n}r_{i}\big(p_{i,j}^{t}-\overline{p}_{i,j}^{t+1}\big)\bigg\}\notag\nonumber \\
 & \le\sum_{j\in\mathcal{J}_{t}}\frac{1}{\eta_{\mu,j}}\mathsf{KL}\big(\bm{\mu}_{j}^{t+1}\parallel\overline{\bm{\mu}}_{j}^{t+1}\big)+(1-\eta)40e\Big(\max_{i}\eta_{p,i}r_{i}\Big)\Big(\max_{j}\eta_{\mu,j}(c_{j}+1/n)\Big)\sum_{i}\frac{1}{\eta_{p,i}}\mathsf{KL}\big(\overline{\bm{p}}_{i}^{t+1}\parallel\bm{p}_{i}^{t}\big) \notag\\
 & \le\sum_{j\in\mathcal{J}_{t}}\frac{1}{\eta_{\mu,j}}\mathsf{KL}\big(\bm{\mu}_{j}^{t+1}\parallel\overline{\bm{\mu}}_{j}^{t+1}\big)+(1-\eta)\sum_{i}\frac{1}{\eta_{p,i}}\mathsf{KL}\big(\overline{\bm{p}}_{i}^{t+1}\parallel\bm{p}_{i}^{t}\big), 
	\label{eq:sum-Jt-first-term-179}
\end{align}
}
provided that (using the choice \eqref{eq:parameter} and the assumption $0<C_3\leq 1$)
{\small
\begin{align}
	\Big(\max_{i}\eta_{p,i}r_{i}\Big)\Big(\max_{j}\eta_{\mu,j}(c_{j}+1/n)\Big)
	\leq \frac{1}{C_3} \Big(\max_{i}\eta_{p,i}r_{i}\Big)\Big(\max_{j}\eta_{\mu,j}(c_{j}+C_3/n)\Big) =\frac{15C_{2}^{2}}{C_3}\le\frac{1}{40e}.
	\label{eq:max-eta-min-eta-condition}
\end{align}
}

We then move on to the second sum on the right-hand side of \eqref{eq:prediction-error} when restricted to $\mathcal{J}_t$. 
Repeating the arguments as above, we can guarantee that
{\small
\begin{align*}
 & \sum_{j\in\mathcal{J}_{t}}\big(\overline{\mu}_{j,+}^{t+1}-\overline{\mu}_{j,-}^{t+1}-\mu_{j,+}^{t,\trunc}+\mu_{j,-}^{t,\trunc}\big)\sum_{i=1}^{n}r_{i}\big(\overline{p}_{i,j}^{t+1}-p_{i,j}^{t+1}\big)\\
 & \qquad\le\sum_{j\in\mathcal{J}_{t}}\frac{1-\eta}{2\eta_{\mu,j}}\big(\overline{\mu}_{j,+}^{t+1}-\overline{\mu}_{j,-}^{t+1}-\mu_{j,+}^{t,\trunc}+\mu_{j,-}^{t,\trunc}\big)^{2}+\frac{1}{2(1-\eta)}\sum_{j\in\mathcal{J}_{t}}\eta_{\mu,j}\bigg\{\sum_{i=1}^{n}r_{i}\big(\overline{p}_{i,j}^{t+1}-p_{i,j}^{t+1}\big)\bigg\}^{2}\\
 & \qquad\le\sum_{j}\frac{1-\eta}{\eta_{\mu,j}}\mathsf{KL}\big(\overline{\bm{\mu}}_{j}^{t+1}\parallel\bm{\mu}_{j}^{t,\trunc}\big)+\sum_{i}\frac{1}{\eta_{p,i}}\KL\big(\bm{p}_{i}^{t+1}\parallel\overline{\bm{p}}_{i}^{t+1}\big), 
\end{align*}
}
%
as long as $0<\eta < 1/2$ (so that $\frac{1}{1-\eta}\leq 2$) and 
Condition~\eqref{eq:max-eta-min-eta-condition} is met; the details are omitted here for brevity.  
Combining this result with \eqref{eq:sum-Jt-first-term-179}, we reach
{\small
\begin{align}
&\sum_{j \in \mathcal{J}_t}\big(\overline{\mu}_{j, +}^{t+1} - \overline{\mu}_{j, -}^{t+1} - \mu_{j, +}^{t+1} + \mu_{j, -}^{t+1}\big)\sum_{i = 1}^n r_i\big(p_{i,j}^{t} - \overline{p}_{i,j}^{t+1}\big) + \sum_{j \in \mathcal{J}_t} \big(\overline{\mu}_{j, +}^{t+1} - \overline{\mu}_{j, -}^{t+1} - \mu_{j, +}^{t, \trunc} + \mu_{j, -}^{t, \trunc}\big)\sum_{i = 1}^nr_i\big(\overline{p}_{i,j}^{t+1} - p_{i,j}^{t+1}\big) \nonumber\\
	&\qquad\qquad\le (1-\eta)\dist\big(\overline{\bm{\zeta}}^{t+1}\parallel \bm{\zeta}^{t, \trunc}\big) + \dist\big(\bm{\zeta}^{t+1}\parallel \overline{\bm{\zeta}}^{t+1}\big). 
	\label{eq:step2-final}
\end{align}
}

\paragraph{Proof of Claim~\eqref{eq:p-bound}}
The analyses for $\overline{p}_{i,j}^{t+1}$ and for $p_{i,j}^{t+1}$ are essentially the same; 
we shall thus only present how to establish the first inequality in \eqref{eq:p-bound} for the sake of brevity. 
According to the update rule~\eqref{eq:update-midpoints}, 
{\small
\begin{align}
	\overline{p}_{i,j}^{t+1} &= \frac{\big(p_{i,j}^{t}\big)^{1-\eta}\exp\Big(-\eta_{p,i}r_{i}\big(0.5w_{i,j}+\mu_{j,+}^{t,\trunc}-\mu_{j,-}^{t,\trunc}\big)\Big)}{\sum_{k = 1}^n \big(p_{i,k}^{t}\big)^{1-\eta}\exp\Big(-\eta_{p,i}r_{i}\big(0.5w_{i,k}+\mu_{k,+}^{t,\trunc}-\mu_{k,-}^{t,\trunc}\big)\Big)}. \label{eq:difference-p-pre}
\end{align}
}
Given that $\|\bm{W}\|_{\infty}=1$, $|\mu_{j,+}^{t,\trunc}-\mu_{j,-}^{t,\trunc} | \leq 1$ and $0<\eta<1$, we can bound
{\small
\begin{subequations}
\label{eq:difference-p-temp}
	{\small
\begin{align}
	&\big(p_{i,j}^{t}\big)^{1-\eta}\exp\Big(-\eta_{p,i}r_{i}\big(0.5w_{i,j}+\mu_{j,+}^{t,\trunc}-\mu_{j,-}^{t,\trunc}\big)\Big) \ge p_{i,j}^{t}\exp\big(-1.5\eta_{p,i} r_i\big)  \\
%
%
	\text{and}\quad & \big(p_{i,j}^{t}\big)^{1-\eta}\exp\Big(-\eta_{p,i}r_{i}\big(0.5w_{i,j}+\mu_{j,+}^{t,\trunc}-\mu_{j,-}^{t,\trunc}\big)\Big)\leq\big(p_{i,j}^{t}\big)^{1-\eta}\exp\big(1.5\eta_{p,i}r_{i})\notag\\
 & \qquad\le\left\{ p_{i,j}^{t}e^{\eta B}\ind\{p_{i,j}^{t}>e^{-B}\}+e^{-(1-\eta)B}\ind\{p_{i,j}^{t}\leq e^{-B}\}\right\} \exp\big(1.5\eta_{p,i}r_{i})\notag\\
 & \qquad\leq p_{i,j}^{t}\exp\big(1.5\eta_{p,i}r_{i}+\eta B\big)+\exp\big(1.5\eta_{p,i}r_{i}-(1-\eta)B\big)
\end{align}
}
\end{subequations}
}
for every $1\leq i,j\leq n$. 
Substituting these two bounds into \eqref{eq:difference-p-pre} leads to
{\small
\begin{align}
\overline{p}_{i,j}^{t+1} 
&\le \frac{p_{i,j}^{t}\exp\big(1.5\eta_{p,i} r_i+\eta B\big) + \exp\big(-(1-\eta)B + 1.5\eta_{p,i} r_i\big)}{\sum_{k = 1}^n p_{i,k}^{t}\exp\big(-1.5\eta_{p,i} r_i\big)} \notag\\
	&= \big(p_{i,j}^{t} + e^{-B} \big) \exp\big(3\eta_{p,i} r_i+\eta B\big) \le 2\big(p_{i,j}^{t} + e^{-B}\big) \label{eq:difference-p}
\end{align}
}
for every $1\leq i,j\leq n$, with the proviso that $3\eta_{p,i} r_i+\eta B \le \log 2$ (which is satisfied under the choice \eqref{eq:parameter} if $C_2$ is small enough).  
Therefore, 
for any $j\in \mathcal{J}_t$, 
{\small
\begin{align}
	 \sum_{i = 1}^n r_i\overline{p}_{i,j}^{t+1} 
	\leq 2  \sum_{i = 1}^n r_i  p_{i,j}^{t}  + 2  \sum_{i = 1}^n r_i e^{-B} 
	< 4e\Big(c_j+ \frac{1}{n}\Big) + 2 e^{-B} 
	<  5e \Big(c_j+ \frac{1}{n}\Big),
\end{align}
}
where we make use of the definition \eqref{eq:defn-Jt-1} of $\mathcal{J}_t$ as well as the fact $\sum_i r_i = 1$, provided that $C_1$ is large enough.

\subsubsection{Step 3: controlling terms with $j \notin \mathcal{J}_t$}
We now move on to the terms with $j \notin \mathcal{J}_t$. 
Employing similar analysis as for~\eqref{eq:difference-p}, we derive 
{\small
\begin{align}
	p_{i,j}^{t+1} &=  \frac{\big(p_{i,j}^{t}\big)^{1-\eta}\exp\big(-\eta_{p,i}r_{i}(0.5w_{i,j}+\overline{\mu}_{j,+}^{t+1}-\overline{\mu}_{j,-}^{t+1})\big)}{\sum_{k = 1}^n \big(p_{i,k}^{t}\big)^{1-\eta}\exp\big(-\eta_{p,i}r_{i}(0.5w_{i,k}+\overline{\mu}_{k,+}^{t+1}-\overline{\mu}_{k,-}^{t+1})\big)} \notag\\
	&\leq \big(p_{i,j}^{t} + e^{-B}\big) \exp\big(3\eta_{p,i} r_i+\eta B\big) 
	\label{eq:pij-t-plus-UB}
\end{align}
}
for any $1\leq i,j\leq n$, which in turn implies that
{\small
\begin{align*}
\sum\nolimits_{i=1}^{n}r_{i}p_{i,j}^{t} & \geq\sum\nolimits_{i=1}^{n}r_{i}\left\{ \big[\exp\big(3\eta_{p,i}r_{i}+\eta B\big)\big]^{-1}p_{i,j}^{t+1}-e^{-B}\right\} \\
 & \geq\sum\nolimits_{i=1}^{n}r_{i}\Big[\exp\Big(\max_{i}3\eta_{p,i}r_{i}+\eta B\Big)\Big]^{-1}p_{i,j}^{t+1}-\sum\nolimits_{i=1}^{n}r_{i}e^{-B}\\
 & =\Big[\exp\Big(\max_{i}3\eta_{p,i}r_{i}+\eta B\Big)\Big]^{-1}\sum\nolimits_{i=1}^{n}r_{i}p_{i,j}^{t+1}-e^{-B}.
\end{align*}
}
Hence, for any $j \notin \mathcal{J}_t$ and any integer $0 < \tau < \frac{1}{3\max_i\eta_{p,i} r_i+\eta B}$, applying the above relation recursively yields
{\small
\begin{align}
\sum_{i=1}^{n}r_{i}p_{i,j}^{t-\tau} & \geq\Big[\exp\Big(\max_{i}3\eta_{p,i}r_{i}+\eta B\Big)\Big]^{-\tau}\sum_{i=1}^{n}r_{i}p_{i,j}^{t}-e^{-B}\sum_{k=0}^{\tau-1}\Big[\exp\Big(\max_{i}3\eta_{p,i}r_{i}+\eta B\Big)\Big]^{-k} \notag\\
	& >\Big[\exp\Big(\max_{i}3\eta_{p,i}r_{i}+\eta B\Big)\Big]^{-\tau}\sum_{i=1}^{n}r_{i}p_{i,j}^{t}-\frac{e^{-B}}{1-\big[\exp\big(\max_{i}3\eta_{p,i}r_{i}+\eta B\big)\big]^{-1}} \label{eq:sum-r-p-relation-179}\\
 & >e^{-1}\sum_{i=1}^{n}r_{i}p_{i,j}^{t}-\frac{e^{-B}}{1-e^{-\eta B}}=e^{-1}\sum_{i=1}^{n}r_{i}p_{i,j}^{t}-\frac{e^{-C_{1}\log\frac{n}{\varepsilon}}}{1-e^{-\frac{C_2^2\varepsilon\sqrt{C_{1}\log\frac{n}{\varepsilon}}}{\log n}}} \notag\\
 & >e^{-1}\sum_{i=1}^{n}r_{i}p_{i,j}^{t}-\frac{0.5}{n}\geq2c_{j}+ 1.5/n, 
	\label{eq:sum-r-p-lower-bound-179}
\end{align}
}
where the penultimate line relies on the condition $\tau < \frac{1}{3\max_i\eta_{p,i} r_i+\eta B}$,
and the last line uses the definition of $\mathcal{J}_t$ in~\eqref{eq:defn-Jt-1} and holds as long as $C_1$ is large enough. 

This lower bound \eqref{eq:sum-r-p-lower-bound-179} already leads to one important observation.  
For any $0 \leq t < \frac{1}{3\max_i\eta_{p,i} r_i+\eta B}$, taking $\tau=t$ leads to 
{\small
\begin{align}
\sum\nolimits_{i=1}^{n}r_{i}p_{i,j}^{0} = \sum\nolimits_{i=1}^{n}r_{i}p_{i,j}^{t-\tau}
	> 2c_{j}+ 1.5/n, 
\end{align}
}
which contradicts our initialization 
$
	\sum_{i=1}^{n}r_{i}p_{i,j}^{0}=\frac{1}{n}\sum_{i=1}^{n}r_{i}=\frac{1}{n}< 2c_j+\frac{1.5}{n}.
$
This essentially implies that 
{\small
\begin{equation}
	\big\{ j: j\notin \mathcal{J}_t \big\} = \emptyset, \qquad \text{for all }0 \leq t < \frac{1}{3\max_i\eta_{p,i} r_i+\eta B} .
	\label{eq:complement-Jt-emptyset}
\end{equation}
}
Thus, it suffices to focus on $t\geq \frac{1}{3\max_i\eta_{p,i} r_i+\eta B}$ when analyzing the sum over $j\notin \mathcal{J}_t$.


Next, recalling the bounds \eqref{eq:difference-p-temp}, one can derive, for every $1\leq i,j\leq n$, 
{\small
\begin{align*}
\overline{p}_{i,j}^{t+1-\tau} & \geq\frac{p_{i,j}^{t-\tau}\exp\big(-1.5\eta_{p,i}r_{i}\big)}{\sum_{k=1}^{n}p_{i,k}^{t-\tau}\exp\big(1.5\eta_{p,i}r_{i}+\eta B\big)+\exp\big(-(1-\eta)B+1.5\eta_{p,i}r_{i}\big)}\\
 & =\frac{p_{i,j}^{t-\tau}\exp\big(-1.5\eta_{p,i}r_{i}\big)}{\exp\big(1.5\eta_{p,i}r_{i}+\eta B\big)+\exp\big(-(1-\eta)B+1.5\eta_{p,i}r_{i}\big)}\\
 & \geq \frac{\big[\exp\big(3\max_{i}\eta_{p,i}r_{i}+\eta B\big)\big]^{-1}}{1+e^{-B}}p_{i,j}^{t-\tau}.
\end{align*}
}
We then further derive: for any $j\notin \mathcal{J}_t$ and any integer $0 < \tau < \frac{1}{6\max_i\eta_{p,i} r_i+2\eta B}$, 
{\small
\begin{align}
\sum_{i=1}^{n}r_{i}\overline{p}_{i,j}^{t+1-\tau} & \geq\frac{\big[\exp\big(3\max_{i}\eta_{p,i}r_{i}+\eta B\big)\big]^{-1}}{1+e^{-B}}\sum_{i=1}^{n}r_{i}p_{i,j}^{t-\tau} \notag\\
 & >\frac{\big[\exp\big(3\max_{i}\eta_{p,i}r_{i}+\eta B\big)\big]^{-\tau-1}}{1+e^{-B}}\sum_{i=1}^{n}r_{i}p_{i,j}^{t} \notag\\
 & \qquad\qquad-\frac{\big[\exp\big(3\max_{i}\eta_{p,i}r_{i}+\eta B\big)\big]^{-1}}{1+e^{-B}}\cdot\frac{e^{-B}}{1-\big[\exp\big(3\max_{i}\eta_{p,i}r_{i}+\eta B\big)\big]^{-1}} \notag\\
 & >e^{-1}\sum_{i=1}^{n}r_{i}p_{i,j}^{t}-\frac{1}{1+e^{-B}}\cdot\frac{e^{-B}}{\exp\big(\frac{3C_{2}}{\sqrt{B}}+\frac{C_{2}^{2}\sqrt{B}\varepsilon}{\log n}\big)-1} \notag\\
 & >e^{-1}\sum\nolimits_{i=1}^{n}r_{i}p_{i,j}^{t}-0.5/n>2c_{j}+1.5/n,
	\label{eq:sum-r-pbar-lower-bound-179}
\end{align}
}
where the second inequality invokes \eqref{eq:sum-r-p-relation-179}, 
and the last line holds as long as $C_1$ is sufficiently large (since $B=C_1\log\frac{n}{\varepsilon}$) and makes use of the assumption that $j\notin \mathcal{J}_t$. 
To explain the penultimate line in \eqref{eq:sum-r-pbar-lower-bound-179}, it suffices to 
combine our parameter choices 
with the two properties below: 
\begin{itemize}
	\item 

Recognizing that $0<\tau<\frac{1}{6\max_{i}\eta_{p,i}r_{i}+2\eta B}$, we
have
\[
\big[\exp\big(3\max_{i}\eta_{p,i}r_{i}+\eta B\big)\big]^{-\tau}\geq\exp\left(-1/2\right).
\]

	\item 

Given that $3\max_{i}\eta_{p,i}r_{i}+\eta B=\frac{3C_{2}}{\sqrt{C_{1}\log(n/\varepsilon)}}+\frac{C_{2}^{2}\varepsilon\sqrt{C_{1}\log(n/\varepsilon)}}{\log n}\ll1$
and that $B=C_{1}\log\frac{n}{\varepsilon}\gg1$, one has 
\[
\frac{1}{(1+e^{-B})\exp\big(3\max_{i}\eta_{p,i}r_{i}+\eta B\big)}\geq\exp\left(-\frac{1}{2}\right).
\]

\end{itemize}

The above  bounds \eqref{eq:sum-r-p-lower-bound-179} and \eqref{eq:sum-r-pbar-lower-bound-179} 
play a useful role for bounding the changes of $\bm{\mu}_{j}^{t,\trunc}$ and $\overline{\bm{\mu}}_{j}^{t}$. 
Consider any $t\geq \frac{1}{3\max_i\eta_{p,i} r_i+\eta B}$ and an integer $\tau_0 = \frac{1}{12\max_i\eta_{p,i} r_i+4\eta B}$. 
\begin{itemize}
	\item Suppose for the moment that $\mu_{j, +}^{t+1-\tau_0} \geq  \mu_{j, -}^{t+1-\tau_0}$. 
Then in view of the elementary fact \eqref{eq:trunc-dimension-2}, we have
{\small
\begin{align*}
	\frac{\mu_{j,+}^{t+1-\tau_0,\trunc}}{\mu_{j,-}^{t+1-\tau_0,\trunc}}
	=\min\bigg\{\frac{\mu_{j,+}^{t+1-\tau_0}}{\mu_{j,-}^{t+1-\tau_0}},e^{B}\bigg\}.
\end{align*}
		}
Taking this together with the update rule~\eqref{eq:update-main} reveals that: for every $j \notin \mathcal{J}_t$,  
		{\small
\begin{align}
\frac{\mu_{j,+}^{t+1-\tau_{0},\trunc}}{\mu_{j,-}^{t+1-\tau_{0},\trunc}} & =\min\Bigg\{\bigg(\frac{\mu_{j,+}^{t-\tau_{0},\trunc}}{\mu_{j,-}^{t-\tau_{0},\trunc}}\bigg)^{1-\eta}\exp\bigg(2\eta_{\mu,j}\Big(\sum_{i=1}^{n}r_{i}\overline{p}_{i,j}^{t+1-\tau_{0}}-c_{j}\Big)\bigg),\ e^{B}\Bigg\}\nonumber \\
 & \geq\min\Bigg\{\frac{\mu_{j,+}^{t-\tau_{0},\trunc}}{\mu_{j,-}^{t-\tau_{0},\trunc}}\cdot\bigg(\frac{\mu_{j,+}^{t-\tau_{0},\trunc}}{\mu_{j,-}^{t-\tau_{0},\trunc}}\bigg)^{-\eta}\exp\bigg(2\eta_{\mu,j}\Big(c_{j}+\frac{1.5}{n}\Big)\bigg),\ e^{B}\Bigg\}\nonumber \\
 & \geq\min\Bigg\{\frac{\mu_{j,+}^{t-\tau_{0},\trunc}}{\mu_{j,-}^{t-\tau_{0},\trunc}}\exp\bigg(2\eta_{\mu,j}\Big(c_{j}+\frac{C_3}{n}\Big)-\eta B\bigg),\ e^{B}\Bigg\}\label{eq:sum-r-pbar-lower-bound-429} \\
 & =\min\Bigg\{\frac{\mu_{j,+}^{t-\tau_{0},\trunc}}{\mu_{j,-}^{t-\tau_{0},\trunc}}\exp\bigg(30C_{2}\sqrt{B}-\frac{C_{2}^{2}\varepsilon}{\log n}\sqrt{B}\bigg),\ e^{B}\Bigg\} \notag \\
 & \geq\min\Bigg\{\frac{\mu_{j,+}^{t-\tau_{0},\trunc}}{\mu_{j,-}^{t-\tau_{0},\trunc}},\ e^{B}\Bigg\}=\frac{\mu_{j,+}^{t-\tau_{0},\trunc}}{\mu_{j,-}^{t-\tau_{0},\trunc}},\label{eq:monotone-mu-ratio}
\end{align}
		}
where the third line relies on \eqref{eq:sum-r-pbar-lower-bound-179}, the assumption $C_3\leq 1$, and the elementary fact $\frac{\mu_{j,+}^{t-\tau_0,\trunc}}{\mu_{j,-}^{t-\tau_0,\trunc}}\leq e^B$ (see \eqref{eq:trunc-dimension-2}), 
the fourth line results from the choice \eqref{eq:parameter}, 
and the last identity holds since $\frac{\mu_{j,+}^{t-\tau_0,\trunc}}{\mu_{j,-}^{t-\tau_0,\trunc}}\leq e^B$ (see \eqref{eq:trunc-dimension-2}). 
Clearly, repeating the above argument recursively yields 
{\small
\begin{align}
	\frac{\mu_{j, +}^{t, \trunc}}{\mu_{j, -}^{t, \trunc}} \geq \frac{\mu_{j, +}^{t-1, \trunc}}{\mu_{j, -}^{t-1, \trunc}}
	\geq \cdots \geq \frac{\mu_{j, +}^{t-\tau_0, \trunc}}{\mu_{j, -}^{t-\tau_0, \trunc}}. 
\end{align}
}
Furthermore, if $\frac{\mu_{j, +}^{t, \trunc}}{\mu_{j, -}^{t, \trunc}} < \exp(B)$, then we have also seen from \eqref{eq:sum-r-pbar-lower-bound-429} that
{\small
\[
\frac{\mu_{j,+}^{t+1-\tau_{0},\trunc}}{\mu_{j,-}^{t+1-\tau_{0},\trunc}}\geq\frac{\mu_{j,+}^{t-\tau_{0},\trunc}}{\mu_{j,-}^{t-\tau_{0},\trunc}}\exp\bigg(2\eta_{\mu,j}\Big(c_{j}+\frac{C_3}{n}\Big)-\eta B\bigg), 
\]
	}
and similarly, 
{\small
\begin{align*}
\frac{\mu_{j,+}^{t,\trunc}}{\mu_{j,-}^{t,\trunc}} & \geq\frac{\mu_{j,+}^{t-\tau_{0},\trunc}}{\mu_{j,-}^{t-\tau_{0},\trunc}}\exp\bigg\{\tau_{0}\Big(2\eta_{\mu,j}\Big(c_{j}+\frac{C_3}{n}\Big)-\eta B\Big)\bigg\}
  \geq\exp\bigg\{\tau_{0}\Big(2\eta_{\mu,j}\Big(c_{j}+\frac{C_3}{n}\Big)-\eta B\Big)\bigg\}\geq e^{B}
\end{align*}
		}
		as long as $\frac{2\eta_{\mu,j} (c_j + C_3/n) - \eta B}{12\max_i\eta_{p,i} r_i+4\eta B} > 2B$ (a condition that is satisfied under our choice of parameters\footnote{To be more specific, we have
$
\frac{2\eta_{\mu,j}(c_{j}+C_{3}/n)-\eta B}{12\max_{i}\eta_{p,i}r_{i}+4\eta B}=\frac{30C_{2}\sqrt{B}-\frac{C_{2}^{2}\varepsilon\sqrt{B}}{\log n}}{12C_{2}/\sqrt{B}+\frac{4C_{2}^{2}\varepsilon\sqrt{B}}{\log n}}\geq\frac{29C_{2}\sqrt{B}}{13C_{2}/\sqrt{B}}>2B.
$
		}). 
This, however, leads to contradiction. Thus,  we necessarily have
{\small
\begin{align}
	\mu_{j, +}^{t, \trunc} / \mu_{j, -}^{t, \trunc} = e^B,
	\qquad 
	\text{if } \mu_{j, +}^{t+1-\tau_0} /  \mu_{j, -}^{t+1-\tau_0} \geq 1. 
	\label{eq:stopping-point-1st-case}
\end{align}
}

\item On the other hand, consider the case where $\mu_{j, +}^{t+1-\tau_0} <  \mu_{j, -}^{t+1-\tau_0}$. 
Then  \eqref{eq:trunc-dimension-2} gives
{\small
\begin{align*}
	\mu_{j,+}^{t+1-\tau_0,\trunc} / \mu_{j,-}^{t+1-\tau_0,\trunc}
	=\max\Big\{ \mu_{j,+}^{t+1-\tau_0}/\mu_{j,-}^{t+1-\tau_0},\,e^{-B}\Big\} .
\end{align*}
}
Repeating the analysis for \eqref{eq:monotone-mu-ratio} tells us that: for every $j \notin \mathcal{J}_t$, 
{\small
\begin{align}
\frac{\mu_{j,+}^{t+1-\tau_{0},\trunc}}{\mu_{j,-}^{t+1-\tau_{0},\trunc}} & \geq\max\Bigg\{\frac{\mu_{j,+}^{t-\tau_{0},\trunc}}{\mu_{j,-}^{t-\tau_{0},\trunc}}\exp\bigg(2\eta_{\mu,j}\Big(c_{j}+\frac{C_3}{n}\Big)-\eta B\bigg),\ e^{-B}\Bigg\}\nonumber \\
 & \geq\max\Bigg\{\frac{\mu_{j,+}^{t-\tau_{0},\trunc}}{\mu_{j,-}^{t-\tau_{0},\trunc}},\ e^{-B}\Bigg\}=\frac{\mu_{j,+}^{t-\tau_{0},\trunc}}{\mu_{j,-}^{t-\tau_{0},\trunc}}. \label{eq:monotone-mu-ratio-oppo}
\end{align}
}
With such monotonicity in place, repeat the argument for \eqref{eq:stopping-point-1st-case} to show that (which we omit for brevity)
{\small
\begin{align}
	\mu_{j, +}^{t, \trunc} / \mu_{j, -}^{t, \trunc} = e^B,
	\qquad 
	\text{if } \mu_{j, +}^{t+1-\tau_0} / \mu_{j, -}^{t+1-\tau_0} < 1, 
	\label{eq:stopping-point-2nd-case}
\end{align}
}
provided that $\frac{2\eta_{\mu,j} (c_j + C_3/n) - \eta B}{12\max_i\eta_{p,i} r_i+4\eta B} > 2B$ (satisfied under our parameter choice). 

\end{itemize}

Recall the update rules \eqref{eq:update-midpoints-123} and \eqref{eq:update-main-135} for $\overline{\mu}_{j,+}^{t+1}$ and $\mu_{j,+}^{t+1}$, respectively. 
Combining \eqref{eq:stopping-point-1st-case} and \eqref{eq:stopping-point-2nd-case} and reusing the argument in \eqref{eq:monotone-mu-ratio} further show that
{\small
\begin{align*}
\frac{\overline{\mu}_{j,+}^{t+1}}{\overline{\mu}_{j,-}^{t+1}} & =\bigg(\frac{\mu_{j,+}^{t,\trunc}}{\mu_{j,-}^{t,\trunc}}\bigg)^{1-\eta}\exp\bigg(2\eta_{\mu,j}\Big(\sum_{i=1}^{n}r_{i}p_{i,j}^{t}-c_{j}\Big)\bigg)\\
 & \geq\frac{\mu_{j,+}^{t,\trunc}}{\mu_{j,-}^{t,\trunc}}\exp\bigg(2\eta_{\mu,j}\Big(c_{j}+\frac{C_{3}}{n}\Big)-\eta B\bigg)\geq\frac{\mu_{j,+}^{t,\trunc}}{\mu_{j,-}^{t,\trunc}}= e^{B}
\end{align*}
}
{\small
\begin{align*}
\text{and}\quad\frac{\mu_{j,+}^{t+1}}{\mu_{j,-}^{t+1}} & =\bigg(\frac{\mu_{j,+}^{t,\trunc}}{\mu_{j,-}^{t,\trunc}}\bigg)^{1-\eta}\exp\bigg(2\eta_{\mu,j}\Big(\sum_{i=1}^{n}r_{i}\overline{p}_{i,j}^{t+1}-c_{j}\Big)\bigg)\\
 & \geq\frac{\mu_{j,+}^{t,\trunc}}{\mu_{j,-}^{t,\trunc}}\exp\bigg(2\eta_{\mu,j}\Big(c_{j}+\frac{C_{3}}{n}\Big)-\eta B\bigg)\geq\frac{\mu_{j,+}^{t,\trunc}}{\mu_{j,-}^{t,\trunc}}= e^{B},
\end{align*}
}
provided that $\frac{2\eta_{\mu,j} (c_j + C_3/n) - \eta B}{12\max_i\eta_{p,i} r_i+4\eta B} > 2B$. 
In turn, these two bounds tell us that 
{\small
\[
\overline{\mu}_{j,-}^{t+1}  =\frac{\overline{\mu}_{j,-}^{t+1}}{\overline{\mu}_{j,+}^{t+1}+\overline{\mu}_{j,-}^{t+1}}\leq\frac{1}{1+e^{B}}\leq e^{-B}\quad\text{and}\quad{\mu}_{j,-}^{t+1}=\frac{\mu_{j,-}^{t+1}}{\mu_{j,+}^{t+1}+\mu_{j,-}^{t+1}}\leq\frac{1}{1+e^{B}}\leq e^{-B}.	
\]
}
Armed with these results as well as \eqref{eq:complement-Jt-emptyset}, we can conclude that for all $t\geq 0$, 
{\small
\begin{align}
 & \sum_{j\notin\mathcal{J}_{t}}\big(\overline{\mu}_{j,+}^{t+1}-\overline{\mu}_{j,-}^{t+1}-\mu_{j,+}^{t+1}+\mu_{j,-}^{t+1}\big)\sum_{i=1}^{n}r_{i}\big(p_{i,j}^{t}-\overline{p}_{i,j}^{t+1}\big)\nonumber\\
 & \qquad\qquad+\sum_{j\notin\mathcal{J}_{t}}\big(\overline{\mu}_{j,+}^{t+1}-\overline{\mu}_{j,-}^{t+1}-\mu_{j,+}^{t,\trunc}+\mu_{j,-}^{t,\trunc}\big)\sum_{i=1}^{n}r_{i}\big(\overline{p}_{i,j}^{t+1}-p_{i,j}^{t+1}\big)\nonumber\\
 &\le\sum_{j\notin\mathcal{J}_{t}}2\big|\overline{\mu}_{j,-}^{t+1}-\mu_{j,-}^{t+1}\big|\bigg|\sum_{i=1}^{n}r_{i}\big(p_{i,j}^{t}-\overline{p}_{i,j}^{t+1}\big)\bigg|+\sum_{j\notin\mathcal{J}_{t}}2\big|\overline{\mu}_{j,-}^{t+1}-\mu_{j,-}^{t,\trunc}\big|\bigg|\sum_{i=1}^{n}r_{i}\big(\overline{p}_{i,j}^{t+1}-p_{i,j}^{t+1}\big)\bigg| \notag\\
 &\leq4\sum_{j\notin\mathcal{J}_{t}}\left(\big|\overline{\mu}_{j,-}^{t+1}-\mu_{j,-}^{t+1}\big|+\big|\overline{\mu}_{j,-}^{t+1}-\mu_{j,-}^{t,\trunc}\big|\right)\le8ne^{-B}.
	\label{eq:step3-final}
\end{align}
}
Here, the first inequality in \eqref{eq:step3-final} follows since  
\begin{align*}
\overline{\mu}_{j,+}^{t+1}-\overline{\mu}_{j,-}^{t+1}-\mu_{j,+}^{t+1}+\mu_{j,-}^{t+1} & =\overline{\mu}_{j,+}^{t+1}-\big(1-\overline{\mu}_{j,+}^{t+1}\big)-\mu_{j,+}^{t+1}+\big(1-\mu_{j,+}^{t+1}\big)=2\big(\overline{\mu}_{j,+}^{t+1}-\mu_{j,+}^{t+1}\big),\\
\overline{\mu}_{j,+}^{t+1}-\overline{\mu}_{j,-}^{t+1}-\mu_{j,+}^{t,\trunc}+\mu_{j,-}^{t,\trunc} & =2\big(\overline{\mu}_{j,+}^{t+1}-\mu_{j,+}^{t,\trunc}\big); 
\end{align*}
the second inequality in \eqref{eq:step3-final} is valid since $\sum_{i=1}^{n}r_{i}=1$; 
and the last inequality in \eqref{eq:step3-final} follows from the above upper bounds on $\overline{\mu}_{j,-}^{t+1}$, $\mu_{j,-}^{t+1}$ and $\mu_{j,-}^{t,\trunc}$.

\subsubsection{Step 4: putting all this together}

Putting \eqref{eq:step2-final} and \eqref{eq:step3-final} together with \eqref{eq:prediction-error}, we arrive at
{\small
\[
\big\langle\overline{\bm{\zeta}}^{t+1}-\bm{\zeta}^{t+1},\,\log\overline{\bm{\zeta}}^{t+1}-\log\bm{\zeta}^{t+1}\big\rangle\leq(1-\eta)\dist\big(\overline{\bm{\zeta}}^{t+1}\parallel\bm{\zeta}^{t,\trunc}\big)+\dist\big(\bm{\zeta}^{t+1}\parallel\overline{\bm{\zeta}}^{t+1}\big)+8ne^{-B}.
\]
}
Substituting it back into \eqref{eq:recursion}, we reach
{\small
\begin{align*}
\dist\big(\bm{\zeta}^{\star}\parallel\bm{\zeta}^{t+1}\big) & \leq(1-\eta)\dist\big(\bm{\zeta}^{\star}\parallel\bm{\zeta}^{t,\trunc}\big)+8ne^{-B}, 
\end{align*}
}
thereby concluding the proof of Claim~\eqref{eq:recursion-bound}.

\section{Discussion}
\label{sec:discussion}

In this paper, we have put forward a first-order method for computing the optimal transport at scale,  
which has been shown to enjoy both intriguing convergence guarantees and favorable numerical performance. 
This is a step we have taken towards closing the theory-practice gap for solving this problem. 
Moving forward, there are several natural research directions to explore. 
To begin with, while the state-of-the-art theory \cite{van2020bipartite} demonstrated the feasibility of a runtime $O(n^2\log^2(1/\varepsilon))$,  
the practical value of the algorithm proposed therein remains unrealized; it would be of great importance to design algorithms that are optimal in theory and practice at once. 
Next, the current algorithm still involves several hyper-parameters to tune, 
and it would be of interest to develop improved versions that are nearly parameter-free. 
Also, there is no shortage of applications where the problems exhibit certain low-dimensional structure 
(e.g., \cite{altschuler2019massively}), 
which could be potentially leveraged to achieve further computational savings. Another natural problem to explore is whether we can solve a more general family of linear programs --- e.g., the ones taking the form 
 $\text{minimize}_{\bm{x}\in \Delta_n} f(\bm{x})+\|\bm{A}\bm{x}-\bm{b}\|_1$ --- using the entropy-regularized extragradient method developed herein. 
We leave these for future investigation.



\section*{Acknowledgements}

Y.~Chen is supported in part by the Alfred P.~Sloan Research Fellowship, the Google Research Scholar Award, the AFOSR grant FA9550-22-1-0198, 
the ONR grant N00014-22-1-2354,  and the NSF grants CCF-2221009, CCF-1907661,  IIS-2218713 and IIS-2218773. 
 Y.~Chi is supported in part by the ONR grant N00014-19-1-2404.

\appendix

\section{Converting a matrix to a transportation plan}
\label{sec:convert-algorithm-P}

Given a general non-negative matrix $\widehat{\bm{P}}\in \mathbb{R}_+^{n\times n}$, 
\cite{altschuler2017near} put forward a simple algorithm that returns a probability matrix $\widetilde{\bm{P}}\in \Delta_{n\times n}$ 
satisfying $\widetilde{\bm{P}}\bm{1}=\bm{r}$ and  $\widetilde{\bm{P}}^{\top}\bm{1}=\bm{c}$ while simultaneously obeying the condition in Lemma~\ref{lem:projection}.  
We include this algorithm here in order to be self-contained; 
here, we recall that $\mathsf{row}_i(\bm{F})$ (resp.~$\mathsf{col}_i(\bm{F})$) denotes the sum of the $i$-th row (resp.~column) of a matrix $\bm{F}$. 

\begin{algorithm}[h]
	\DontPrintSemicolon
	\SetNoFillComment
	\vspace{-0.3ex}
	\textbf{Input}: $\widehat{\bm{P}} \in \mathbb{R}_+^{n\times n}$, and two probability vectors $\bm{r}=[r_i]_{1\leq i\leq n}, \bm{c}=[c_i]_{1\leq i\leq n} \in \Delta_n$. \\
	\For{$ i = 1$ \KwTo $n$}{
		$x_i = \min \big\{ \frac{r_i}{\mathsf{row}_i(\widehat{\bm{P}})}, \, 1\big\}$.  
	}
	
	$\bm{F}^{\prime}= \mathsf{diag}([x_i]_{1\leq i\leq n})\widehat{\bm{P}} $. \\
	\For{$ i = 1$ \KwTo $n$}{
		$y_i = \min \big\{ \frac{c_i}{\mathsf{col}_i(\bm{F}^{\prime})}, \, 1\big\}$. 
	}	
	$\bm{F}^{\prime\prime}= \bm{F}^{\prime}\mathsf{diag}([y_i]_{1\leq i\leq n}) .$ \\
	$\bm{e}_{\mathrm{r}} = \bm{r} - \mathsf{row}(\bm{F}^{\prime\prime})$; $\bm{e}_{\mathrm{c}} = \bm{c} - \mathsf{col}(\bm{F}^{\prime\prime})$. \\
	\textbf{Output:}  $\widetilde{\bm{P}}= \bm{F}^{\prime\prime} + \bm{e}_{\mathrm{r}}\bm{e}_{\mathrm{c}}^{\top} / \| \bm{e}_{\mathrm{r}} \|_1$.
	\caption{Converting a matrix $\widehat{\bm{P}}$ to a feasible transportation plan \cite{altschuler2017near}.\label{alg:convert-P}}
\end{algorithm}

\section{Proof of Equations~\eqref{eq:identity} and \eqref{eq:prediction-error}}
\label{sec:proof-identity}

\paragraph{Proof of the identity~\eqref{eq:identity}}
The optimizer of the regularized minimax problem \eqref{eq:game-entropy} necessarily satisfies the following optimality condition: for each $1\leq j\leq n$, 
{\small
\begin{align}
	\log \mu_{j, s}^{\star,\mathsf{reg}} = \alpha_{\mu,j} + \frac{\eta_{\mu,j}}{\eta}s\bigg(\sum_{i=1}^{n}r_{i}p_{i,j}^{\star,\mathsf{reg}}-c_{j}\bigg)
	\qquad s\in \{ +, -\}  
	\label{eq:optimality-condition-mu-js}
\end{align}
}
for some normalization factor $\alpha_{\mu,j}$;  
this can be easily seen by setting the gradient of the objective function to zero and utilizing the constraint $\bm{\mu}_j^{\star,\mathsf{reg}}\in \Delta_2$ as well as $\tau_{\mu,j}=\frac{\eta}{\eta_{\mu,j}}$. 
Additionally, the update rule~\eqref{eq:update-main} implies that
{\small
\begin{equation}
	\log\mu_{j,s}^{t+1}=\beta_{\mu,j}+(1-\eta)\log\mu_{j,s}^{t,\trunc}+\eta\cdot\frac{\eta_{\mu,j}}{\eta} s \bigg(\sum_{i=1}^{n}r_{i}\overline{p}_{i,j}^{t+1}-c_{j}\bigg),
	\qquad s\in \{ +, - \}, \label{eq:log-condition-mu-t-mu-trunc-UR}
\end{equation}
}
where $\beta_{\mu,j}$ is some normalization constant. 
Taking the preceding two identities together and using the basic facts $\big\langle \overline{\bm{\mu}}_j^{t+1} - \bm{\mu}_j^{\star,\mathsf{reg}}, \bm{1}\big\rangle = 0$ give
{\small
\begin{align*}
&\big\langle \overline{\bm{\mu}}_j^{t+1} - \bm{\mu}_j^{\star,\mathsf{reg}}, \, \log \bm{\mu}_j^{t+1} - (1-\eta)\log \bm{\mu}_j^{t, \trunc} - \eta\log \bm{\mu}_j^{\star,\mathsf{reg}}\big\rangle \\
&\qquad = \eta_{\mu,j}\big(\overline{\mu}_{j,+}^{t+1}-\overline{\mu}_{j,-}^{t+1} - \mu_{j,+}^{\star,\mathsf{reg}} + \mu_{j,-}^{\star,\mathsf{reg}}\big)\cdot \sum\nolimits_{i=1}^{n}r_{i}\big(\overline{p}_{i,j}^{t+1}-p_{i,j}^{\star,\mathsf{reg}}\big) 
\end{align*}
}
for any $1 \leq j\leq n$. 
A similar argument also leads to 
{\small
\begin{align*}
&\big\langle \overline{\bm{p}}_i^{t+1} - \bm{p}_i^{\star,\mathsf{reg}}, \, \log \bm{p}_i^{t+1} - (1-\eta)\log \bm{p}_i^{t} - \eta\log \bm{p}_i^{\star,\mathsf{reg}}\big\rangle \\
&\qquad= \eta_{p,i}\sum\nolimits_{j=1}^n \big(\overline{p}_{i,j}^{t+1} - p_{i,j}^{\star,\mathsf{reg}}\big) \cdot r_{i}\big( \mu_{j,+}^{\star,\mathsf{reg}}-\mu_{j,-}^{\star,\mathsf{reg}} - \overline{\mu}_{j,+}^{t+1}+\overline{\mu}_{j,-}^{t+1}\big) 
\end{align*}
}
for any $1 \leq i\leq n$. Putting the above two identities together gives 
{\small
\begin{align}
	&\big\langle \overline{\bm{\zeta}}^{t+1} - \bm{\zeta}^{\star}, \, \log \bm{\zeta}^{t+1} - (1-\eta)\log \bm{\zeta}^{t, \trunc} - \eta\log \bm{\zeta}^{\star}\big\rangle \notag\\
	&\qquad= \sum\nolimits_{j = 1}^n\frac{1}{\eta_{\mu,j}}\big\langle \overline{\bm{\mu}}_j^{t+1} - \bm{\mu}_j^{\star,\mathsf{reg}}, \, \log \bm{\mu}_j^{t+1} - (1-\eta)\log \bm{\mu}_j^{t, \trunc} - \eta\log \bm{\mu}_j^{\star,\mathsf{reg}}\big\rangle \notag\\
	&\qquad\qquad+ \sum\nolimits_{i=1}^n\frac{1}{\eta_{p,i}}\big\langle \overline{\bm{p}}_i^{t+1} - \bm{p}_i^{\star,\mathsf{reg}}, \, \log \bm{p}_i^{t+1} - (1-\eta)\log \bm{p}_i^{t} - \eta\log \bm{p}_i^{\star,\mathsf{reg}}\big\rangle \notag\\
	&\qquad = \sum\nolimits_{j = 1}^n \big(\overline{\mu}_{j,+}^{t+1}-\overline{\mu}_{j,-}^{t+1} - \mu_{j,+}^{\star,\mathsf{reg}} + \mu_{j,-}^{\star,\mathsf{reg}}\big)\cdot \sum\nolimits_{i=1}^{n}r_{i}\big(\overline{p}_{i,j}^{t+1}-p_{i,j}^{\star,\mathsf{reg}}\big) 
	\notag\\
	&\qquad\qquad 
	+ \sum\nolimits_{i=1}^n \sum\nolimits_{j=1}^n \big(\overline{p}_{i,j}^{t+1} - p_{i,j}^{\star,\mathsf{reg}}\big) \cdot r_{i}\big( \mu_{j,+}^{\star,\mathsf{reg}}-\mu_{j,-}^{\star,\mathsf{reg}} - \overline{\mu}_{j,+}^{t+1}+\overline{\mu}_{j,-}^{t+1}\big) 
	\  = 0.
\end{align}
}
%

\paragraph{Proof of the identity~\eqref{eq:prediction-error}}

Repeating similar arguments as in the proof of \eqref{eq:identity} and using the update rules~\eqref{eq:update-midpoints} and~\eqref{eq:update-main}, 
we can also deduce that: for any $j$, 
{\small
\begin{align*}
\big\langle\log\overline{\bm{\mu}}_{j}^{t+1},\overline{\bm{\mu}}_{j}^{t+1}-\bm{\mu}_{j}^{t+1}\big\rangle & =\Big\langle(1-\eta)\log\bm{\mu}_{j}^{t,\trunc}+\eta_{\mu,j}\Big(\sum_{i=1}^{n}r_{i}p_{i,j}^{t}-c_{j}\Big)\left[\begin{array}{c}
1\\
-1
\end{array}\right],\,\overline{\bm{\mu}}_{j}^{t+1}-\bm{\mu}_{j}^{t+1}\Big\rangle,\\
\big\langle\log\bm{\mu}_{j}^{t+1},\overline{\bm{\mu}}_{j}^{t+1}-\bm{\mu}_{j}^{t+1}\big\rangle & =\Big\langle(1-\eta)\log\bm{\mu}_{j}^{t,\trunc}+\eta_{\mu,j}\Big(\sum_{i=1}^{n}r_{i}\overline{p}_{i,j}^{t+1}-c_{j}\Big)\left[\begin{array}{c}
1\\
-1
\end{array}\right],\,\overline{\bm{\mu}}_{j}^{t+1}-\bm{\mu}_{j}^{t+1}\Big\rangle,
\end{align*}
}
and as a result, 
{\small
\begin{align*}
\frac{1}{\eta_{\mu,j}}\big\langle\log\overline{\bm{\mu}}_{j}^{t+1}-\log\bm{\mu}_{j}^{t+1},\overline{\bm{\mu}}_{j}^{t+1}-\bm{\mu}_{j}^{t+1}\big\rangle 
	& =\bigg\langle\sum_{i=1}^{n}r_{i}\big(p_{i,j}^{t}-\overline{p}_{i,j}^{t+1}\big)\left[\begin{array}{c}
1\\
-1
\end{array}\right],\,\overline{\bm{\mu}}_{j}^{t+1}-\bm{\mu}_{j}^{t+1}\bigg\rangle\\
 & =\big(\overline{\mu}_{j,+}^{t+1}-\overline{\mu}_{j,-}^{t+1}-\mu_{j,+}^{t+1}+\mu_{j,-}^{t+1}\big)\sum_{i=1}^{n}r_{i}\big(p_{i,j}^{t}-\overline{p}_{i,j}^{t+1}\big).
\end{align*}
}
Similarly, the update rules~\eqref{eq:update-midpoints} and~\eqref{eq:update-main} also indicate that
{\small
\begin{align*}
\big\langle\log\overline{\bm{p}}_{i}^{t+1},\,\overline{\bm{p}}_{i}^{t+1}-\bm{p}_{i}^{t+1}\big\rangle 
	& =\Big\langle(1-\eta)\log\bm{p}_{i}^{t} - \eta_{p,i}r_{i}\big(0.5\bm{w}_{i}+\bm{\mu}_{+}^{t,\trunc}-\bm{\mu}_{-}^{t,\trunc}\big),\,\overline{\bm{p}}_{i}^{t+1}-\bm{p}_{i}^{t+1}\Big\rangle\\
\big\langle\log\bm{p}_{i}^{t+1},\,\overline{\bm{p}}_{i}^{t+1}-\bm{p}_{i}^{t+1}\big\rangle 
	& =\Big\langle(1-\eta)\log\bm{p}_{i}^{t} - \eta_{p,i}r_{i}\big(0.5\bm{w}_{i}+\overline{\bm{\mu}}_{+}^{t+1}-\overline{\bm{\mu}}_{-}^{t+1}\big),\,\overline{\bm{p}}_{i}^{t+1}-\bm{p}_{i}^{t+1}\Big\rangle
\end{align*}
}
with $\bm{\mu}_{s}^{t,\trunc}\coloneqq[\mu_{j,s}^{t,\trunc}]_{1\leq j\leq n}$
and $\bm{\mu}_{s}^{t}\coloneqq[\mu_{j,s}^{t}]_{1\leq j\leq n}$ for
all $s\in\{+,-\}$, which yield
{\small
\begin{align*}
\frac{1}{\eta_{p,i}}\big\langle\log\overline{\bm{p}}_{i}^{t+1}-\log\bm{p}_{i}^{t+1},\overline{\bm{p}}_{i}^{t+1}-\bm{p}_{i}^{t+1}\big\rangle 
	& = -r_i \Big\langle\bm{\mu}_{+}^{t,\trunc}-\bm{\mu}_{-}^{t,\trunc}-\overline{\bm{\mu}}_{+}^{t+1}+\overline{\bm{\mu}}_{-}^{t+1},\,\overline{\bm{p}}_{i}^{t+1}-\bm{p}_{i}^{t+1}\Big\rangle\\
 & = - \sum_{j=1}^{n}r_{i}\big(\mu_{j,+}^{t,\trunc}-\mu_{j,-}^{t,\trunc}-\overline{\mu}_{j,+}^{t+1}+\overline{\mu}_{j,-}^{t+1}\big)\big(\overline{p}_{i,j}^{t+1}-p_{i,j}^{t+1}\big).
\end{align*}
}
Taking the above results together, we arrive at the advertised relation: 
{\small
\begin{align*}
\big\langle\log\overline{\bm{\zeta}}^{t+1}-\log\bm{\zeta}^{t+1},\overline{\bm{\zeta}}^{t+1}-\bm{\zeta}^{t+1}\big\rangle & =\sum\nolimits_{j=1}^{n}\frac{1}{\eta_{\mu,j}}\big\langle\log\overline{\bm{\mu}}_{j}^{t+1}-\log\bm{\mu}_{j}^{t+1},\,\overline{\bm{\mu}}_{j}^{t+1}-\bm{\mu}_{j}^{t+1}\big\rangle\\
 & \qquad+\sum\nolimits_{i=1}^{n}\frac{1}{\eta_{p,i}}\big\langle\log\overline{\bm{p}}_{i}^{t+1}-\log\bm{p}_{i}^{t+1},\,\overline{\bm{p}}_{i}^{t+1}-\bm{p}_{i}^{t+1}\big\rangle\\
 & =\sum\nolimits_{j=1}^{n}\sum\nolimits_{i=1}^{n}\big(\overline{\mu}_{j,+}^{t+1}-\overline{\mu}_{j,-}^{t+1}-\mu_{j,+}^{t+1}+\mu_{j,-}^{t+1}\big)r_{i}\big(p_{i,j}^{t}-\overline{p}_{i,j}^{t+1}\big)\nonumber\\
 & \qquad+\sum\nolimits_{j=1}^{n}\sum\nolimits_{i=1}^{n}\big(\overline{\mu}_{j,+}^{t+1}-\overline{\mu}_{j,-}^{t+1}-\mu_{j,+}^{t,\trunc}+\mu_{j,-}^{t,\trunc}\big)r_{i}\big(\overline{p}_{i,j}^{t+1}-p_{i,j}^{t+1}\big).
\end{align*}
}

\bibliographystyle{apalike}
\bibliography{reference-OT}

\begin{thebibliography}{}

\bibitem[Allen-Zhu et~al., 2017]{allen2017much}
Allen-Zhu, Z., Li, Y., Oliveira, R., and Wigderson, A. (2017).
\newblock Much faster algorithms for matrix scaling.
\newblock In {\em IEEE Annual Symposium on Foundations of Computer Science
  (FOCS)}, pages 890--901.

\bibitem[Allen-Zhu and Orecchia, 2015]{allen2015nearly}
Allen-Zhu, Z. and Orecchia, L. (2015).
\newblock Nearly-linear time positive {LP} solver with faster convergence rate.
\newblock In {\em Annual ACM symposium on Theory of Computing}, pages 229--236.

\bibitem[Altschuler et~al., 2019]{altschuler2019massively}
Altschuler, J., Bach, F., Rudi, A., and Niles-Weed, J. (2019).
\newblock Massively scalable {S}inkhorn distances via the {N}ystr{\"o}m method.
\newblock {\em NeurIPS}, 32.

\bibitem[Altschuler et~al., 2017]{altschuler2017near}
Altschuler, J., Niles-Weed, J., and Rigollet, P. (2017).
\newblock Near-linear time approximation algorithms for optimal transport via
  sinkhorn iteration.
\newblock {\em Advances in neural information processing systems}, 30.

\bibitem[Altschuler, 2022]{altschuler2022flows}
Altschuler, J.~M. (2022).
\newblock Flows, scaling, and entropy revisited: a unified perspective via
  optimizing joint distributions.
\newblock {\em arXiv preprint arXiv:2210.16456}.

\bibitem[An et~al., 2022]{an2022efficient}
An, D., Lei, N., Xu, X., and Gu, X. (2022).
\newblock Efficient optimal transport algorithm by accelerated gradient
  descent.
\newblock {\em AAAI Conference on Artificial Intelligence}, 36(9):10119--10128.

\bibitem[Ao et~al., 2023]{ao2022asynchronous}
Ao, R., Cen, S., and Chi, Y. (2023).
\newblock Asynchronous gradient play in zero-sum multi-agent games.
\newblock In {\em International Conference on Learning Representations (ICLR)}.

\bibitem[Arjovsky et~al., 2017]{arjovsky2017wasserstein}
Arjovsky, M., Chintala, S., and Bottou, L. (2017).
\newblock Wasserstein generative adversarial networks.
\newblock In {\em International conference on machine learning}, pages
  214--223. PMLR.

\bibitem[Blanchet et~al., 2018]{blanchet2018towards}
Blanchet, J., Jambulapati, A., Kent, C., and Sidford, A. (2018).
\newblock Towards optimal running times for optimal transport.
\newblock {\em arXiv preprint arXiv:1810.07717}.

\bibitem[Carlier, 2022]{carlier2022linear}
Carlier, G. (2022).
\newblock On the linear convergence of the multimarginal {S}inkhorn algorithm.
\newblock {\em SIAM Journal on Optimization}, 32(2):786--794.

\bibitem[Cen et~al., 2022a]{cen2022independent}
Cen, S., Chen, F., and Chi, Y. (2022a).
\newblock Independent natural policy gradient methods for potential games:
  Finite-time global convergence with entropy regularization.
\newblock In {\em IEEE Conference on Decision and Control (CDC)}.

\bibitem[Cen et~al., 2022b]{cen2022fast}
Cen, S., Cheng, C., Chen, Y., Wei, Y., and Chi, Y. (2022b).
\newblock Fast global convergence of natural policy gradient methods with
  entropy regularization.
\newblock {\em Operations Research}, 70(4):2563--2578.

\bibitem[Cen et~al., 2023]{cen2022faster}
Cen, S., Chi, Y., Du, S.~S., and Xiao, L. (2023).
\newblock Faster last-iterate convergence of policy optimization in zero-sum
  {M}arkov games.
\newblock In {\em ICLR}.

\bibitem[Cen et~al., 2021]{cen2021fast}
Cen, S., Wei, Y., and Chi, Y. (2021).
\newblock Fast policy extragradient methods for competitive games with entropy
  regularization.
\newblock {\em Advances in Neural Information Processing Systems},
  34:27952--27964.

\bibitem[Chakrabarty and Khanna, 2021]{chakrabarty2021better}
Chakrabarty, D. and Khanna, S. (2021).
\newblock Better and simpler error analysis of the {S}inkhorn--{K}nopp
  algorithm for matrix scaling.
\newblock {\em Mathematical Programming}, 188(1):395--407.

\bibitem[Chambolle and Contreras, 2022]{chambolle2022accelerated}
Chambolle, A. and Contreras, J.~P. (2022).
\newblock Accelerated {B}regman primal-dual methods applied to optimal
  transport and {W}asserstein {B}arycenter problems.
\newblock {\em arXiv:2203.00802}.

\bibitem[Cuturi, 2013]{cuturi2013sinkhorn}
Cuturi, M. (2013).
\newblock Sinkhorn distances: Lightspeed computation of optimal transport.
\newblock {\em Advances in neural information processing systems}, 26.

\bibitem[Daskalakis and Panageas, 2018]{daskalakis2018last}
Daskalakis, C. and Panageas, I. (2018).
\newblock Last-iterate convergence: Zero-sum games and constrained min-max
  optimization.
\newblock {\em arXiv preprint arXiv:1807.04252}.

\bibitem[Dvurechensky et~al., 2018]{dvurechensky2018computational}
Dvurechensky, P., Gasnikov, A., and Kroshnin, A. (2018).
\newblock Computational optimal transport: Complexity by accelerated gradient
  descent is better than by {S}inkhorn's algorithm.
\newblock In {\em International conference on machine learning}, pages
  1367--1376.

\bibitem[Feydy et~al., 2019]{feydy2019interpolating}
Feydy, J., S{\'e}journ{\'e}, T., Vialard, F.-X., Amari, S.-i., Trouv{\'e}, A.,
  and Peyr{\'e}, G. (2019).
\newblock Interpolating between optimal transport and mmd using {S}inkhorn
  divergences.
\newblock In {\em International Conference on Artificial Intelligence and
  Statistics}, pages 2681--2690.

\bibitem[Gayraud et~al., 2017]{gayraud2017optimal}
Gayraud, N.~T., Rakotomamonjy, A., and Clerc, M. (2017).
\newblock Optimal transport applied to transfer learning for {P300} detection.
\newblock In {\em Graz Brain-Computer Interface Conference}, page~6.

\bibitem[Geist et~al., 2019]{geist2019theory}
Geist, M., Scherrer, B., and Pietquin, O. (2019).
\newblock A theory of regularized {M}arkov decision processes.
\newblock In {\em International Conference on Machine Learning}, pages
  2160--2169. PMLR.

\bibitem[Genevay et~al., 2016]{genevay2016stochastic}
Genevay, A., Cuturi, M., Peyr{\'e}, G., and Bach, F. (2016).
\newblock Stochastic optimization for large-scale optimal transport.
\newblock {\em Advances in neural information processing systems}, 29.

\bibitem[Ghosal and Nutz, 2022]{ghosal2022convergence}
Ghosal, P. and Nutz, M. (2022).
\newblock On the convergence rate of {S}inkhorn's algorithm.
\newblock {\em arXiv preprint arXiv:2212.06000}.

\bibitem[Guminov et~al., 2021]{guminov2021combination}
Guminov, S., Dvurechensky, P., Tupitsa, N., and Gasnikov, A. (2021).
\newblock On a combination of alternating minimization and nesterov's momentum.
\newblock In {\em International Conference on Machine Learning}, pages
  3886--3898. PMLR.

\bibitem[Guo et~al., 2020]{guo2020fast}
Guo, W., Ho, N., and Jordan, M. (2020).
\newblock Fast algorithms for computational optimal transport and wasserstein
  barycenter.
\newblock In {\em International Conference on Artificial Intelligence and
  Statistics}, pages 2088--2097. PMLR.

\bibitem[Harker and Pang, 1990]{harker1990finite}
Harker, P.~T. and Pang, J.-S. (1990).
\newblock Finite-dimensional variational inequality and nonlinear
  complementarity problems: a survey of theory, algorithms and applications.
\newblock {\em Mathematical programming}, 48(1):161--220.

\bibitem[Hsieh et~al., 2019]{hsieh2019convergence}
Hsieh, Y.-G., Iutzeler, F., Malick, J., and Mertikopoulos, P. (2019).
\newblock On the convergence of single-call stochastic extra-gradient methods.
\newblock {\em NeurIPS}, 32.

\bibitem[Idel, 2016]{idel2016review}
Idel, M. (2016).
\newblock A review of matrix scaling and {S}inkhorn's normal form for matrices
  and positive maps.
\newblock {\em arXiv preprint arXiv:1609.06349}.

\bibitem[Jambulapati et~al., 2019]{jambulapati2019direct}
Jambulapati, A., Sidford, A., and Tian, K. (2019).
\newblock A direct tilde $\tilde{O}(1/\epsilon)$ iteration parallel algorithm
  for optimal transport.
\newblock {\em Advances in Neural Information Processing Systems}, 32.

\bibitem[Kalantari et~al., 2008]{kalantari2008complexity}
Kalantari, B., Lari, I., Ricca, F., and Simeone, B. (2008).
\newblock On the complexity of general matrix scaling and entropy minimization
  via the {RAS} algorithm.
\newblock {\em Mathematical Programming}, 112(2):371--401.

\bibitem[Kantorovich, 1942]{kantorovich1942translocation}
Kantorovich, L.~V. (1942).
\newblock On the translocation of masses.
\newblock In {\em Dokl. Akad. Nauk. USSR (NS)}, volume~37, pages 199--201.

\bibitem[Kim et~al., 2013]{kim2013guided}
Kim, Y.~M., Mitra, N.~J., Huang, Q., and Guibas, L. (2013).
\newblock Guided real-time scanning of indoor objects.
\newblock In {\em Computer Graphics Forum}, volume~32, pages 177--186. Wiley
  Online Library.

\bibitem[Knight, 2008]{knight2008sinkhorn}
Knight, P.~A. (2008).
\newblock The {S}inkhorn--{K}nopp algorithm: convergence and applications.
\newblock {\em SIAM Journal on Matrix Analysis and Applications},
  30(1):261--275.

\bibitem[Korpelevich, 1976]{korpelevich1976extragradient}
Korpelevich, G.~M. (1976).
\newblock The extragradient method for finding saddle points and other
  problems.
\newblock {\em Matecon}, 12:747--756.

\bibitem[Kostic et~al., 2022]{kostic2022batch}
Kostic, V.~R., Salzo, S., and Pontil, M. (2022).
\newblock Batch {G}reenkhorn algorithm for entropic-regularized multimarginal
  optimal transport: Linear rate of convergence and iteration complexity.
\newblock In {\em International Conference on Machine Learning}, pages
  11529--11558.

\bibitem[Kuhn, 1956]{kuhn1956variants}
Kuhn, H.~W. (1956).
\newblock Variants of the {H}ungarian method for assignment problems.
\newblock {\em Naval research logistics quarterly}, 3(4):253--258.

\bibitem[Lahn et~al., 2019]{lahn2019graph}
Lahn, N., Mulchandani, D., and Raghvendra, S. (2019).
\newblock A graph theoretic additive approximation of optimal transport.
\newblock {\em Advances in Neural Information Processing Systems}, 32.

\bibitem[Lan, 2022]{lan2022policy}
Lan, G. (2022).
\newblock Policy mirror descent for reinforcement learning: Linear convergence,
  new sampling complexity, and generalized problem classes.
\newblock {\em Mathematical programming}, pages 1--48.

\bibitem[Lee and Sidford, 2014]{lee2014path}
Lee, Y.~T. and Sidford, A. (2014).
\newblock Path finding methods for linear programming: Solving linear programs
  in $\widetilde{O} (\sqrt{rank})$ iterations and faster algorithms for maximum
  flow.
\newblock In {\em IEEE Annual Symposium on Foundations of Computer Science
  (FOCS)}, pages 424--433. IEEE.

\bibitem[Lehmann et~al., 2021]{Lehmann2021note}
Lehmann, T., von Renesse, M.-K., Sambale, A., and Uschmajew, A. (2021).
\newblock A note on overrelaxation in the sinkhorn algorithm.
\newblock {\em Optimization Letters}, 16(8):2209--2220.

\bibitem[Liang and Stokes, 2019]{liang2019interaction}
Liang, T. and Stokes, J. (2019).
\newblock Interaction matters: A note on non-asymptotic local convergence of
  generative adversarial networks.
\newblock In {\em AISTATS}, pages 907--915.

\bibitem[Lin et~al., 2022]{lin2022efficiency}
Lin, T., Ho, N., and Jordan, M.~I. (2022).
\newblock On the efficiency of entropic regularized algorithms for optimal
  transport.
\newblock {\em Journal of Machine Learning Research}, 23(137):1--42.

\bibitem[Mai et~al., 2022]{mai2022a}
Mai, V.~V., Lindb{\"a}ck, J., and Johansson, M. (2022).
\newblock A fast and accurate splitting method for optimal transport: analysis
  and implementation.
\newblock In {\em ICLR}.

\bibitem[McKelvey and Palfrey, 1995]{mckelvey1995quantal}
McKelvey, R.~D. and Palfrey, T.~R. (1995).
\newblock Quantal response equilibria for normal form games.
\newblock {\em Games and economic behavior}, 10(1):6--38.

\bibitem[Mei et~al., 2020]{mei2020global}
Mei, J., Xiao, C., Szepesvari, C., and Schuurmans, D. (2020).
\newblock On the global convergence rates of softmax policy gradient methods.
\newblock In {\em ICML}, pages 6820--6829.

\bibitem[Mertikopoulos et~al., 2018a]{mertikopoulos2018optimistic}
Mertikopoulos, P., Lecouat, B., Zenati, H., Foo, C.-S., Chandrasekhar, V., and
  Piliouras, G. (2018a).
\newblock Optimistic mirror descent in saddle-point problems: Going the extra
  (gradient) mile.
\newblock In {\em International Conference on Learning Representations}.

\bibitem[Mertikopoulos et~al., 2018b]{mertikopoulos2018cycles}
Mertikopoulos, P., Papadimitriou, C., and Piliouras, G. (2018b).
\newblock Cycles in adversarial regularized learning.
\newblock In {\em Symposium on Discrete Algorithms}, pages 2703--2717. SIAM.

\bibitem[Mertikopoulos and Sandholm, 2016]{mertikopoulos2016learning}
Mertikopoulos, P. and Sandholm, W.~H. (2016).
\newblock Learning in games via reinforcement and regularization.
\newblock {\em Mathematics of Operations Research}, 41(4):1297--1324.

\bibitem[Mokhtari et~al., 2020a]{mokhtari2020unified}
Mokhtari, A., Ozdaglar, A., and Pattathil, S. (2020a).
\newblock A unified analysis of extra-gradient and optimistic gradient methods
  for saddle point problems: Proximal point approach.
\newblock In {\em International Conference on Artificial Intelligence and
  Statistics}, pages 1497--1507.

\bibitem[Mokhtari et~al., 2020b]{mokhtari2020convergence}
Mokhtari, A., Ozdaglar, A.~E., and Pattathil, S. (2020b).
\newblock Convergence rate of ${O}(1/k)$ for optimistic gradient and
  extragradient methods in smooth convex-concave saddle point problems.
\newblock {\em SIAM Journal on Optimization}, 30(4):3230--3251.

\bibitem[Munkres, 1957]{munkres1957algorithms}
Munkres, J. (1957).
\newblock Algorithms for the assignment and transportation problems.
\newblock {\em Journal of the society for industrial and applied mathematics},
  5(1):32--38.

\bibitem[Nemirovski, 2004]{nemirovski2004prox}
Nemirovski, A. (2004).
\newblock Prox-method with rate of convergence ${O}(1/t)$ for variational
  inequalities with {L}ipschitz continuous monotone operators and smooth
  convex-concave saddle point problems.
\newblock {\em SIAM Journal on Optimization}, 15(1):229--251.

\bibitem[Nesterov, 2007]{nesterov2007dual}
Nesterov, Y. (2007).
\newblock Dual extrapolation and its applications to solving variational
  inequalities and related problems.
\newblock {\em Mathematical Programming}, 109(2):319--344.

\bibitem[Neu et~al., 2017]{neu2017unified}
Neu, G., Jonsson, A., and G{\'o}mez, V. (2017).
\newblock A unified view of entropy-regularized markov decision processes.
\newblock {\em arXiv preprint arXiv:1705.07798}.

\bibitem[Pele and Werman, 2009]{pele2009fast}
Pele, O. and Werman, M. (2009).
\newblock Fast and robust earth mover's distances.
\newblock In {\em 2009 IEEE 12th international conference on computer vision},
  pages 460--467. IEEE.

\bibitem[Peyr{\'e} et~al., 2019]{peyre2019computational}
Peyr{\'e}, G., Cuturi, M., et~al. (2019).
\newblock Computational optimal transport: With applications to data science.
\newblock {\em Foundations and Trends{\textregistered} in Machine Learning},
  11(5-6):355--607.

\bibitem[Quanrud, 2018]{quanrud2018approximating}
Quanrud, K. (2018).
\newblock Approximating optimal transport with linear programs.
\newblock {\em arXiv preprint arXiv:1810.05957}.

\bibitem[Rakhlin and Sridharan, 2013]{rakhlin2013optimization}
Rakhlin, A. and Sridharan, K. (2013).
\newblock Optimization, learning, and games with predictable sequences.
\newblock {\em arXiv preprint arXiv:1311.1869}.

\bibitem[Rubner et~al., 2000]{rubner2000earth}
Rubner, Y., Tomasi, C., and Guibas, L.~J. (2000).
\newblock The earth mover's distance as a metric for image retrieval.
\newblock {\em International journal of computer vision}, 40(2):99--121.

\bibitem[Savas et~al., 2019]{savas2019entropy}
Savas, Y., Ahmadi, M., Tanaka, T., and Topcu, U. (2019).
\newblock Entropy-regularized stochastic games.
\newblock In {\em 2019 IEEE 58th Conference on Decision and Control (CDC)},
  pages 5955--5962. IEEE.

\bibitem[Sedrakyan and Sedrakyan, 2018]{sedrakyan2018algebraic}
Sedrakyan, H. and Sedrakyan, N. (2018).
\newblock {\em Algebraic inequalities}.
\newblock Springer.

\bibitem[Sherman, 2017]{sherman2017area}
Sherman, J. (2017).
\newblock Area-convexity, $\ell_{\infty}$ regularization, and undirected
  multicommodity flow.
\newblock In {\em Annual ACM SIGACT Symposium on Theory of Computing}, pages
  452--460.

\bibitem[Sinkhorn, 1967]{sinkhorn1967diagonal}
Sinkhorn, R. (1967).
\newblock Diagonal equivalence to matrices with prescribed row and column sums.
\newblock {\em The American Mathematical Monthly}, 74(4):402--405.

\bibitem[Solomon et~al., 2015]{solomon2015convolutional}
Solomon, J., De~Goes, F., Peyr{\'e}, G., Cuturi, M., Butscher, A., Nguyen, A.,
  Du, T., and Guibas, L. (2015).
\newblock Convolutional {W}asserstein distances: Efficient optimal
  transportation on geometric domains.
\newblock {\em ACM Transactions on Graphics}, 34(4):1--11.

\bibitem[Tseng, 1995]{tseng1995linear}
Tseng, P. (1995).
\newblock On linear convergence of iterative methods for the variational
  inequality problem.
\newblock {\em Journal of Computational and Applied Mathematics},
  60(1-2):237--252.

\bibitem[Tsybakov, 2009]{tsybakov2004introduction}
Tsybakov, A.~B. (2009).
\newblock {\em Introduction to nonparametric estimation}.
\newblock Springer.

\bibitem[van~den Brand et~al., 2020]{van2020bipartite}
van~den Brand, J., Lee, Y.-T., Nanongkai, D., Peng, R., Saranurak, T., Sidford,
  A., Song, Z., and Wang, D. (2020).
\newblock Bipartite matching in nearly-linear time on moderately dense graphs.
\newblock {\em IEEE Annual Symposium on Foundations of Computer Science}, pages
  919--930.

\bibitem[Villani, 2009]{villani2009optimal}
Villani, C. (2009).
\newblock {\em Optimal transport: old and new}, volume 338.
\newblock Springer.

\bibitem[von Neumann, 1928]{v1928theorie}
von Neumann, J. (1928).
\newblock Zur theorie der gesellschaftsspiele.
\newblock {\em Mathematische annalen}, 100(1):295--320.

\bibitem[Wei et~al., 2021]{wei2020linear}
Wei, C.-Y., Lee, C.-W., Zhang, M., and Luo, H. (2021).
\newblock Linear last-iterate convergence in constrained saddle-point
  optimization.
\newblock In {\em ICLR}.

\bibitem[Werman et~al., 1985]{werman1985distance}
Werman, M., Peleg, S., and Rosenfeld, A. (1985).
\newblock A distance metric for multidimensional histograms.
\newblock {\em Computer Vision, Graphics, and Image Processing},
  32(3):328--336.

\bibitem[Xie et~al., 2022a]{xie2022accelerated}
Xie, Y., Luo, Y., and Huo, X. (2022a).
\newblock An accelerated stochastic algorithm for solving the optimal transport
  problem.
\newblock {\em arXiv preprint arXiv:2203.00813}.

\bibitem[Xie et~al., 2022b]{xie2022solving}
Xie, Y., Luo, Y., and Huo, X. (2022b).
\newblock Solving a special type of optimal transport problem by a modified
  {H}ungarian algorithm.
\newblock {\em arXiv preprint arXiv:2210.16645}.

\bibitem[Zhan et~al., 2023]{zhan2021policy}
Zhan, W., Cen, S., Huang, B., Chen, Y., Lee, J.~D., and Chi, Y. (2023).
\newblock Policy mirror descent for regularized reinforcement learning: A
  generalized framework with linear convergence.
\newblock {\em SIAM Journal on Optimization}.

\end{thebibliography}

\end{document}